\documentclass[11pt,a4paper]{article}

\usepackage{microtype}
\usepackage{graphicx}
\usepackage{subfigure}
\usepackage{booktabs}

\usepackage{amssymb}
\usepackage{algorithm}
\usepackage{algpseudocode}
\usepackage{amsmath,bm}
\usepackage{array}
\usepackage{natbib}
\usepackage{enumerate}
\usepackage{hyperref}
\usepackage{cleveref}

\graphicspath{{./figs1/}{./figs2/}{./figs3/}{./figs4/}}

\newcommand{\bs}{\boldsymbol}

\newcommand{\mb}{\mathbf}

\begin{document}

\title{Learning pose variations within shape population\\
	by constrained mixtures of factor analyzers}
\author{Xilu Wang\\
        xwang666@wisc.edu}
\date{}
\maketitle

\begin{abstract}
Mining and learning the shape variability of underlying population has benefited the applications including parametric shape modeling, 3D animation, and image segmentation. The current statistical shape modeling method works well on learning unstructured shape variations without obvious pose changes (relative rotations of the body parts). Studying the pose variations within a shape population involves segmenting the shapes into different articulated parts and learning the transformations of the segmented parts. This paper formulates the pose learning problem as mixtures of factor analyzers. The segmentation is obtained by components posterior probabilities and the rotations in pose variations are learned by the factor loading matrices. To guarantee that the factor loading matrices are composed by rotation matrices, constraints are imposed and the corresponding closed form optimal solution is derived. Based on the proposed method, the pose variations are automatically learned from the given shape populations. The method is applied in motion animation where new poses are generated by interpolating the existing poses in the training set. The obtained results are smooth and realistic.
\end{abstract}

\section{Introduction}

Statistical shape modeling (SSM) \cite{heimann2009statistical,dryden1998statistical} has become a powerful tool of learning the shape variations from given population, which has been widely used in a variety of applications including image segmentation \cite{heimann2009statistical}, 3D animation \cite{Hasler2009A}, parametric shape modeling \cite{baek2012parametric,chu2010exemplar}, and shape matching \cite{wang2016statistical}. SSM assumes that the shapes are Gaussian distributed and use principal component analysis \cite{dryden1998statistical} to extract the mean shape and eigen shapes which span a linear shape space. The above assumption is accurate when the population does not contain large pose variations. Otherwise in application like motion animation \cite{Hasler2009A}, where relative rotations of the body parts are involved, the underlying shape space is nonlinear and the linear interpolation of any two shapes will not produce a valid shape.

To provide better shape modeling, the Shape Completion and Animation for PEople (SCAPE) method \cite{Anguelov2005SCAPE} separates pose variations from unstructured shape variations using articulated skeleton model. The pose variations are modeled by the rigid motions (rotations and translations) of the articulated segments of the skeleton. The skeleton is obtained from the articulated object model in \cite{Anguelov2012Recovering}, which is automatically recovered by the Markov Network algorithm given a set of registered \cite{amberg2007optimal} shapes in different poses represented by triangle meshes. The Markov Network algorithm finds the segmentation of the shapes and rigid transformations of the segmented parts that maximize the posterior probability of the input shapes conditioned on the template shape. To guarantee that the segmentation is spatially coherent, soft and hard contiguity constraints are imposed, which makes it difficult to fine tune the weight that balances the data probability and the contiguity constraints. The Markov Network algorithm also doesn't work well on the objects that have large non-rigid deformations (compared to articulated objects) and annealing is exploited in such situation \cite{Anguelov2012Recovering}.

In this study we find that clustering and dimensionality reduction are the essences of recovering articulated object model from registered triangle meshes of the training shapes. Assigning each vertex a label indicating the corresponding articulated part is clustering. Finding the rotations and translations that map the points on the same corresponding positions of the training shapes to a single point on the reference shape is dimensionality reduction. Mixtures of factor analyzers (MFA) \cite{ghahramani1996algorithm,MclachlanModelling} conducting clustering and dimensionality reduction simultaneously whose convergence is proved \cite{MclachlanModelling,Meng1993Maximum,meng1997algorithm}. Different from the Markov Network algorithm in \cite{Anguelov2012Recovering}, MFA maximizes the complete data log-probability and the latent variable is assumed to be Gaussian distributed which automatically penalizes the spatially-disjoint segmentation results thus guarantees the spatial contiguity. The only hyper-parameter in MFA is the number of clusters.

This paper formulates the problem of learning the pose variations from given shape population as the problem of mixtures of factor analyzers. Assume we have $n_s$ number of training shapes and each training shape is sampled by $n_v$ number of points. The inputs to the MFA algorithm are $n_v$ number of data vectors each is obtained by concatenating the points on the same corresponding positions of the $n_s$ training shapes. The outputs are the factor analyzers composed by the mixture proportions, the factor loading matrices, and the mixture variances. Each factor analyzer corresponds to an articulated part of the shapes. The vertices' labels are calculated from the component posterior probability of the data vectors. To avoid arbitrary factor loading matrices, constraints are imposed and the corresponding closed form optimal solution is derived. Thought in \cite{Baek2010Mixtures,tang2012deep,MclachlanModelling} different constraints are imposed on the factor loading matrices, the goal is to reduce unnecessary degrees of freedom so to avoid bad local minimums. In this paper the loading matrices are explicitly constrained to be composed by rotation matrices. Based on the proposed method, the pose variations are automatically learned from the given shape populations. The method is applied in motion animation where new poses are generated by interpolating the existing poses in the training set. The obtained results are smooth and realistic. The shape data used in this paper is from \cite{sumner2004deformation}.

This paper is organized as follows, in Section 2 the problem of learning pose variations from given shape population is formulated as problem of mixtures of factor analyzers. In Section 3 the hierarchical optimization approach that automatically refines the initial MFA result is presented. In Section 4 the algorithm of pose interpolation is demonstrated. Section 5 shows the experimental results on different shapes. This paper is concluded in Section 6. The derivation of the closed form optimal solution is elaborated in the Appendix.

\section{Related Work}

In \cite{Schaefer2007Example} an example based skeleton extraction approach is proposed. Given a set of example meshes that share the same mesh connectivity, the approach uses hierarchical face clustering to group adjacent mesh elements into different mesh regions each corresponds to a rigid part of the mesh body. In the hierarchical clustering, the adjacent mesh elements are merged by the error of rigid transformations from the corresponding mesh elements of the example meshes to the reference mesh. Though spatial coherency is automatically imposed in \cite{Schaefer2007Example} by merging adjacent mesh regions, updating the rigid transformation error in each iteration is a costly operation and the hierarchical clustering is sensitive to local nonrigid transformations.

In \cite{de2008automatic} a spectral clustering approach is proposed to segment the mesh examples into different rigid parts. The inputs to the spectral clustering are the motion trajectories of the seed vertices among the example meshes and the outputs are $k$ clusters of vertices corresponding to $k$ rigid parts. The time complexity of spectral clustering is $O(N^3)$ due to eigen-analysis of the affinity matrix, where $N$ is the number of vertices. That's why instead of the full mesh vertices, seed vertices are used which are obtained by curvature based segmentation approach.

In \cite{James2005Skinning} mean shift clustering of rotation sequences of the mesh elements is used to segment the example shapes into corresponding rigid parts. Each rotation sequence is composed by the rotation matrices between the triangle elements of the example shapes and the corresponding element of the reference shape. However, the segmentation obtained is lack of spatial coherency and shows broken patches in the areas of non-rigid deformation (e.g. shoulders).

In \cite{tierny2008fast} a Reeb graph based approach is proposed for kinematic skeleton extraction of 3D dynamic mesh sequences. The contours of the Reeb graph that locally maximizes the local length deviation metric are identified as motion boundaries by which the kinematic skeleton is extracted. The local length deviation metric measures the extent of local length preservation across the mesh sequences. Though simple and efficient, the Reeb graph based approach generally suffers from robustness issues due to the discrete nature of mesh.

In \cite{le2012smooth} K-means clustering is used to cluster the vertices of example meshes into different rigid groups. In the assignment step of the K-means clustering, each vertex is associated to the group that has the smallest rigid transformation error. The rigid transformation from each vertex group of an example shape to the reference shape is calculated by the Kabsch algorithm \cite{kabsch1978discussion}. Similarly, the drawback of the above approach is lacking of spatial coherency.


Recently, deep auto-encoders \cite{hinton2006reducing} has been applied in learning shape variations from a population \cite{shu2018deforming,nash2017shape,li2017grass}. In \cite{shu2018deforming} deforming autoencoders for images are introduced to disentangle shape from appearance. However, the learned field deformations is not capable of describing the articulated motions. In \cite{nash2017shape}, given a set of part-segmented objects with dense point correspondences, the shape variational auto-encoder (ShapeVAE) is capable of synthesizing novel, realistic shapes. However, the method requires pre-segmented training shapes. In \cite{li2017grass} a recursive neural net (RvNN) based autoencoder is proposed for encoding and synthesis of 3D shapes, which are represented by voxels. The main limitations of the above approaches are: 1) only take data with an underlying Euclidean structure (e.g. 1D sequences, 2D or 3D images) thus can not capture articulated motions of the pose variations; 2) require large number of training data of the same class of shapes (e.g. 3701/1501 in \cite{nash2017shape}) which are not always available due to the tedious process of obtaining neat 3D shape models from either 3D scanning (e.g. scanning $\Rightarrow$ point clouds $\Rightarrow$ boundary triangulation) or images (e.g. image $\Rightarrow$ segmentation $\Rightarrow$ boundary triangulation), while the proposed MFA based approach is capable of preciously learning the pose variations from just tens of shape models.

In the review paper \cite{bronstein2017geometric} geometric deep learning techniques that generalize (structured) deep neural models to non-Euclidean domains such as graphs and manifolds are introduced. In \cite{litany2017deep} a deep residual network is proposed that takes dense descriptor fields defined on two shapes as input, and outputs a soft map between the two given objects. In \cite{boscaini2016learning} an Anisotropic Convolutional Neural Network (ACNN) is proposed to learn the dense correspondences between deformable shapes where classical convolutions are replaced by projections over a set of oriented anisotropic diffusion kernels. In \cite{yi2017syncspeccnn} a Synchronized Spectral CNN is proposed for 3D Shape Segmentation. However, until now I have seen any literature in geometric deep learning that aims articulated shape segmentation and learns pose variations from underlying shape population.

\section{Learning Pose Variations within Shape Population}

The goal of this work is to learn the pose variations within the given shape population. To be more specific, it learns: 1) the vertex labels that segment the shape into a set of rigid parts whose shape is approximately invariant across the different poses; 2) the rotations and translations associated with the rigid parts across the different poses; 3) muscle movements of the rigid parts of the pose variations.

\begin{figure}[ht]
	\centering
	\subfigure{
		\includegraphics[width=0.22\textwidth,trim=1cm 4cm 1cm 5cm, clip]{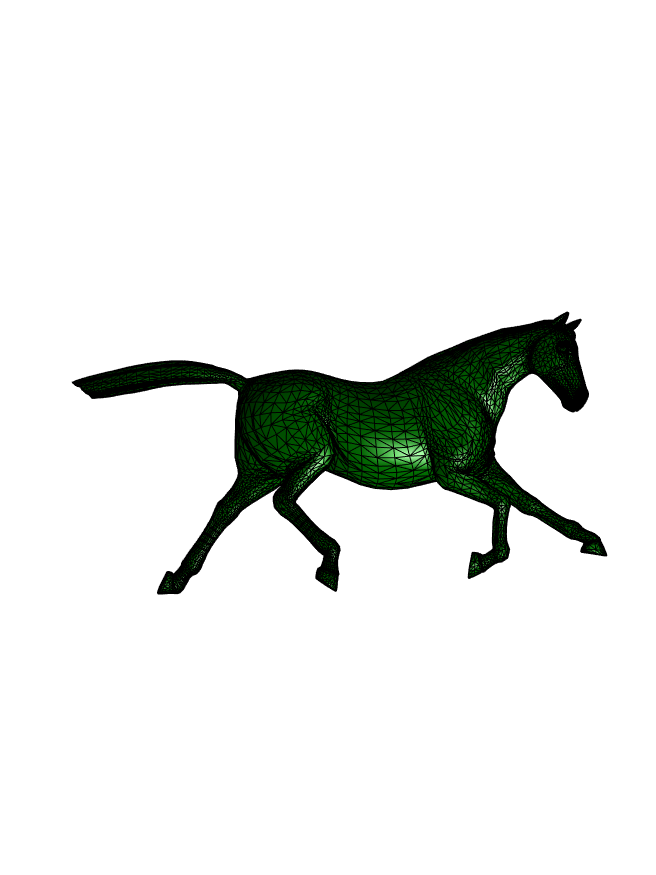}
		\label{fig:11horsesA}
	}
	\subfigure{
		\includegraphics[width=0.24\textwidth,trim=1cm 4.5cm 0.5cm 4.5cm, clip]{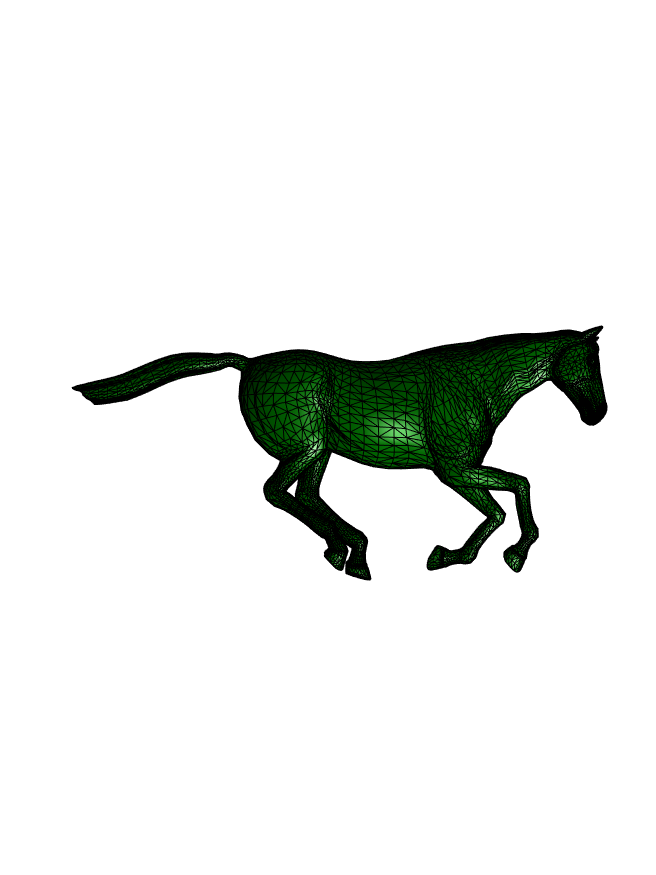}
	}
	\subfigure{
		\includegraphics[width=0.16\textwidth,trim=1cm 3cm 0.5cm 3cm, clip]{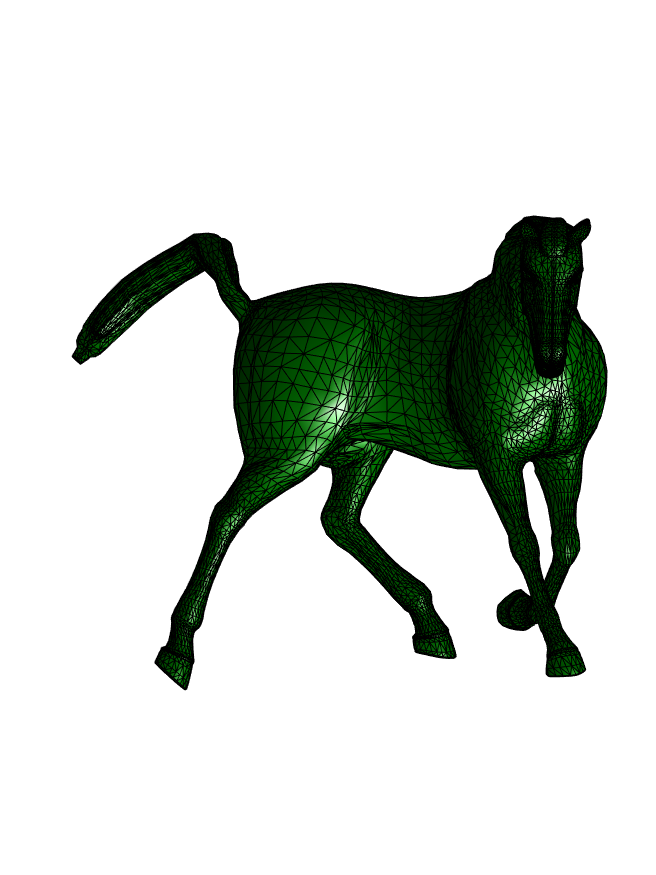}
	}
	\subfigure{
		\includegraphics[width=0.2\textwidth,trim=1cm 4cm 0.5cm 5cm, clip]{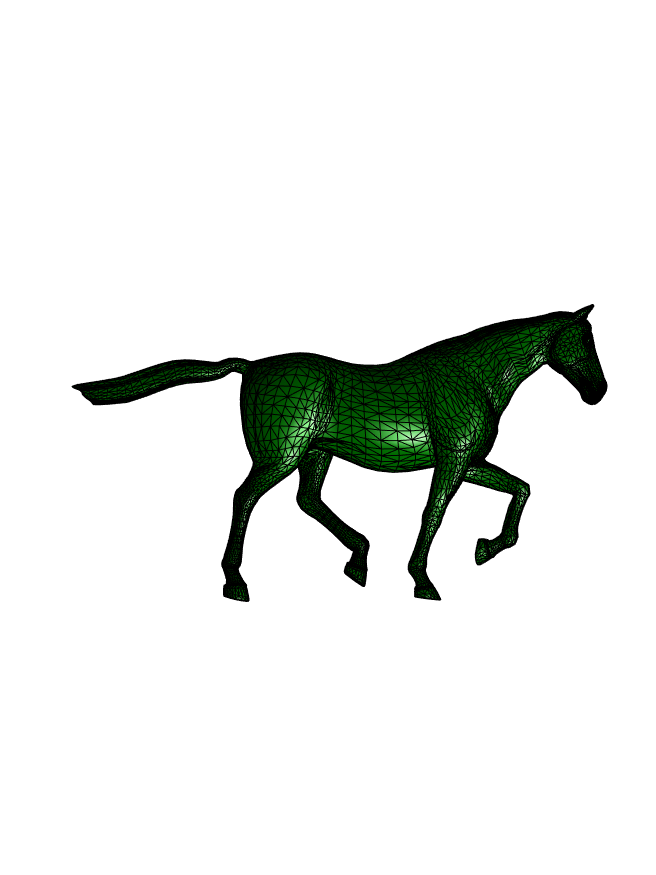}
	}
	\subfigure{
		\includegraphics[width=0.2\textwidth,trim=1cm 4cm 0.5cm 4.5cm, clip]{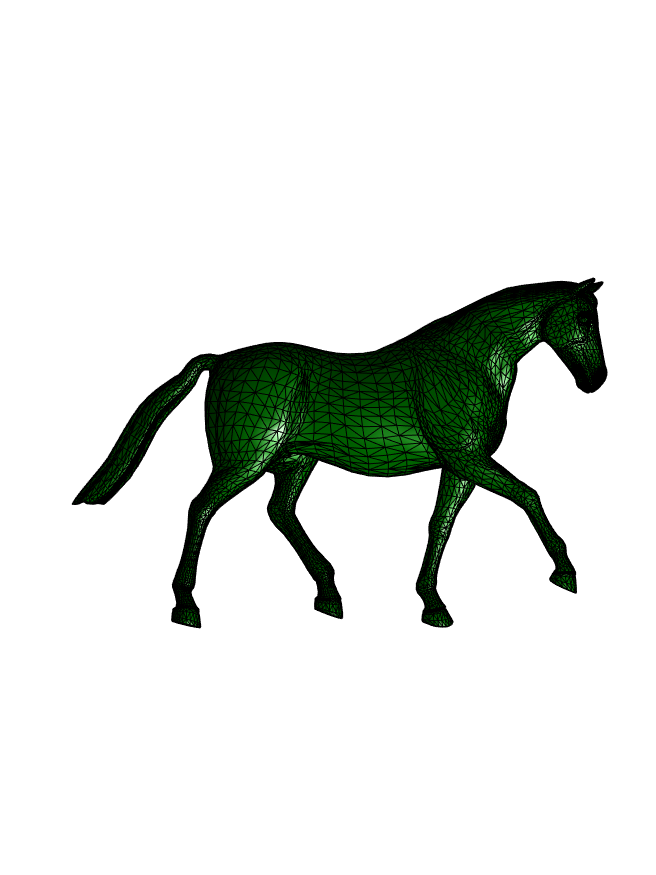}
	}
	\subfigure{
		\includegraphics[width=0.2\textwidth,trim=1cm 4cm 0.5cm 4.5cm, clip]{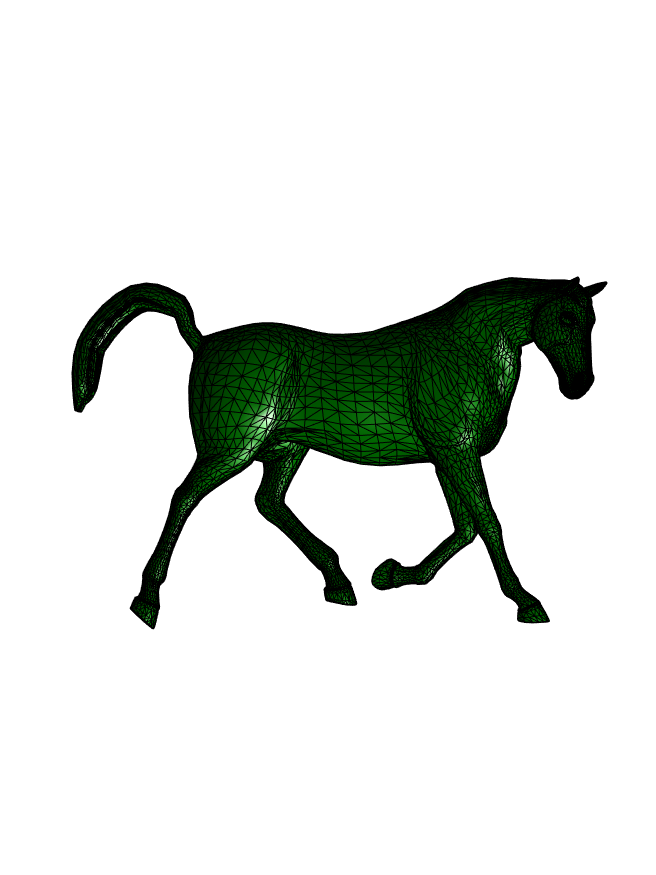}
	}
	\subfigure{
		\includegraphics[width=0.22\textwidth,trim=1cm 4.5cm 1cm 5cm, clip]{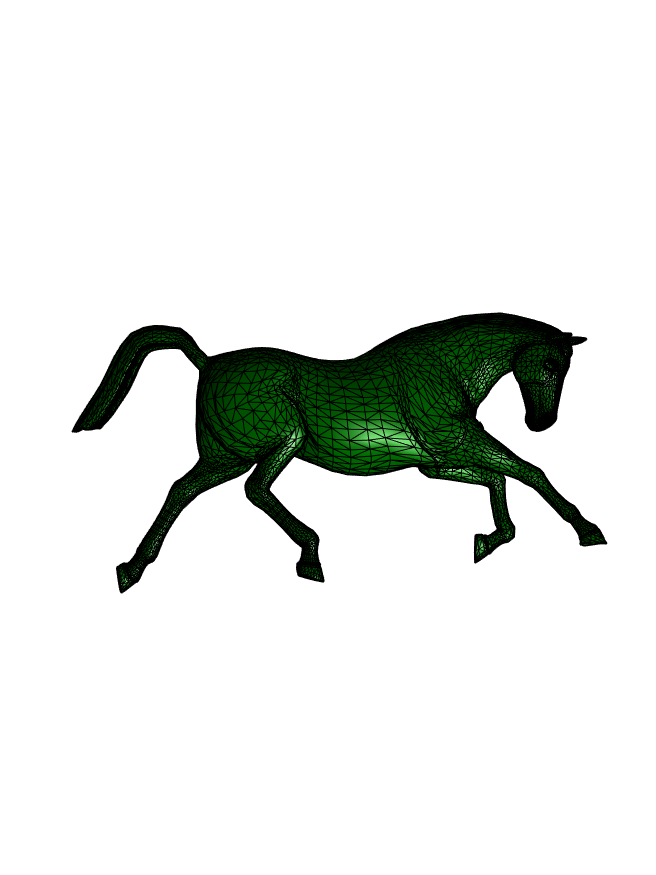}
	}
	\subfigure{
		\includegraphics[width=0.18\textwidth,trim=1cm 3cm 0.5cm 3.5cm, clip]{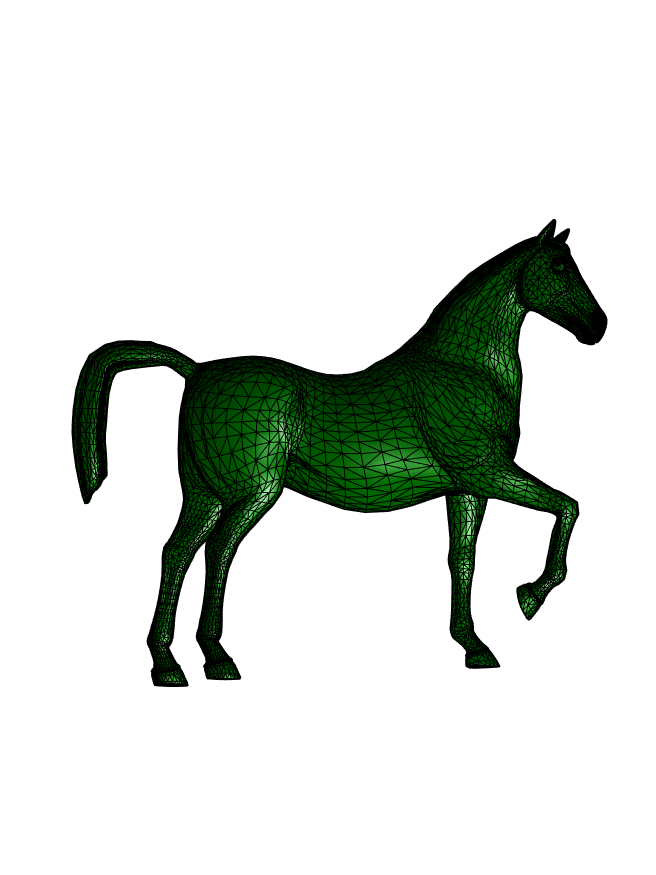}
	}
	\subfigure{
		\includegraphics[width=0.2\textwidth,trim=1cm 4cm 0.5cm 5cm, clip]{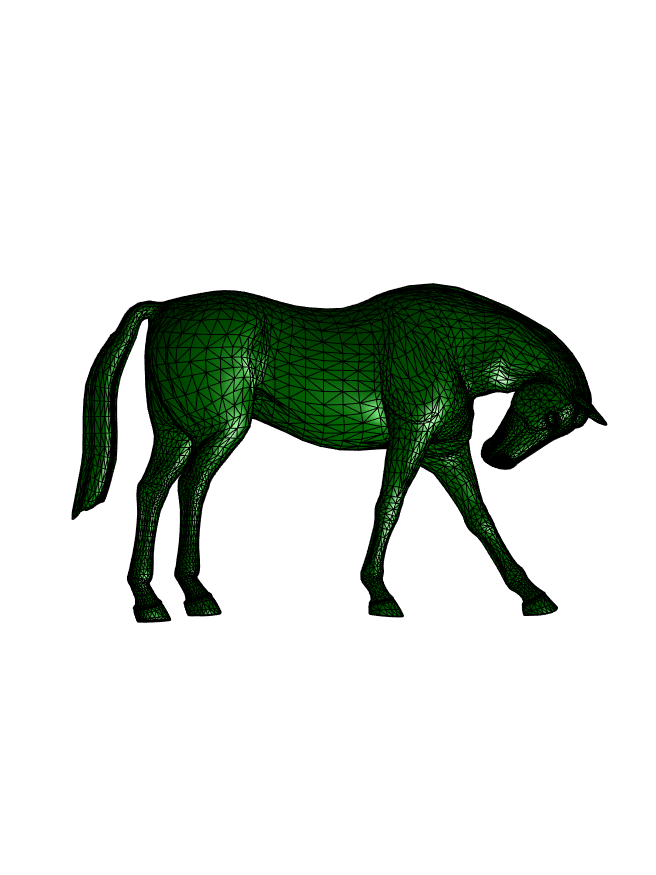}
	}
	\subfigure{
		\includegraphics[width=0.19\textwidth,trim=1cm 3.5cm 0.5cm 4cm, clip]{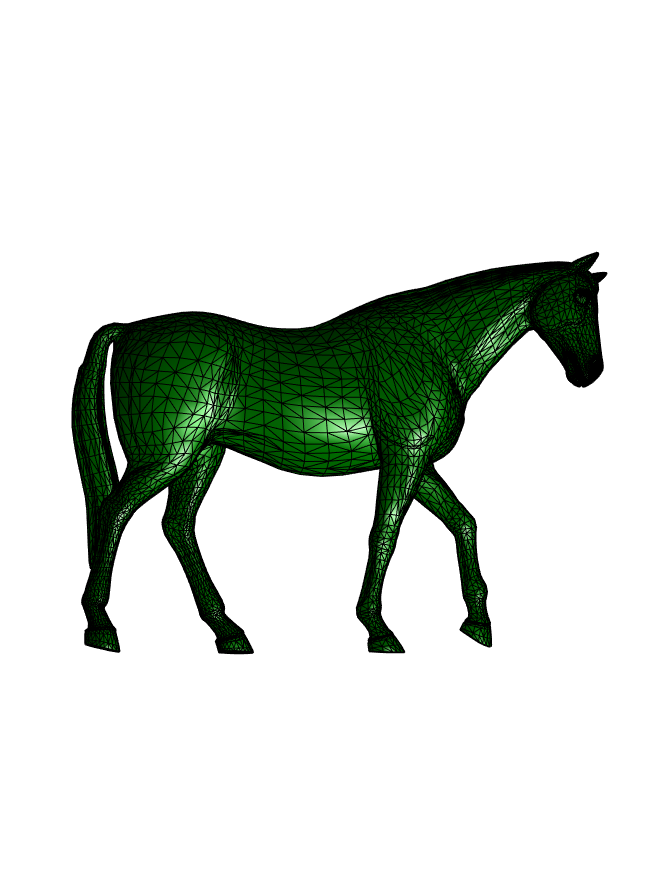}
	}
	\subfigure{
		\includegraphics[width=0.2\textwidth,trim=1cm 3cm 0.5cm 3cm, clip]{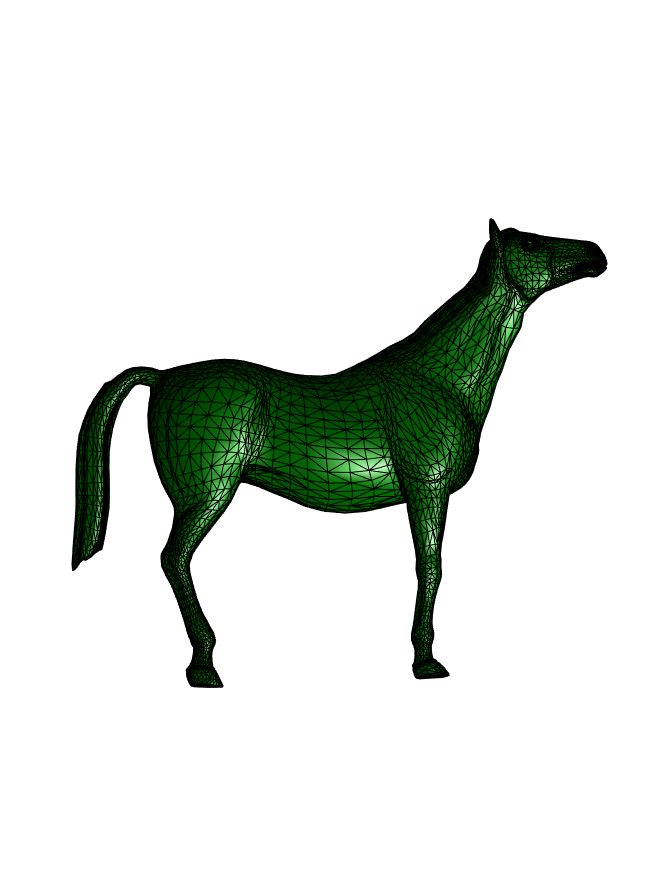}
	}
	\caption{The 11 training shapes in different poses.}
	\label{fig:11horses}
\end{figure}

As shown in Figure \ref{fig:11horses} we have a set of training shapes $\mb S_1, \mb S_2,..., \mb S_{n_s}$ in triangle meshes that are sampled by the same number of points in correspondence:
\begin{equation}
\mb S_i = [\mb v_1^i, \mb v_2^i,...,\mb v_{n_v}^i], i = 1,...,n_s,
\end{equation}
where $\mb v_j^i$ is vertex $j$ on the $i$th shape.

Assume that each training shape is composed by $m$ rigid parts, and assume $I_j \in \{1, 2, ..., m\}$ is the label of vertex $j$ indicating the part Id, we have:
\begin{eqnarray}
\mb v_j^i = \mb R_{I_j}^i\mb v_j^{ref} + \mb b_{I_j}^i + \boldsymbol\epsilon_{j}^i,j = 1,...,n_v,i = 1,...,n_s,
\label{eqn:poseMovement}
\end{eqnarray}
where $\mb v_j^{ref}$ is the $j$th vertex on the reference shape in the latent space, $\mb R_{I_j}^i$ and $\mb b_{I_j}^i$ are the rotation and translation associated with rigid part $I_j$ of shape $\mb S_i$, and $\boldsymbol\epsilon_{j}^i$ is the residual caused by the muscle movement of the pose variation. It should be noted that the rotations and translations are the same for the vertices belong to the same rigid part of the same shape, otherwise different.

Concatenate the rigid motions of the vertices on the same corresponding positions of the $n_s$ shapes in \eqref{eqn:poseMovement} we have:
\begin{equation}
\mb h_j = \mb R_{I_j}\mb v_j^{ref}  + \mb b_{I_j} + \bs\epsilon_j, j = 1,...,n_v,
\label{eqn:dimensionReductRaw}
\end{equation}
where
\begin{eqnarray}
\mb h_j = \left[\begin{array}{c}
\mb v_j^1\\
\vdots\\
\mb v_j^{n_s}
\end{array}\right] 
,
\mb R_{I_j} = \left[\begin{array}{c}
\mb R_{I_j}^1\\
\vdots\\
\mb R_{I_j}^{n_s}
\end{array}\right]
,
\mb b_{I_j} = \left[\begin{array}{c}
\mb b_{I_j}^1\\
\vdots\\
\mb b_{I_j}^{n_s}
\end{array}\right]
,
\bs\epsilon_j = \left[\begin{array}{c}
\boldsymbol\epsilon_{j}^1\\
\vdots\\
\boldsymbol\epsilon_{j}^{n_s}
\end{array}\right].\nonumber
\end{eqnarray}
From Equation \eqref{eqn:dimensionReductRaw} we see that the high-dimensional vector $\mb h_j$ has its image $\mb v_j^{ref}$ in the low-dimensional space. Our goal is to learn the vertex labels $I_j,j=1,...,n_v$, the rotations and translations $\mb R_k^{i}, \mb b_k^{i},k=1,...,m,i=1,...n_s$, and the muscle movements $\boldsymbol\epsilon_j^i,j=1,...,n_v,i=1,...n_s$, which is essentially a clustering problem with dimensionality reduction, and the mixtures of factor analyzers is a natural choice to address the problem. Differently from general mixtures of factor analyzers, as shown by equation \eqref{eqn:dimensionReductRaw}, the factor loadings are composed by rotation matrices, which forms our constraints.

\subsection{Mixtures of Factor Analyzers}

We formulate the problem as mixtures of factor analyzers:
\begin{equation}
\mb h = \mb A_k\mb z + \mb b_k + \boldsymbol\epsilon_k,~k=1,...,m,
\label{eqn:dimensionReductFinal}
\end{equation}
where $\mb h$ is the random variable in the high dimensional space whose samplings are $\{\mb h_j\}$, $\mb z$ is the latent variable which is normally distributed with zero mean and unit diagonal covariance matrix, $\mb A_k=\mb R_{k}\bs\Lambda_k$ is the factor loading matrix of the $k$th mixture, $\bs\Lambda_k$ is the diagonal matrix that scales the dimensions of $\mb z$, $\mb b_k$ is the corresponding mean vector, and $\boldsymbol\epsilon_k$ is the residual vector of the $k$th mixture which is Gaussian distributed with zero mean and diagonal covariance matrix $\bs\Phi_k$. It should be noted that ${\mb v_j^{ref}}$ could be viewed as samplings of $\bs\Lambda_{I_j}E(z|\mb h_j,I_j)$ in the latent space.

In the following $\boldsymbol{\Theta}_k = \{\mb A_k, \mb b_k, \bs\Phi_k, \pi_k\}$ is used to represent the parameters of the $k$th mixture, where $\pi_k$ is the probability that an arbitrary data belongs to the $k$th mixture, and $\boldsymbol{\Theta} = \{\boldsymbol{\Theta}_k, k = 1,2,...,m\}$ is used to represent the full parameter sets of the mixtures.

Since $\mb z$ and $\bs\epsilon_k$ are Gaussian distributed, from Equation \eqref{eqn:dimensionReductFinal} we have that the joint probability of $\mb h$ and $\mb z$ under the $k$th mixture is also Gaussian distributed:
\begin{eqnarray}
P(\mb h, \mb z|\bs\Theta_k) \sim N(
\left[
\begin{array}{c}
\mb b_k\\
\mb 0
\end{array}
\right] 
,
\left[
\begin{array}{cc}
\mb A_k\mb A_k^T + \bs\Phi_k & \mb A_k\\
\mb A_k^T & \mb I
\end{array}
\right] 
),
\label{eqn:MFAJointProbability}
\end{eqnarray} 
where $N(\cdot,\cdot)$ stands for Gaussian distribution with the first parameter its mean and the second parameter its covariance. From Equation \ref{eqn:MFAJointProbability} we have the marginal distribution of $\mb h$ under the $k$th mixture:
\begin{equation}
P(\mb h|\bs\Theta_k) \sim N(\mb b_k,\mb A_k\mb A_k^T + \bs\Phi_k),
\label{eqn:PhMarginal}
\end{equation}
the conditional distribution of $\mb h$ given $\mb z$:
\begin{eqnarray}
\mb P(\mb h | \mb z,\bs\Theta_k) \sim N(\mb A_k\mb z + \mb b_k,\bs\Phi_k),
\label{eqn:PhConditional}
\end{eqnarray}
and the conditional distribution of $\mb z$ given $\mb h$:
\begin{equation}
P(\mb z | \mb h,\bs\Theta_k) \sim N(\bs\beta_k(\mb h - \mb b_k), \mb I - \bs\beta_k\mb A_k),
\label{eqn:PzConditional}
\end{equation}
where $\bs\beta_k = \mb A_k^T(\mb A_k\mb A_k^T + \bs\Phi_k)^{-1}$.

Our goal is to find the parameters $\boldsymbol{\Theta}$ that maximize the posterior probability of the data $\{\mb h_j,j=1,...,n_v\}$:
\begin{equation}
P(\mb S | \boldsymbol{\Theta}) = \prod_{j = 1}^{n_v} \sum_{k=1}^{m}\pi_kP(\mb h_j | \bs\Theta_k),
\label{eqn:PosteriorDataRaw1}
\end{equation}
where $P(\mb h_j | \bs\Theta_k)$ is the posterior probability of $\mb h_j$ given that it belongs to the $k$th mixture and is calculated by Equation \eqref{eqn:PhMarginal}. Taking logarithm of equation \eqref{eqn:PosteriorDataRaw1} and set it as the objective function we have
\begin{eqnarray}
\max_{\bs\Theta} \log P(\mb S | \boldsymbol{\Theta}) =\sum_{j = 1}^{n_v} \log\left(\sum_{k=1}^{m}\pi_kP(\mb h_j | \bs\Theta_k)\right).
\label{eqn:LikelihoodData}
\end{eqnarray}

\subsection{Alternating Expectation–Conditional Maximization}

Optimization formula \eqref{eqn:LikelihoodData} is in the form of mixtures of Gaussian, which does not have closed form solution but can be efficiently maximized by the algorithm of Expectation Maximization (EM) or Alternating Expectation–Conditional Maximization (AECM) algorithm. AECM replaces the M-step of the EM algorithm by computationally simpler conditional maximization steps (CM-step). Here we choose AECM algorithm due to its computational simplicity and faster convergence.

Assume $\boldsymbol{\Theta}^{l}= \{ \mb A_k^l, \mb b_k^l, \bs\Phi_k^l, \tilde{\pi}_k^l, k = 1,2,...,m\}$ are estimations of the parameters in the $l$th iteration, by Expectation Maximization we have:
\begin{eqnarray}
\boldsymbol{\Theta}^{l+1} = \max_{\bs\Theta} && \sum_{j = 1}^{n_v}E_{p(k,\mb z|\mb h_j,\bs\Theta^{l})}\log P(\mb h_j, k, \mb z | \bs\Theta)\nonumber\\
\Rightarrow\max_{\bs\Theta} && \sum_{j = 1}^{n_v}\left(\sum_{k=1}^{m}P(k | \mb h_j, \bs\Theta^{l})\int_{\mb z}P(\mb z|\mb h_j, \bs\Theta_k^l)\log P(\mb h_j, k, \mb z | \bs\Theta)\right)\nonumber\\
\Rightarrow\max_{\bs\Theta} && \sum_{j = 1}^{n_v}\left(\sum_{k=1}^{m}P(k | \mb h_j, \bs\Theta^l)\int_{\mb z}P(\mb z|\mb h_j, \bs\Theta_k^l)\log P(\mb h_j| \mb z, \bs\Theta_k)\right) \nonumber\\
&&+ \sum_{j = 1}^{n_v}\left(\sum_{k=1}^{m}P(k | \mb h_j, \bs\Theta^l)\log \pi_k\right),
\label{eqn:EMFormula}
\end{eqnarray}
where the last line is obtained by factoring out $P(\mb h_j, k, \mb z | \bs\Theta) = \pi_kP(\mb h_j|\mb z,\bs\Theta_k)P(\mb z)$ and dropping $P(\mb z)$ which is irrelevant to the optimization parameters. Note that $E_{p(k,\mb z|\mb h_j,\bs\Theta^l)}$ (abbreviated as $E(k,\mb z|\mb h_j)$ when there is no ambiguity) means expectation with the conditional probability $P(k,\mb z|\mb h_j,\bs\Theta^l) = P(k | \mb h_j, \bs\Theta^l)P(\mb z|\mb h_j, \bs\Theta_k^l)$, where
\begin{equation}
P(k | \mb h_j, \bs\Theta^l) = \frac{\pi_k^lP(\mb h_j | \bs\Theta_k^l)}{\sum_{k=1}^{m}\pi_k^lP(\mb h_j | \bs\Theta_k^l)},
\label{eqn:Responsibilities}
\end{equation}
is the posterior probability that $\mb h_j$ belongs to mixture $k$ given the parameter estimations $\bs\Theta^l$. It is called the responsibility in \cite{MclachlanModelling}.

Formula \eqref{eqn:EMFormula} is maximized with the Alternating Expectation–Conditional Maximization (AECM) Algorithm with two cycles \cite{MclachlanModelling}, each cycle contains one E-step and one CM-Step. The first cycle is the same with \cite{MclachlanModelling}, the second cycle has taken into consideration the rotational constraints and derived the corresponding optimal closed form solutions. In the following we note $\gamma_{kj}^l=P(k|\mb h_j,\bs\Theta^l)$ for simplicity.

Alternating Expectation–Conditional Maximization:
\begin{enumerate}
\item Initialization: $l=0$\\
Initial estimates of the parameters $\bs{\Theta}^0 = \{ \mb A_k^0, \mb b_k^0, \boldsymbol{\Phi}_k^0, \pi_k^0, k = 1,2,...,m\}$ as in \cite{MclachlanModelling}.
\item Cycle 1:\\
E Step: compute the responsibilities $\{\gamma_{kj}^{l+1}\}$ by \eqref{eqn:Responsibilities} using the current estimations $\boldsymbol{\Theta}^l$.\\
CM Step: compute $\{ \pi_k^{l+1} \}$ and $\{ \mb b_k^{l+1} \}$ using the current responsibilities $\{\gamma_{kj}^{l+1}\}$, and the parameters $\{\mb A_k^l\}$ and $\{\bs{\Phi}_k^l\}$:
\begin{eqnarray}
\pi_k^{l+1} &=& \sum_{j = 1}^{n_v}\gamma_{kj}^{l+1}/n_v,~k=1,...,m.
\label{eqn:tildepi}\\
\mb b_k^{l+1} &=& \frac{\sum_{j = 1}^{n_v}\gamma_{kj}^{l+1}\mb h_j}{\sum_{j = 1}^{n_v}\gamma_{kj}^{l+1}},~k=1,...,m.
\label{eqn:tildeb}
\end{eqnarray}
\item Cycle 2:\\
E Step: update the responsibilities $\{\gamma_{kj}^{l+1}\}$ using the parameters $\{\mb b_k^{l+1}\},\{\pi_k^{l+1}\}$ and $\{\mb A_k^l\}, \{\bs{\Phi}_k^l\}$.\\
CM Step 2: compute $\{ \mb A_k^{l+1} \}$ and $\{ \boldsymbol{\Phi}_k^{l+1} \}$ using the current responsibilities and $\{\mb b_k^{l+1}\},\{\pi_k^{l+1}\}$:
\begin{eqnarray}
\mb A_k^{l+1} &=& \mb R_k^{l+1}\bs\Lambda_k^{l+1},~k=1,...,m.
\label{eqn:tildeA}\\
\bs{\Phi}_k^{l+1} &=& \text{diag}\sum_{j=1}^{n_v}\left(\frac{\gamma_{kj}^{l+1}}{\pi_kn_v}(\mb h_j-\mb b_k^{l+1})(\mb h_j-\mb b_k^{l+1})^T\nonumber\right.\\
&&-2\frac{\gamma_{kj}^{l+1}}{\pi_kn_v}(\mb h_j-\mb b_k^{l+1})E_{p(\mb z|\mb h_j)}^k(\mb z)^T{\mb A_k^{l+1}}^T\nonumber\\
&&\left.+\mb A_k^{l+1}\frac{\gamma_{kj}^{l+1}}{\pi_kn_v}E_{p(\mb z|\mb h_j)}(\mb z \mb z^T){\mb A_k^{l+1}}^T\right).\label{eqn:tildePhi}
\end{eqnarray}
The covariance matrix of the $k$th mixture (rigid part) of the $i$th shape in the $l+1$th iteration is:
\begin{equation}
\mb C_{ki}^{l+1}=\sum_{j=1}^{n_v}\frac{\gamma_{kj}^{l+1}}{\pi_k n_v}(\mb v_j^i - \mb b_k^{i,l+1})(\mb v_j^i - \mb b_k^{i,l+1})^T.
\label{eqn:mixtureCovariance}
\end{equation}
Assume $\bs\Sigma_{ki}^{l+1}$ is the diagonal matrix of the eigenvalues of $\mb C_{ki}^{l+1}$, then $\bs\Lambda_k^{l+1}$ is estimated as:
\begin{equation}
\bs\Lambda_k^{l+1} = \sqrt{\frac{1}{n_s}\sum_{i=1}^{n_s}\bs\Sigma_{ki}^{l+1}}.
\label{eqn:optLambda}
\end{equation}
Assume $\mb B_{ki}^{l+1} = \bs\Lambda_k^{l+1}\sum_{j=1}^{n_v}\gamma_{kj}E_{p(\mb z|\mb h_j,\Theta_k)}(\mb z)(\mb v_j^i-\mb b_k^{i,l+1})^T$, and $\mb U_{ki}^{l+1}\mb D_{ki}^{l+1}{\mb V_{ki}^{l+1}}^T$ is the singular value decomposition of $\mb B_{ki}^{l+1}$, we have:
\begin{equation}
\mb R_{k}^{i,l+1} = \mb V_{ki}^{l+1}{\mb U_{ki}^{l+1}}^T,~i=1,...,n_s.
\label{eqn:optRot}
\end{equation}
For the derivations of $\bs{\Phi}_k^{l+1}$, $\bs\Lambda_k^{l+1}$, and $\mb R_{k}^{i,l+1}$ and the way to ensure the right handedness of $\mb R_{k}^{i,l+1}$ please refer to the section of Appendix.
\item Repeat (ii) and (iii) until converge.
\end{enumerate}

\section{Hierarchical Optimization Algorithm}

As noted in \cite{tang2012deep} when the number of mixtures is large, MFA could converge to bad local minimums. In order to avoid bad local minimums we designed a hierarchical optimization algorithm. In the following we use the parameter set $\bs\Theta_k$ to denote the $k$th factor analyzer.

\begin{algorithm}
\caption{Hierarchical Optimization Algorithm}
\label{alg:HierarchOpt}
\begin{itemize}
	\item Inputs:\\
	The training shapes $\mb S_1,...,\mb S_{n_s}$.\\
	The initial number of mixtures $m$.
	\item $\{\bs\Theta_1,...,\bs\Theta_{m}\}$ = AECM($\{\mb S_i\}$, m).
	\item For each factor analyzer $\bs\Theta_k,k\in\{1,2,...,m\}$:
	\begin{enumerate}
		\item Initialize: $m_k = 1$.
		\item $\{\mb S_i^k\}$ = the rigid parts correspond to $\bs\Theta_k$ of $\{\mb S_i\}$.
		\item $\{\bs\Theta_{k1},...,\bs\Theta_{km}\}$ = AECM($\{\mb S_i^k\}$, $m_k$).
		\item $\text{err} = \sum_{j=1}^{m_k}\text{trace}(\bs\Phi_{kj})/(3n_sm_k)$.
		\item Break if $\text{err} < 1.0e-5$ or err cease to decrease as $m_k$ increases.
		\item $m_k=m_k + 1$, go to step 3.
	\end{enumerate}
	\item $\{I_j\}\Leftarrow\{\bs\Theta_{kj},k=1,..,m,j=1,...,m_k\}$.
	\item Outputs: $\{\bs\Theta_1,...,\bs\Theta_{N}\}$ = AECM($\{\mb S_i\}$, $\{I_j\}$).
\end{itemize}
\end{algorithm}

As shown in Algorithm \ref{alg:HierarchOpt}, The inputs are the training shapes $\mb S_1,...,\mb S_{n_s}$ and the initial number of mixtures $m$, which is usually small. In the upper level the coarse factor analyzers $\{\bs\Theta_1,...,\bs\Theta_{m}\}$ are obtained by the AECM algorithm AECM($\{\mb S_i\}$, m) with the input data $\{\mb S_i\}$ and the number of mixtures $m$. Then for each factor analyzer $\bs\Theta_k,k\in\{1,2,...,m\}$ (each coarse rigid part), refinement is conducted until the average variances is less that $1.0e-5$ or the variances cease to decrease as $m_k$ increases. Note that $\bs\Phi_k$ is the average squared error between the real shapes and corresponding images in the latent space. In this study, all the shapes are scaled to be within the unit box. After the refinements, the vertices labels $\{I_j,j=1,...,n_v\}$ are obtained from the refined factor analyzers $\Leftarrow\{\bs\Theta_{kj},k=1,..,m,j=1,...,m_k\}$ and are used to initialize the final optimization AECM($\{\mb S_i\}$, $\{I_j\}$).

Given the factor analyzers $\bs\Theta=\{\bs\Theta_1,...,\bs\Theta_{N}\}$, each vertex label is obtained by associating the vertex to the mixture that has the maximum posterior probability:
\begin{eqnarray}
I_j = \max_{k=1,...,m} P(\mb k|\mb h_j,\bs\Theta),~j=1,...,n_v.
\label{eqn:AssignLabels}
\end{eqnarray}

\begin{figure}[ht]
	\centering
	\subfigure[]{
		\includegraphics[width=0.28\textwidth,trim=1cm 4cm 1cm 5cm, clip]{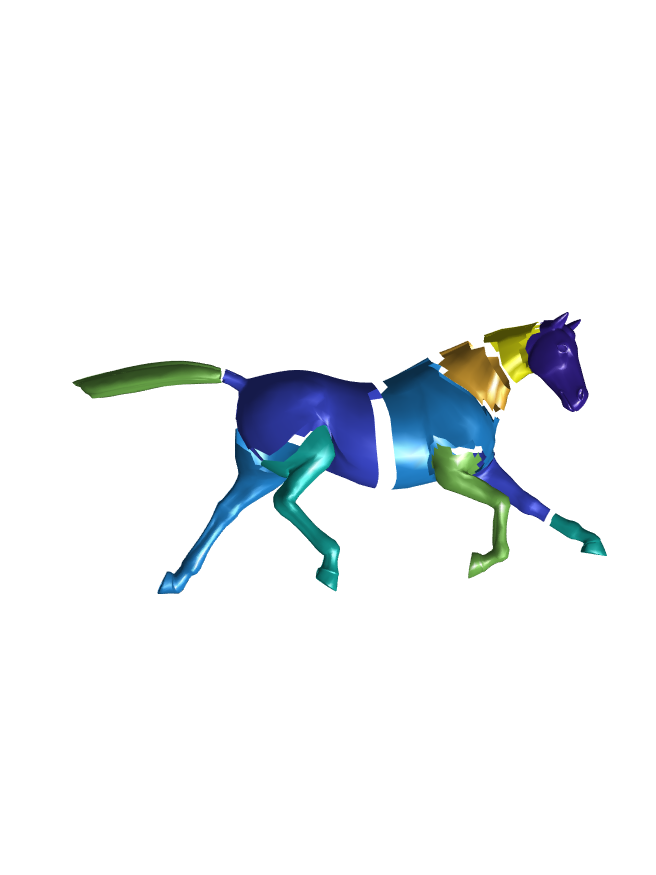}
		\label{fig:hierarchicalOptimizeA}
	}
	\subfigure[]{
		\includegraphics[width=0.28\textwidth,trim=1cm 4cm 1cm 5cm, clip]{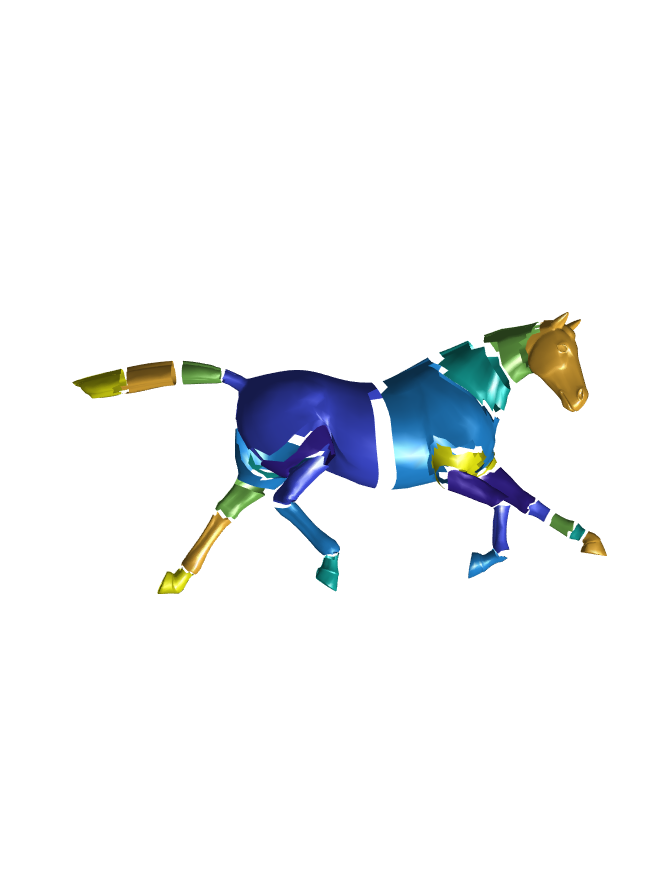}
		\label{fig:hierarchicalOptimizeB}
	}
	\subfigure[]{
		\includegraphics[width=0.28\textwidth,trim=1cm 4cm 1cm 5cm, clip]{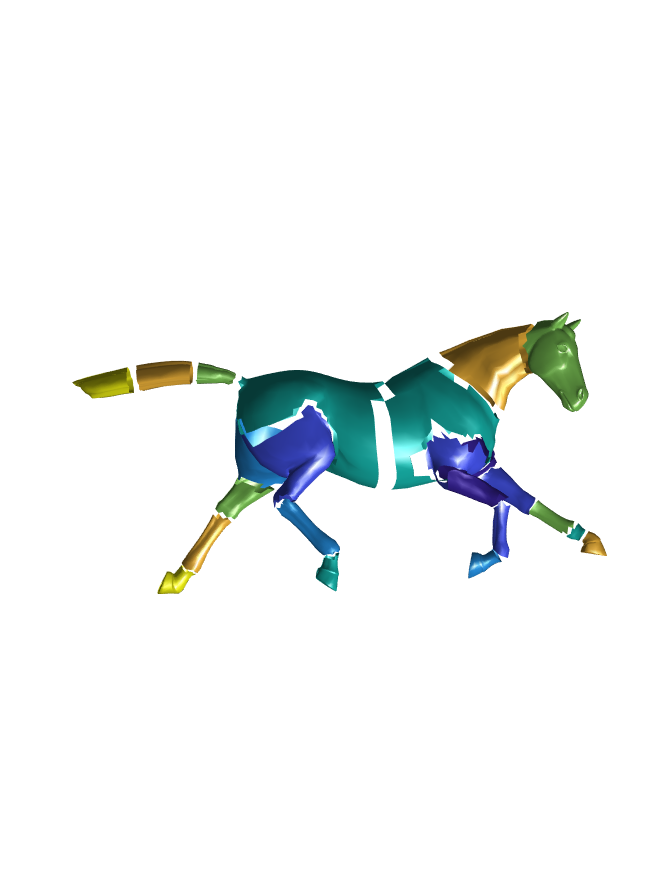}
		\label{fig:hierarchicalOptimizeC}
	}
	\caption{Hierarchical optimization: a) coarse factor analyzers with 11 mixtures; b) refinements that obtain 23 mixtures; c) final results with 23 mixtures.}
	\label{fig:hierarchicalOptimize}
\end{figure}

As shown in Figure \ref{fig:hierarchicalOptimize} are the results of hierarchical optimization on the horse training data. Figure \ref{fig:hierarchicalOptimizeA} shows the results of coarse factor analyzers, in which the horse is segmented into 11 rigid parts. The gaps between the different rigid parts in the figures are obtained by not showing the triangles on the common boundaries. Figure \ref{fig:hierarchicalOptimizeB} shows the results of refinements. Figure \ref{fig:hierarchicalOptimizeC} shows the 23 mixtures (rigid parts) obtained in the final stage, from which we could see that all the joints on the legs are separated, the head, neck, and body are also separated since the poses of training shapes include movements of the head and neck. The tail is also segmented into 3 parts, which swings in the running.

The images of the vertices in the latent space are obtained by:
\begin{eqnarray}
\mb v_{j}^{ref} = \bs\Lambda_{I_j} E_{p(\mb z|\mb h_j,\bs\Theta_{I_j})}(\mb z),~j=1,...,n_v.
\label{eqn:VerticesImages}
\end{eqnarray}

\begin{figure}[ht]
	\centering
	\subfigure{
		\includegraphics[width=0.16\textwidth,trim=2cm 0.5cm 2cm 2cm, clip]{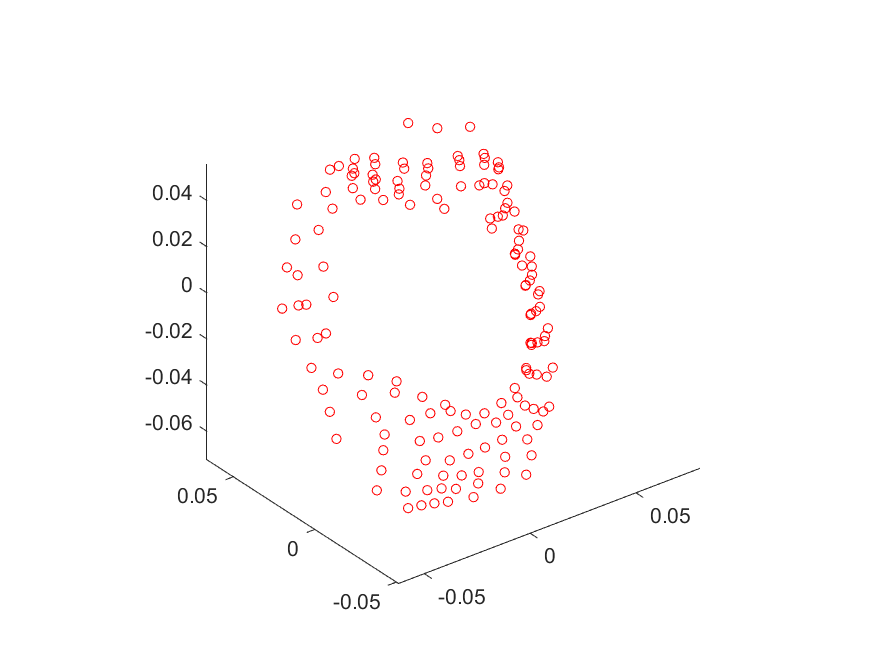}
	}
	\subfigure{
		\includegraphics[width=0.16\textwidth,trim=2cm 0.5cm 2cm 2cm, clip]{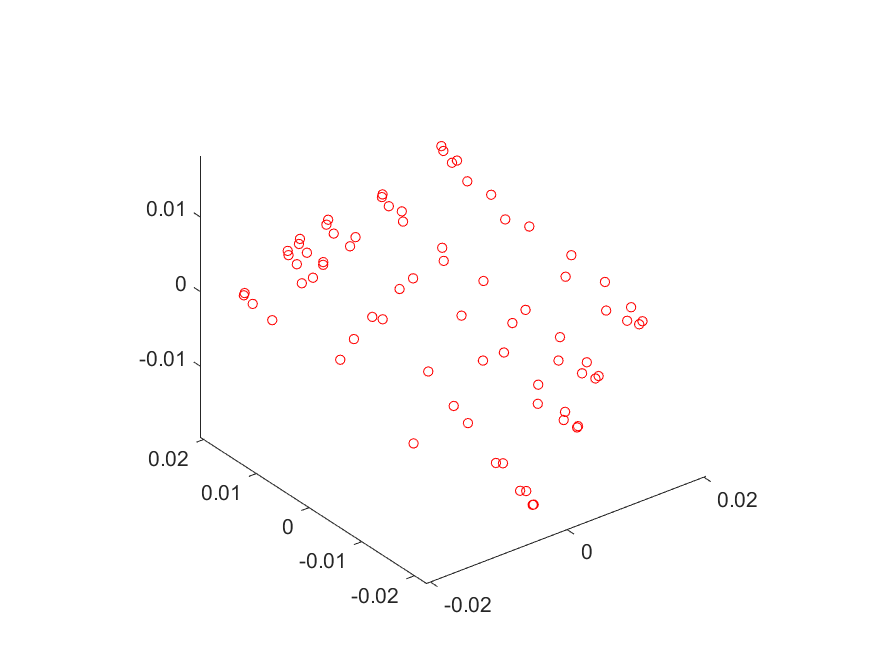}
	}
	\subfigure{
		\includegraphics[width=0.16\textwidth,trim=2cm 0.5cm 2cm 2cm, clip]{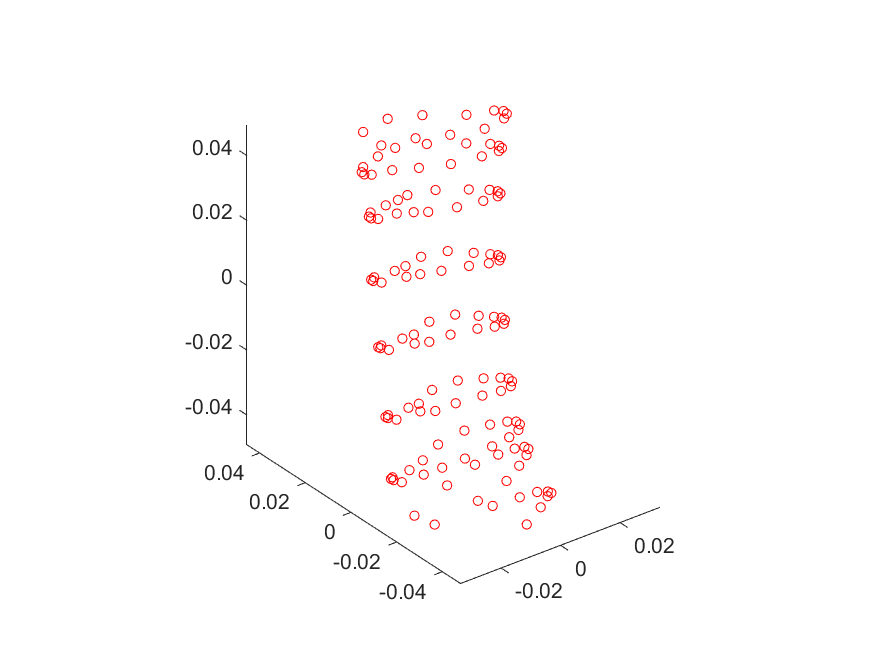}
	}
	\subfigure{
		\includegraphics[width=0.16\textwidth,trim=2cm 0.5cm 2cm 2cm, clip]{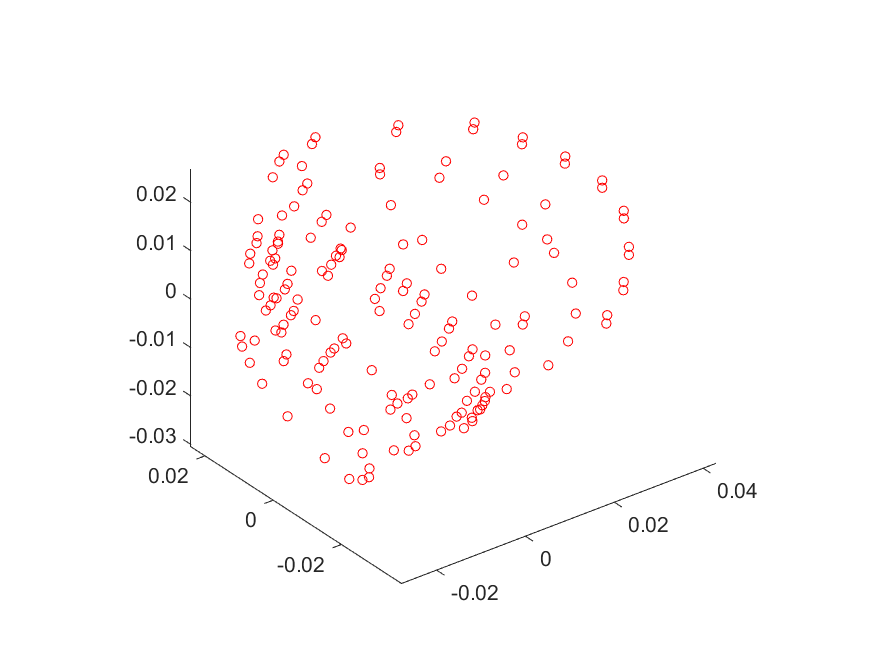}
	}
	\subfigure{
		\includegraphics[width=0.16\textwidth,trim=2cm 0.5cm 2cm 2cm, clip]{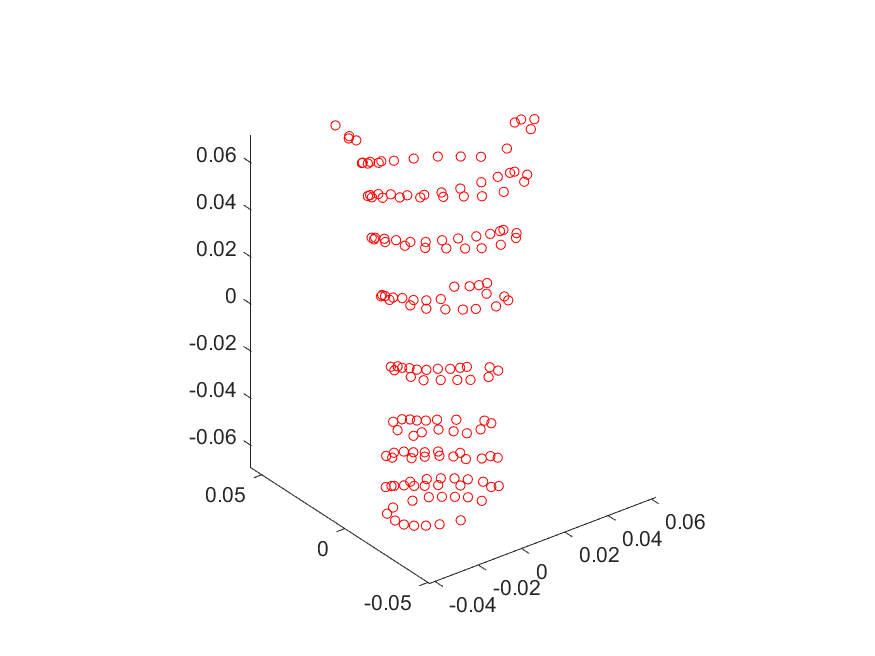}
	}
	\subfigure{
		\includegraphics[width=0.16\textwidth,trim=2cm 0.5cm 2cm 2cm, clip]{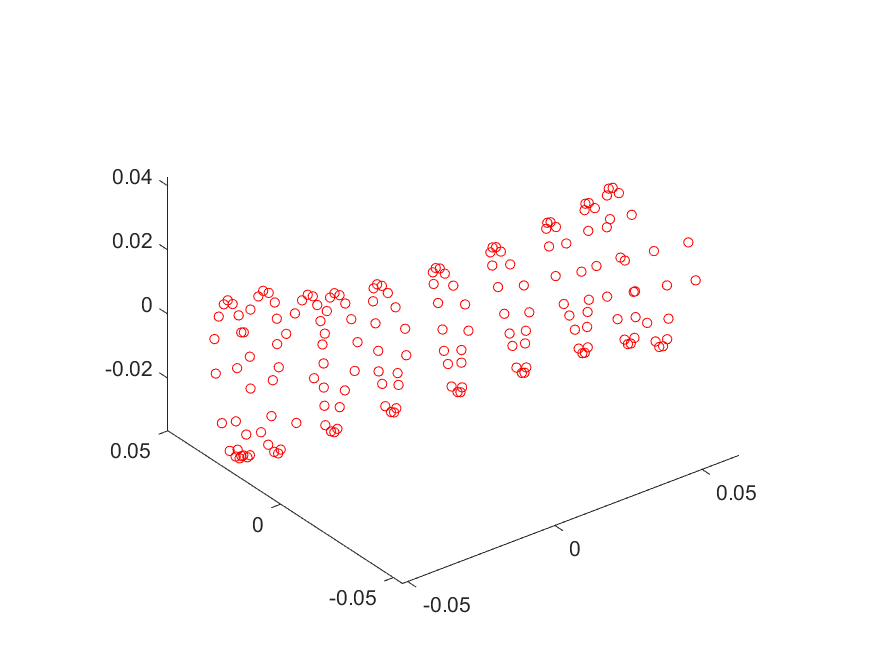}
	}
	\subfigure{
		\includegraphics[width=0.16\textwidth,trim=2cm 0.5cm 2cm 2cm, clip]{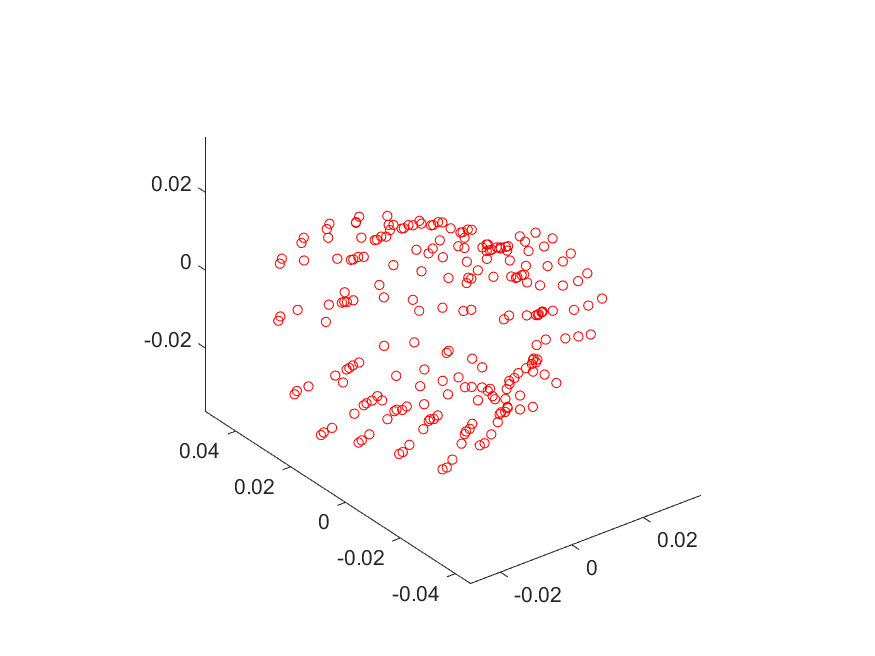}
	}
	\subfigure{
		\includegraphics[width=0.16\textwidth,trim=2cm 0.5cm 2cm 2cm, clip]{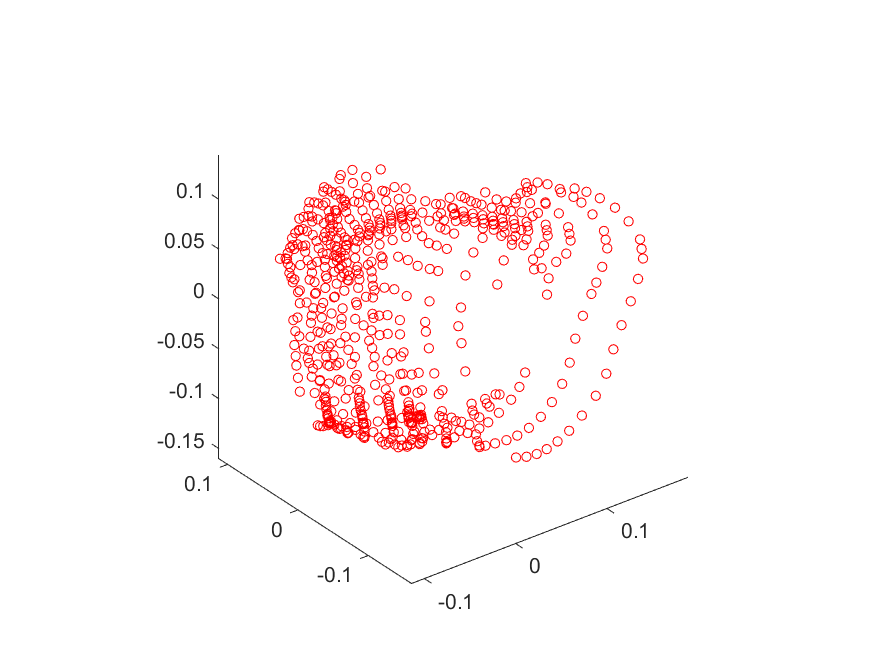}
	}
	\subfigure{
		\includegraphics[width=0.16\textwidth,trim=2cm 0.5cm 2cm 2cm, clip]{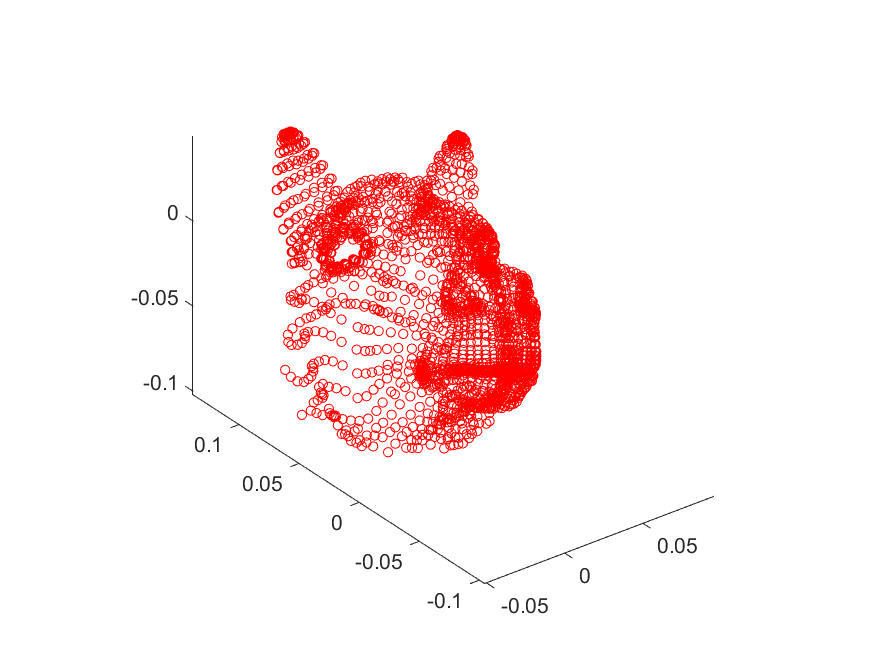}
	}
	\subfigure{
		\includegraphics[width=0.16\textwidth,trim=2cm 0.5cm 2cm 2cm, clip]{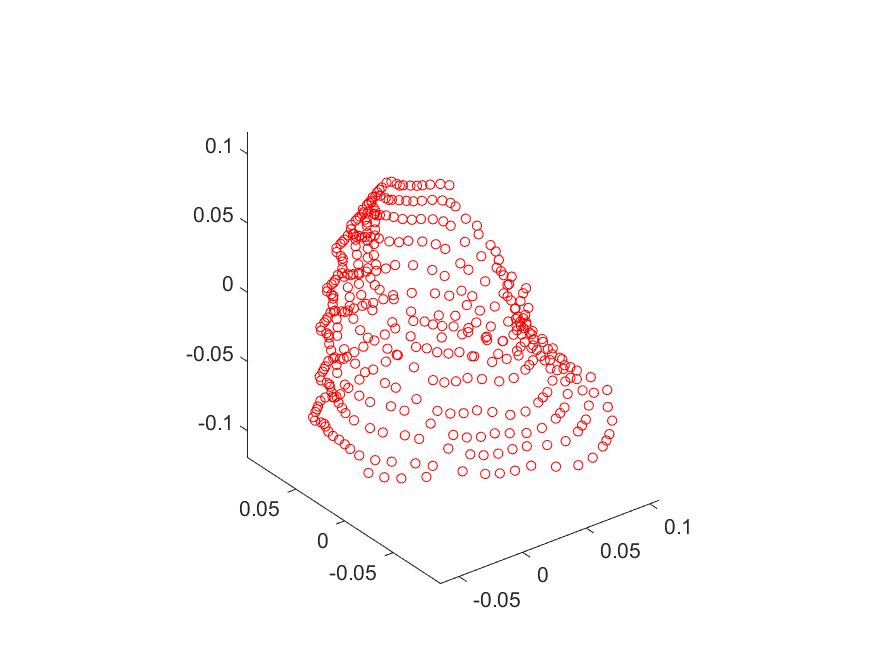}
	}
	\subfigure{
		\includegraphics[width=0.16\textwidth,trim=2cm 0.5cm 2cm 2cm, clip]{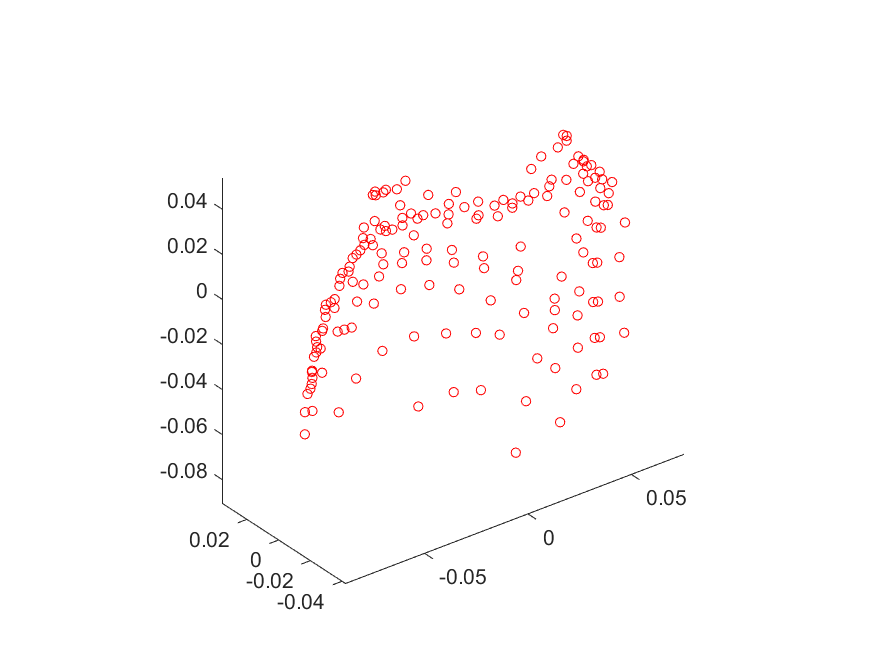}
	}
	\subfigure{
		\includegraphics[width=0.16\textwidth,trim=2cm 0.5cm 2cm 2cm, clip]{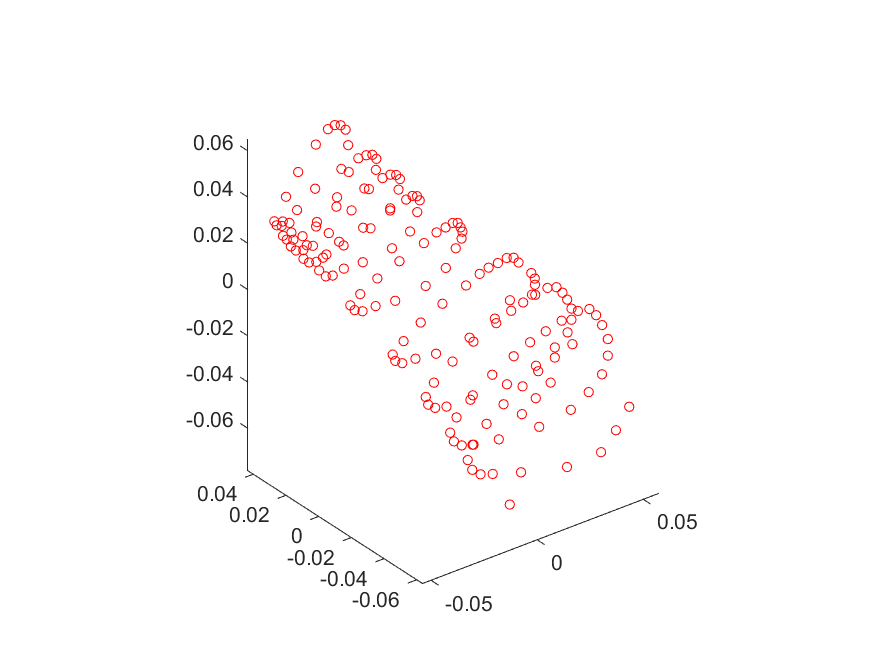}
	}
	\subfigure{
		\includegraphics[width=0.16\textwidth,trim=2cm 0.5cm 2cm 2cm, clip]{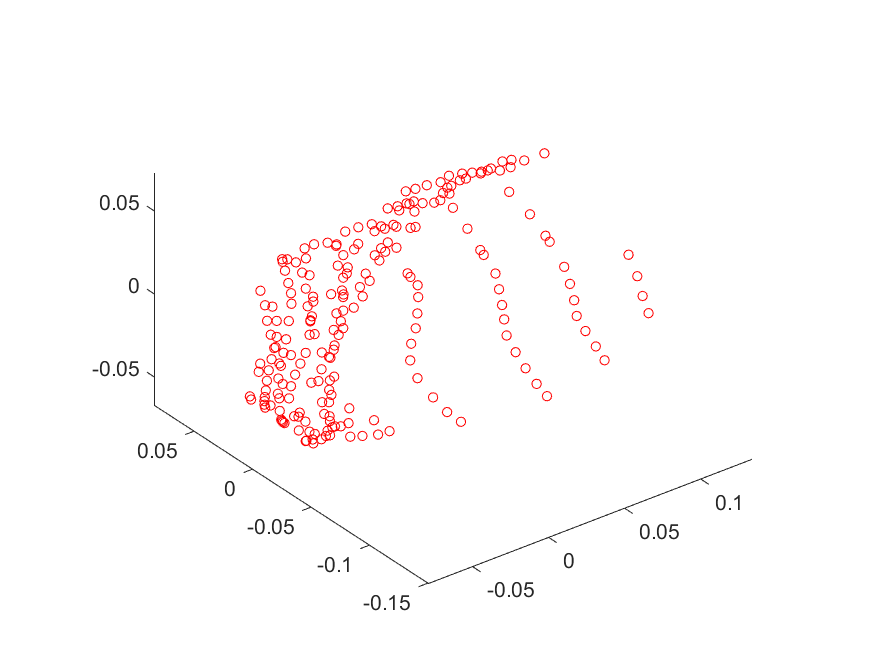}
	}
	\subfigure{
		\includegraphics[width=0.16\textwidth,trim=2cm 0.5cm 2cm 2cm, clip]{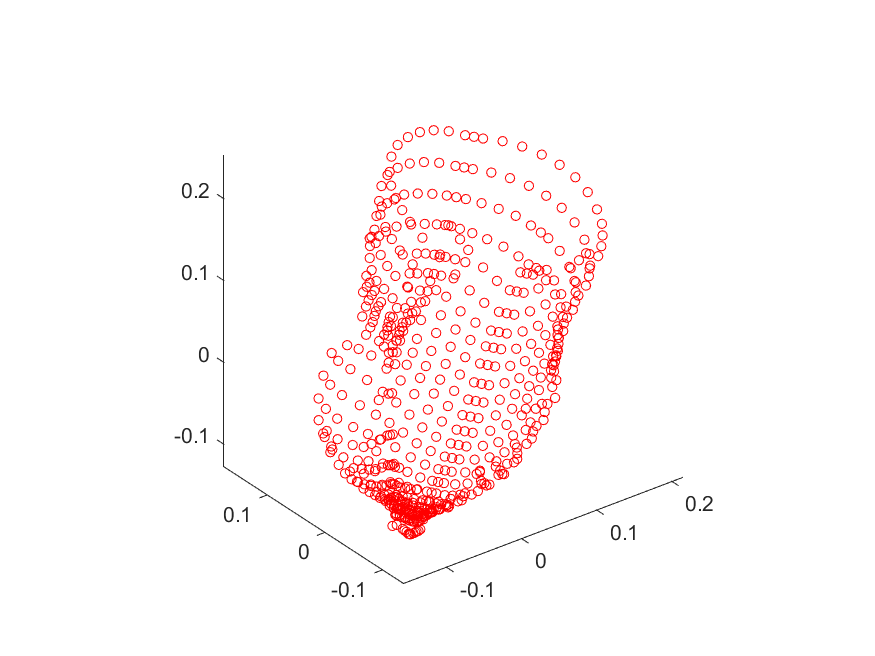}
	}
	\subfigure{
		\includegraphics[width=0.16\textwidth,trim=2cm 0.5cm 2cm 2cm, clip]{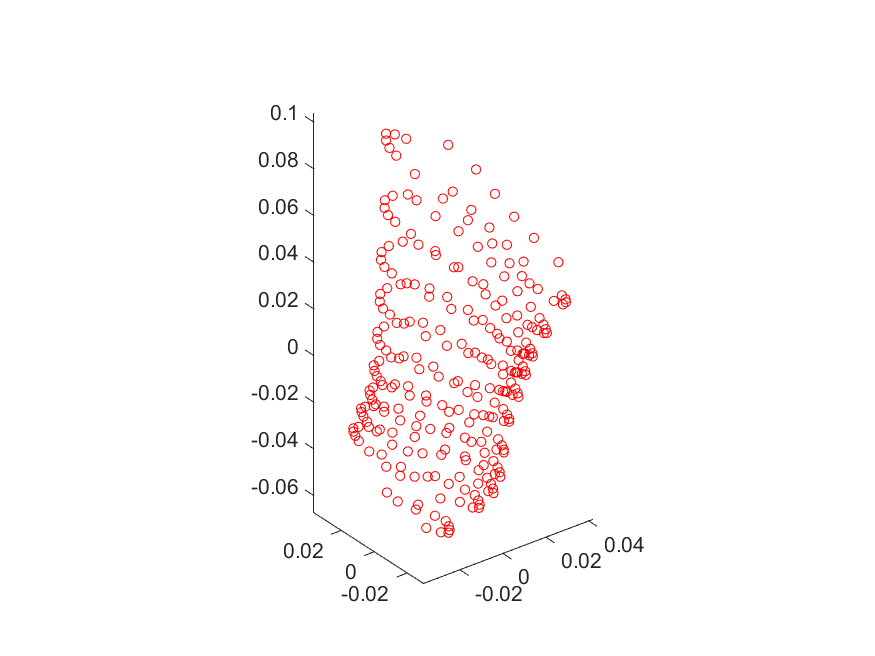}
	}
	\subfigure{
		\includegraphics[width=0.16\textwidth,trim=2cm 0.5cm 2cm 2cm, clip]{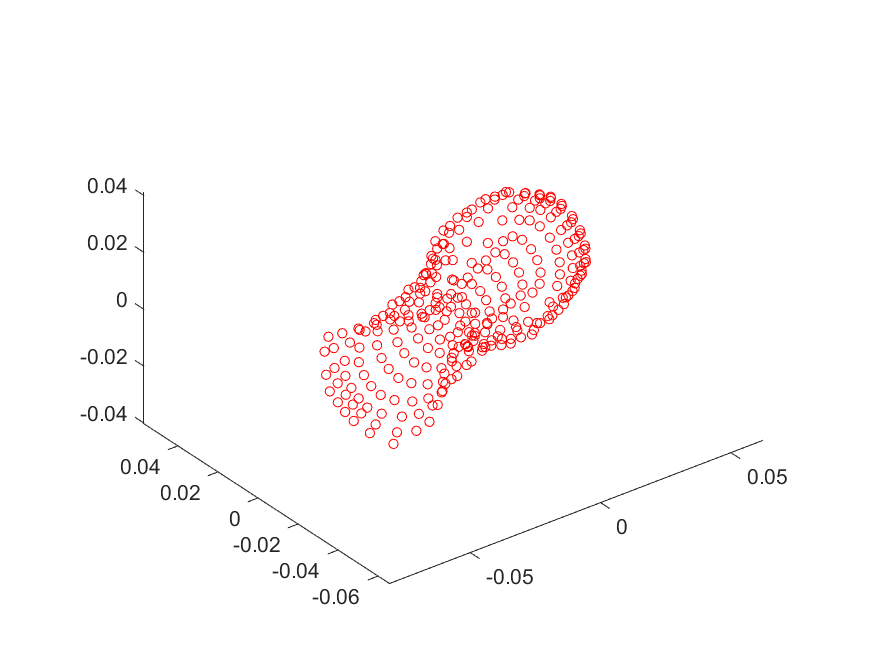}
	}
	\subfigure{
		\includegraphics[width=0.16\textwidth,trim=2cm 0.5cm 2cm 2cm, clip]{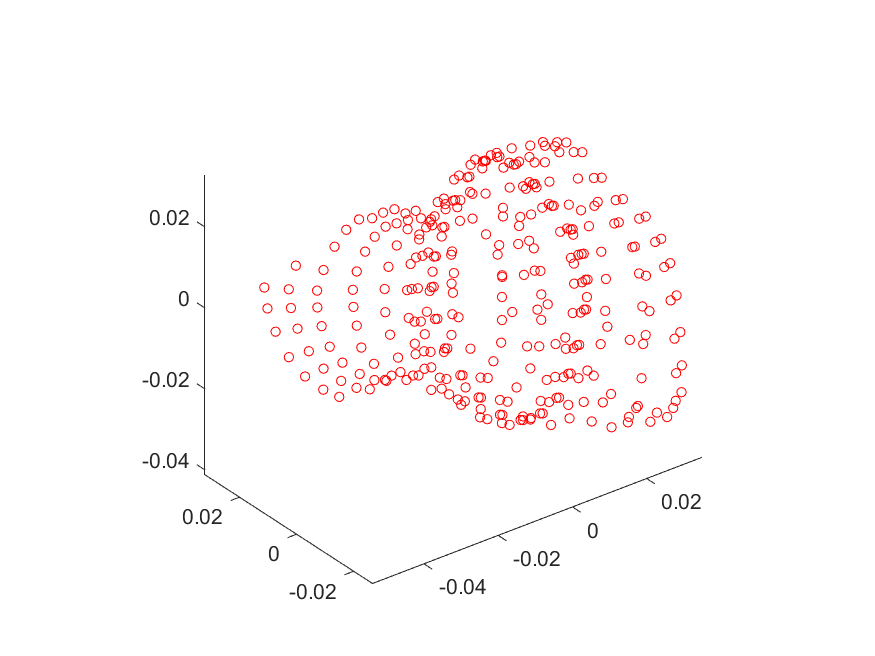}
	}
	\subfigure{
		\includegraphics[width=0.16\textwidth,trim=2cm 0.5cm 2cm 2cm, clip]{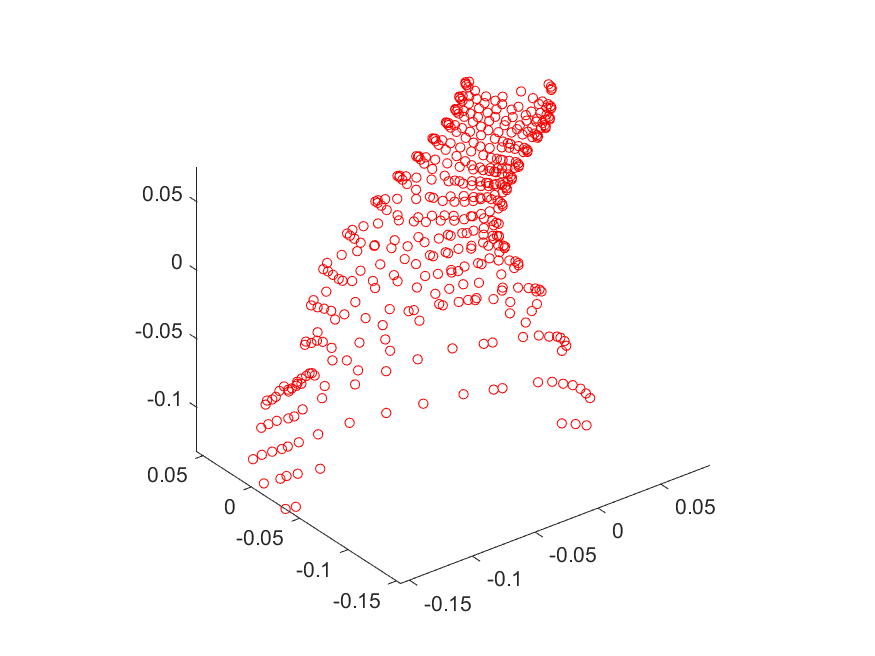}
	}
	\subfigure{
		\includegraphics[width=0.16\textwidth,trim=2cm 0.5cm 2cm 2cm, clip]{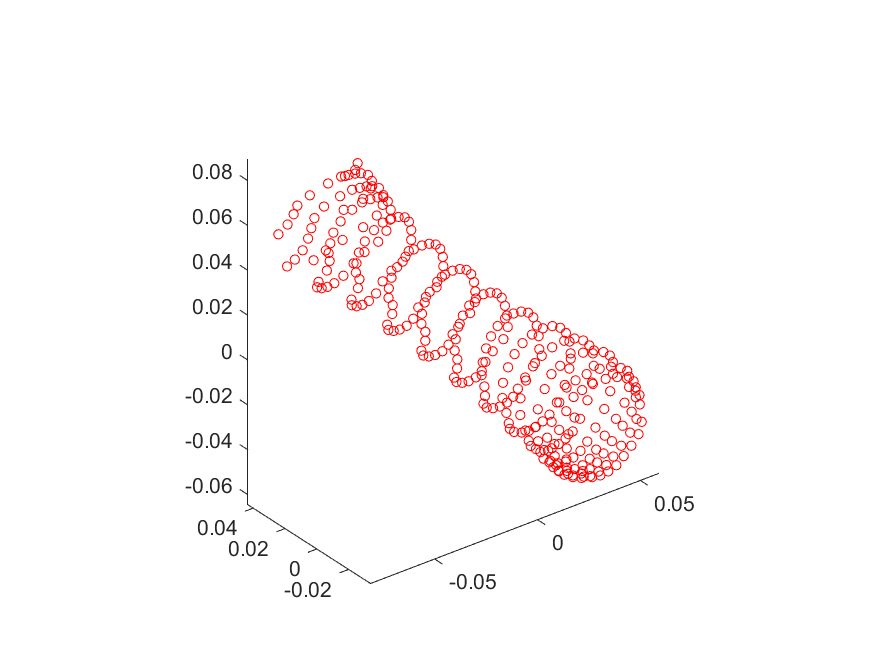}
	}
	\subfigure{
		\includegraphics[width=0.16\textwidth,trim=2cm 0.5cm 2cm 2cm, clip]{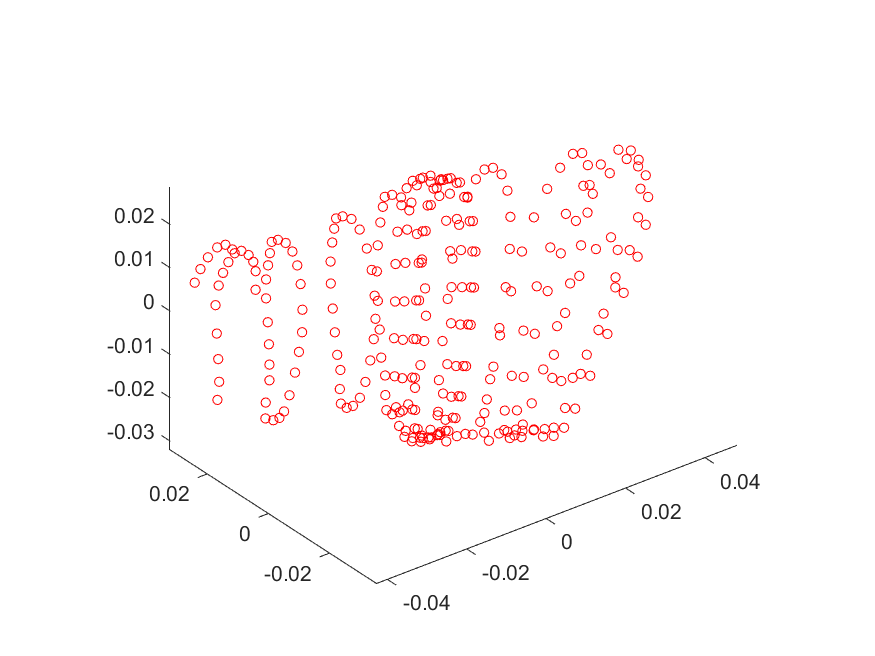}
	}
	\subfigure{
		\includegraphics[width=0.16\textwidth,trim=2cm 0.5cm 2cm 2cm, clip]{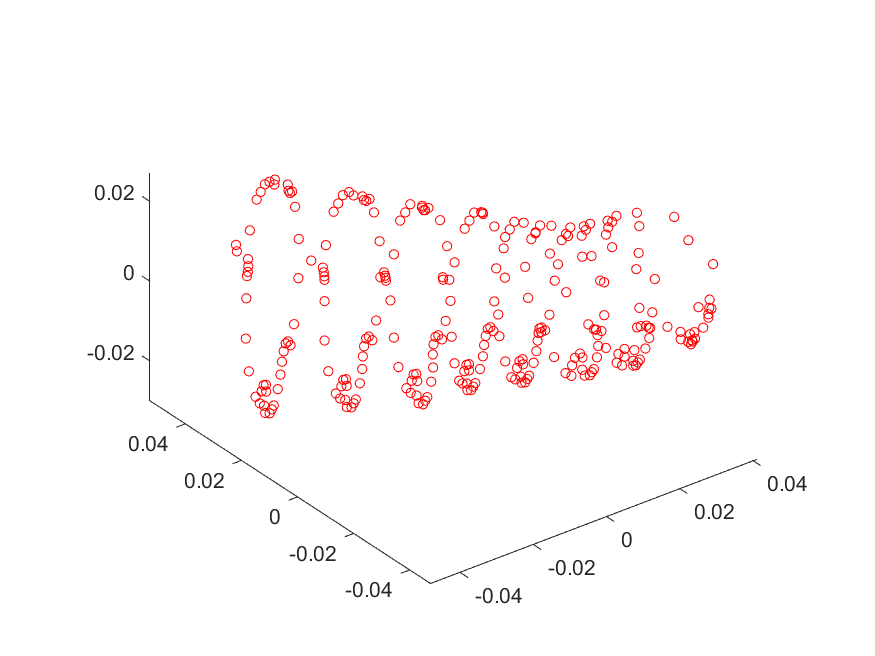}
	}
	\subfigure{
		\includegraphics[width=0.16\textwidth,trim=2cm 0.5cm 2cm 2cm, clip]{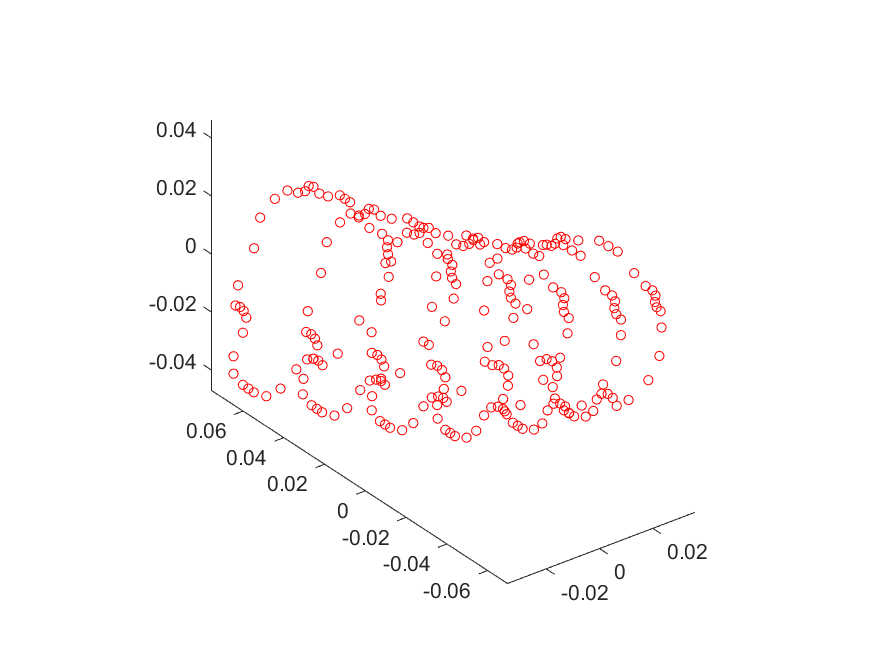}
	}
	\subfigure{
		\includegraphics[width=0.16\textwidth,trim=2cm 0.5cm 2cm 2cm, clip]{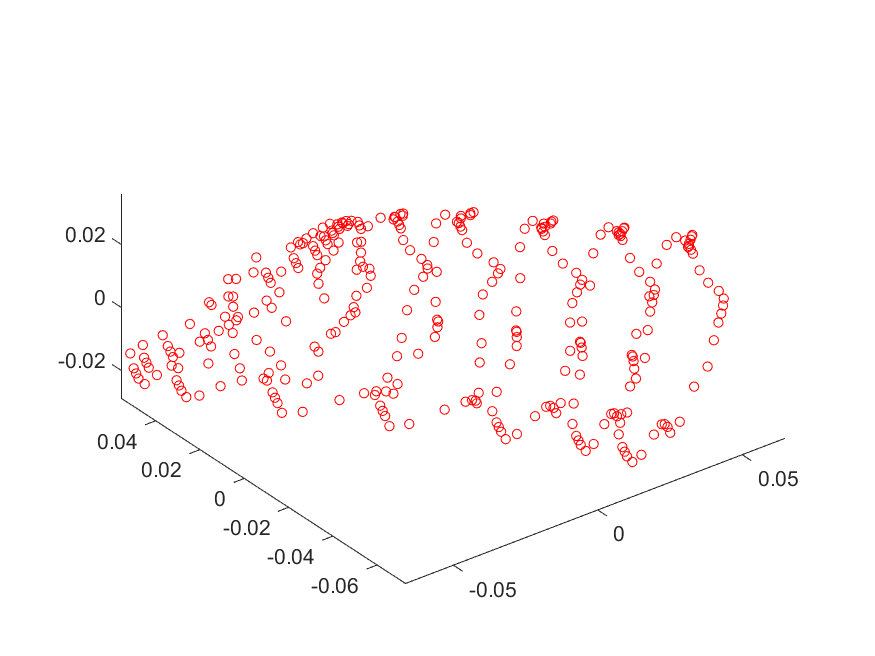}
	}
	\caption{The images of the rigid parts in the latent space of the horse training data.}
	\label{fig:horseLatentSpace}
\end{figure}

As shown in Figure \ref{fig:horseLatentSpace} are images of the rigid parts $\mb S_k^{ref}={\mb v_j^{ref},\forall I_j=k},k=1,...m$ in the latent space of the horse training data. From the images we can see that the horse is precisely segmented by the hierarchical optimization algorithm. From the rigid parts in the latent space the horse shapes can be reconstructed by the learned rotations and translations, for example, the vertices of the $i$th shape ($i\in[1,...,n_s]$) are reconstructed as:
\begin{eqnarray}
\mb R_{I_j}^i\mb v_{j}^{ref} + \mb b_{I_j}^i, j=1,...,n_v.
\label{eqn:SkeletonReconstruct}
\end{eqnarray}

\begin{figure}[ht]
	\centering
	\subfigure{
		\includegraphics[width=0.22\textwidth,trim=1cm 4cm 1cm 5cm, clip]{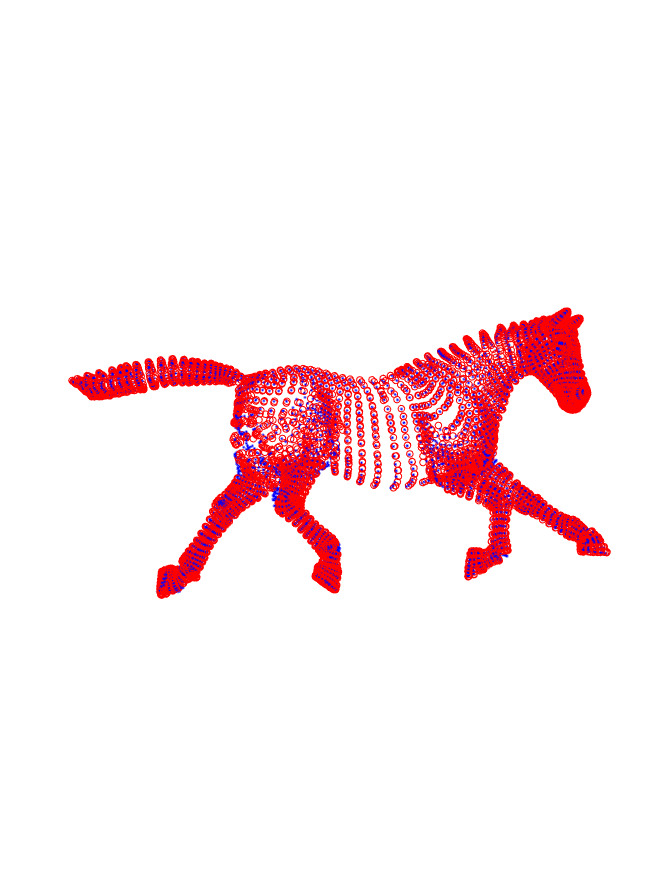}
		\label{fig:11horsesReconstructA}
	}
	\subfigure{
		\includegraphics[width=0.24\textwidth,trim=1cm 4.5cm 0.5cm 4.5cm, clip]{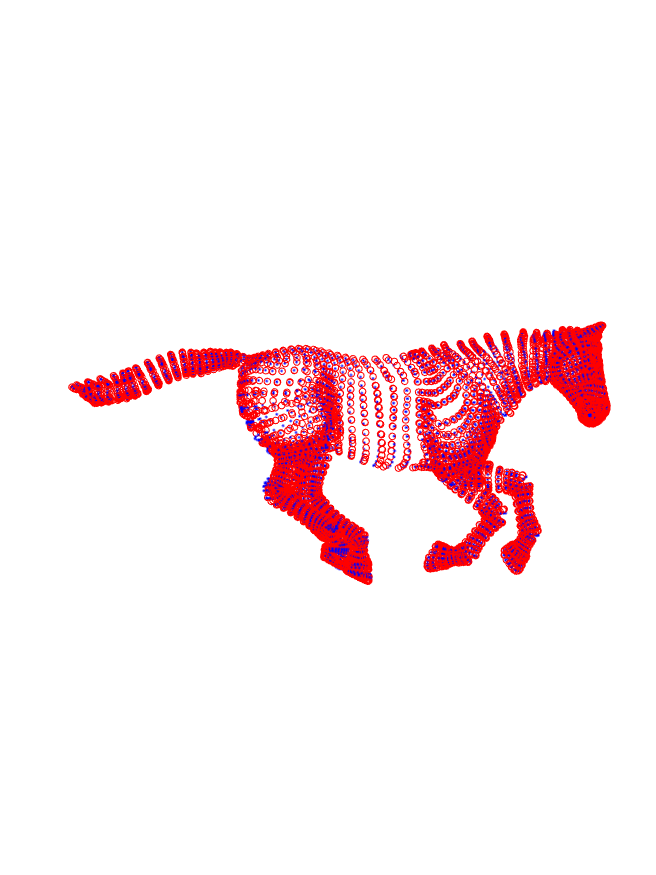}
	}
	\subfigure{
		\includegraphics[width=0.16\textwidth,trim=1cm 3cm 0.5cm 3cm, clip]{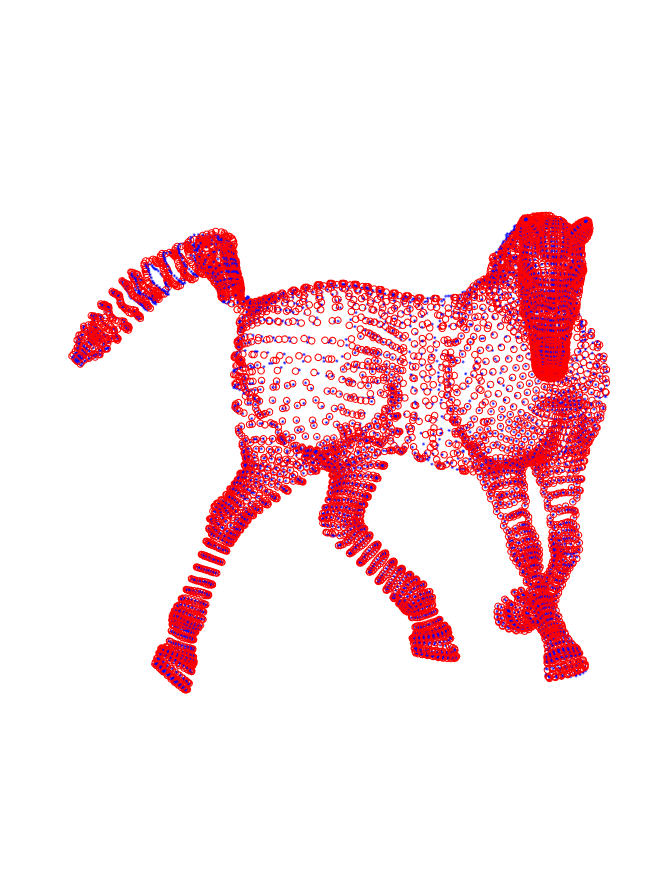}
	}
	\subfigure{
		\includegraphics[width=0.2\textwidth,trim=1cm 4cm 0.5cm 5cm, clip]{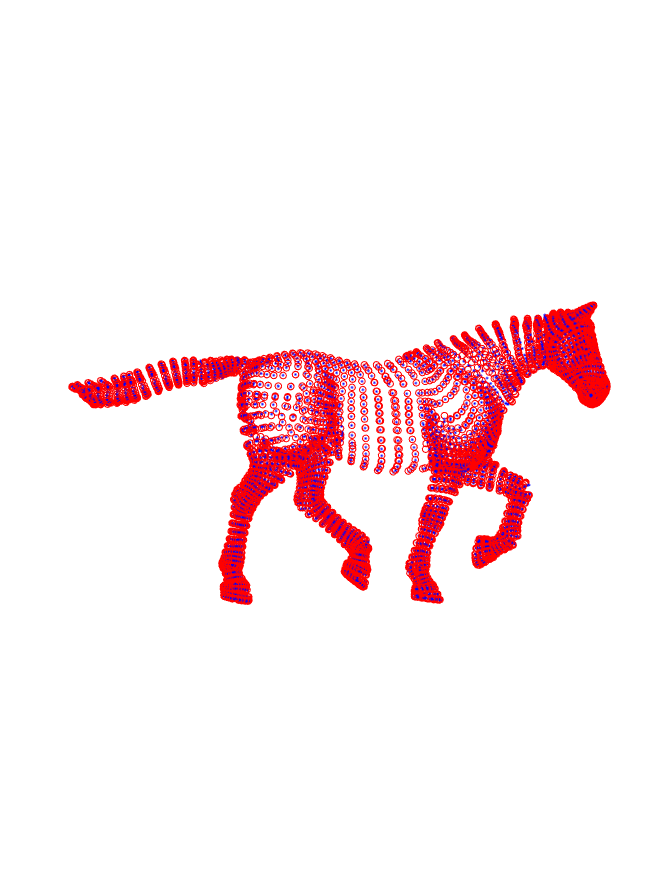}
	}
	\subfigure{
		\includegraphics[width=0.2\textwidth,trim=1cm 4cm 0.5cm 4.5cm, clip]{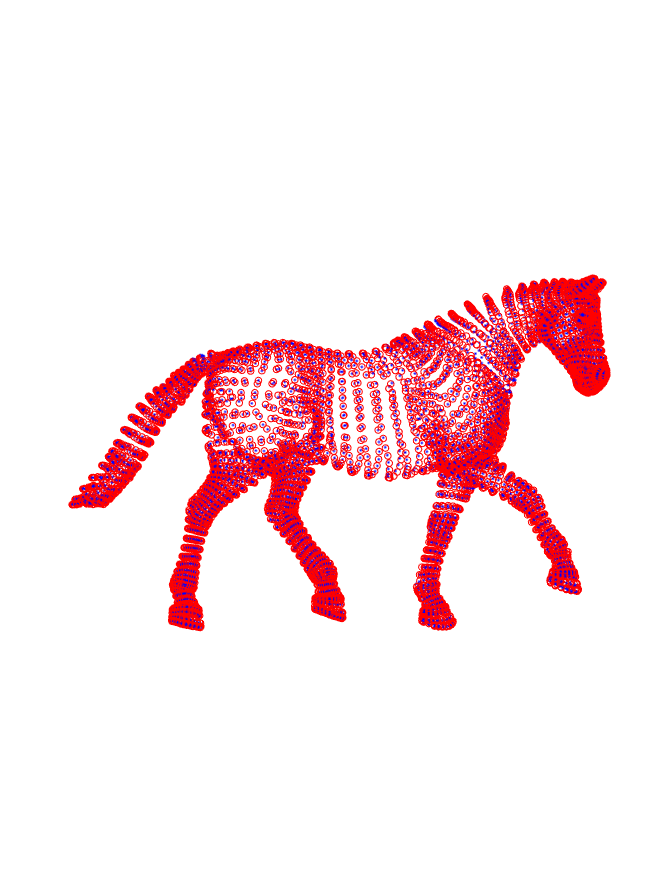}
	}
	\subfigure{
		\includegraphics[width=0.2\textwidth,trim=1cm 4cm 0.5cm 4.5cm, clip]{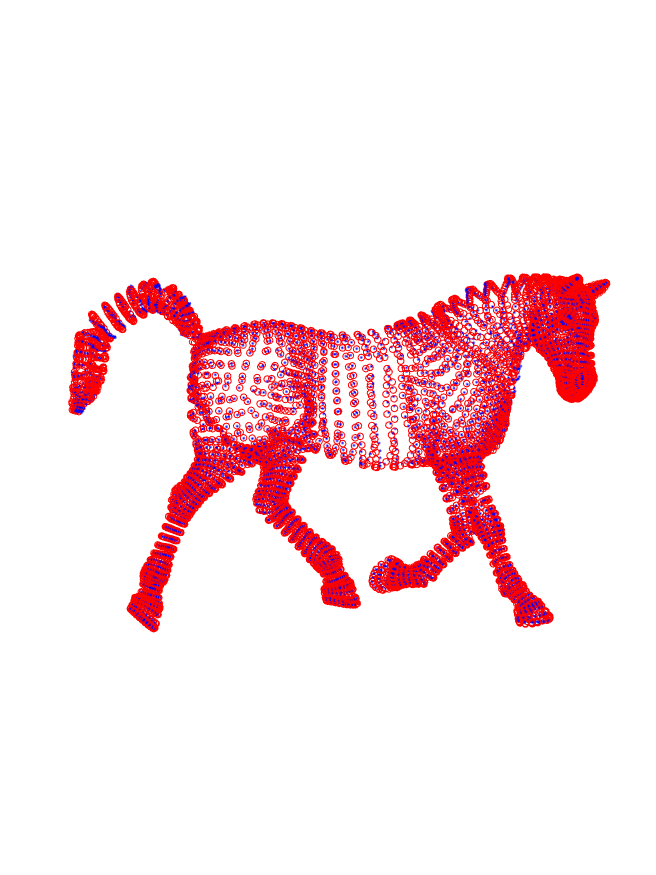}
	}
	\subfigure{
		\includegraphics[width=0.22\textwidth,trim=1cm 4.5cm 1cm 5cm, clip]{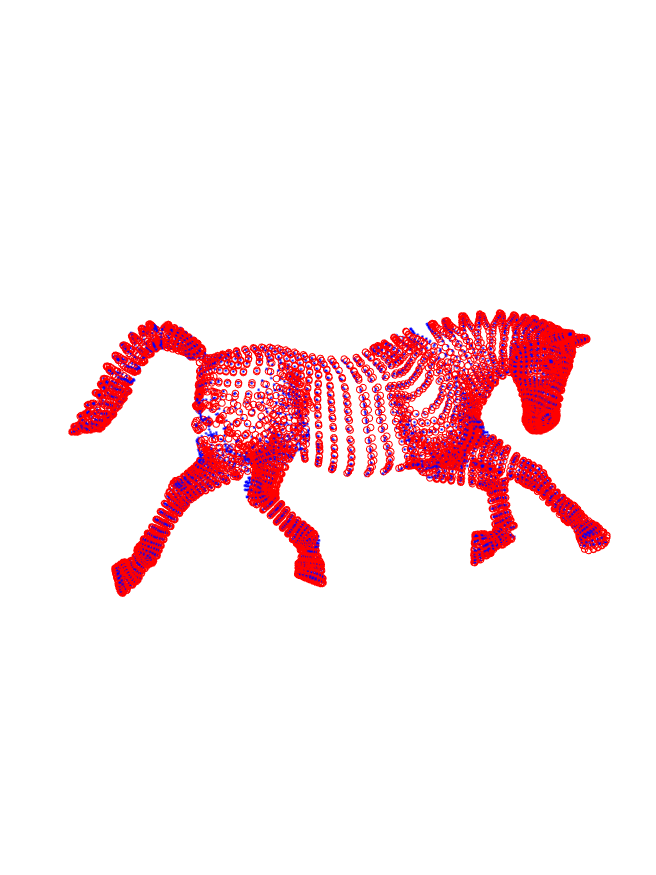}
	}
	\subfigure{
		\includegraphics[width=0.18\textwidth,trim=1cm 3cm 0.5cm 3.5cm, clip]{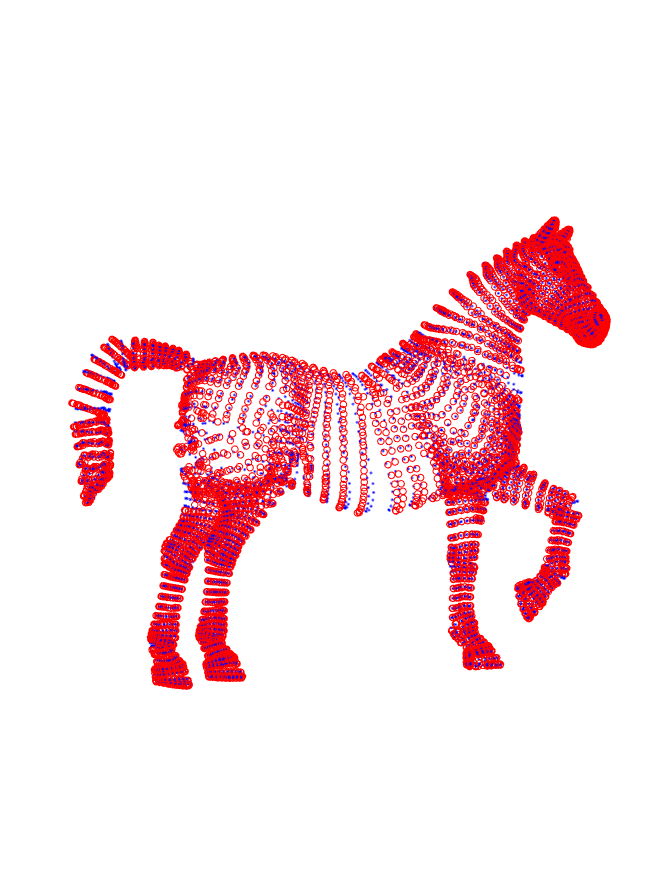}
	}
	\subfigure{
		\includegraphics[width=0.2\textwidth,trim=1cm 4cm 0.5cm 5cm, clip]{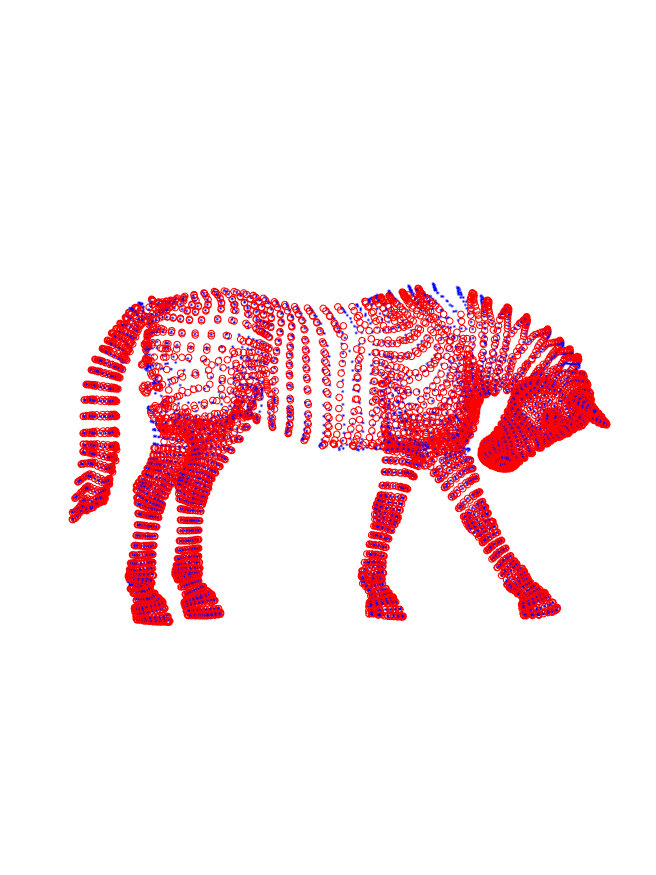}
	}
	\subfigure{
		\includegraphics[width=0.19\textwidth,trim=1cm 3.5cm 0.5cm 4cm, clip]{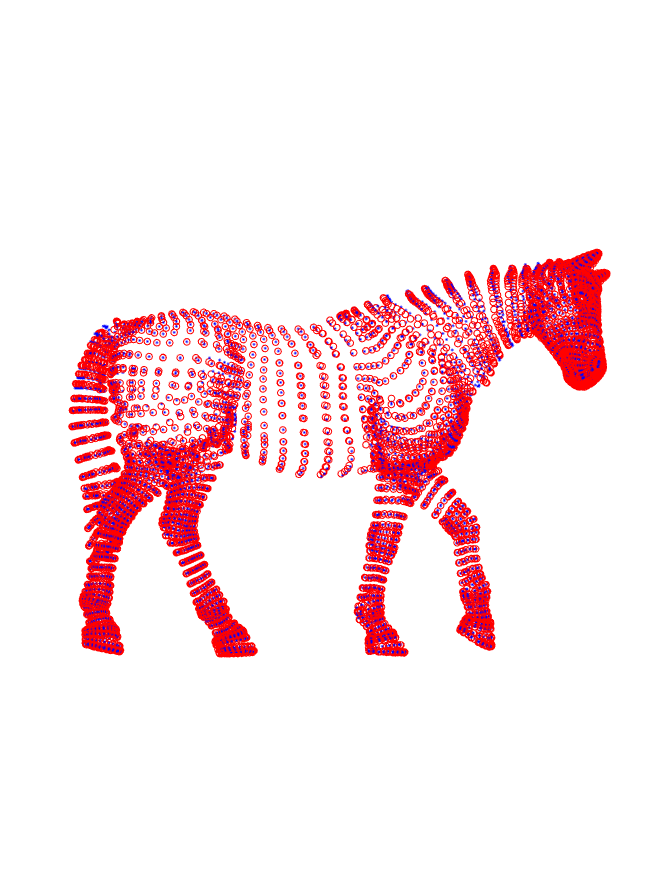}
	}
	\subfigure{
		\includegraphics[width=0.2\textwidth,trim=1cm 3cm 0.5cm 3cm, clip]{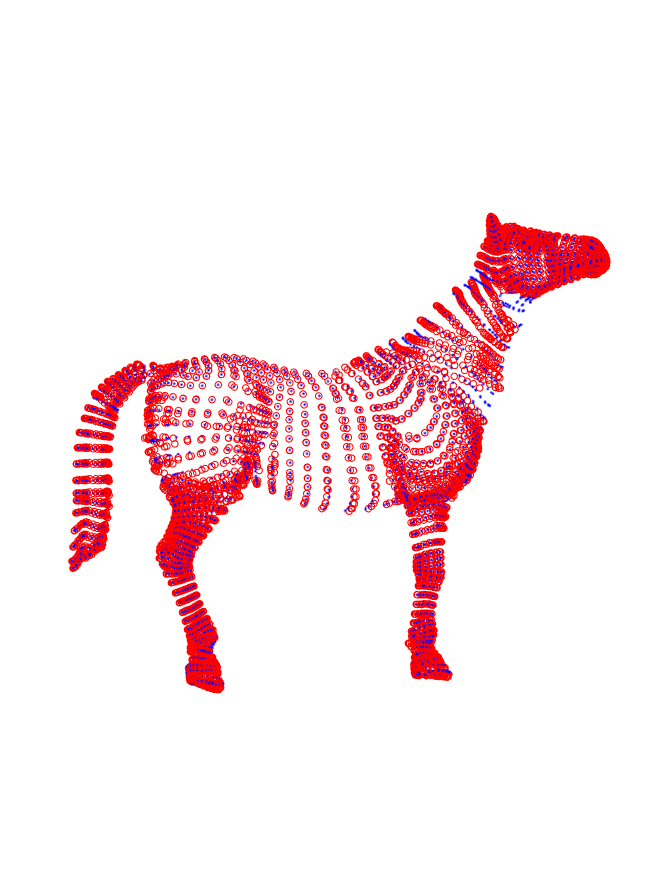}
	}
	\caption{The 11 horse shapes reconstructed from the latent space by the learned rotations and translations. Red: reconstructed shapes; Blue: original shapes.}
	\label{fig:11horsesReconstruct}
\end{figure}

As shown in Figure \ref{fig:11horsesReconstruct} are the 11 horse shapes reconstructed from the latent space by the learned rotations and translations. It can be seen that the learned rigid pose variations (rotations and translations) have captured the major variations in the shape population. The remaining small discrepancies between the reconstructed shape (red points) and original shape (blue points) are caused by the muscle movements corresponding to the different poses.

\section{Pose Interpolation}

Through the hierarchical AECM optimization, we have the mixtures of factor analyzers $\{\bs\Theta_k,k=1,...,N\}$, from which we have the vertices labels $I_j,j=1,...,n_v$ that indicate which rigid part each vertex belongs to, the images of the rigid parts in the latent space $\mb S_k^{ref},k=1,...m$, and the rotations and translations $\mb R_k^i,\mb b_k^i,k=1,...,m,i=1,..,n_s$ of the mixtures. Based on the above information, given any two shapes we can interpolate their poses.

\begin{algorithm}
\caption{Pose Interpolation}
\label{alg:Interpolate}
\begin{itemize}
\item Inputs: two shapes $\mb S_i$ and $\mb S_j$, parameter $t\in[0, 1]$.
\item $r = \min_{k=1,...,m}~\text{Angle}(\mb R_k^i,\mb R_k^j)$.
\item $visit.insert(r), queue.push(r), parent[r] = r$.
\item Until $queue$ is empty
\begin{enumerate}
\item $c = queue.pop()$.
\item $\mb R_c^t$ = Slerp($\mb R_c^i$, $\mb R_c^j$, $t$).
\item if $p = parent[c]$ equals $c$:\\
$\mb b_c^t$ = $(1-t)\mb b_c^i+t\mb b_c^j$.
\item else:$\mb b_c^t = \mb R_p^t\mb E^p(\mb z|\mb J_{pc}) + \mb b_p^t - \mb R_c^t \mb E^c(\mb z|\mb J_{pc})$
\item For each adjacent part $k$ of $c$\\
If $k$ is not in $visit$:\\
$visit.insert(k), queue.push(k), parent[k] = c$
\end{enumerate}
\item Return $\{\mb R_k^t\mb S_k^{ref}+\mb b_k^t,k=1,...,m\}$.
\end{itemize}
\end{algorithm}

As shown in Algorithm \ref{alg:Interpolate}, given two shapes $\mb S_i$, $\mb S_j$, and the interpolation parameter $t\in[0,1]$, the "root part" $r$ is chosen to have the smallest rotation from $\mb S_i$ to $\mb S_j$. Starting from root part $r$, a broad first search is conducted to interpolate one rigid part by one rigid part. For the rigid part $c$, the rotation $\mb R_c^t$ is interpolated using the spherical linear interpolation of quaternion \cite{lengyel2001mathematics}: Slerp($\mb R_c^i$, $\mb R_c^j$, $t$); the translation $\mb b_c^t$ is linearly interpolated if $c$ is the root part, otherwise it is calculated by:
\begin{eqnarray}
\mb R_c^t E_{p(\mb z | \mb J_{pc}, \bs\Theta_c)}(\mb z) + \mb b_c^t = \mb R_p^t\mb E_{p(\mb z | \mb J_{pc}, \bs\Theta_p)}(\mb z) + \mb b_p^t,
\label{eqn:interpolateTranslation}
\end{eqnarray}
where $p$ is the parent part of $c$, $\mb J_{pc} = \{ \mb J_{pc}^1,...,\mb J_{pc}^{n_s} \}$, $\mb J_{pc}^i$ is the point of contacting 
between parts $p$ and $c$ of the $i$th training shape, which is decided by the mass center of the triangles that are in between parts $p$ and $c$, $E_{p(\mb z | \mb J_{pc}, \bs\Theta_c)}(\mb z)$ is the image of $\mb J_{pc}$ in the latent space of mixture $c$, similar for $E_{p(\mb z | \mb J_{pc}, \bs\Theta_p)}(\mb z)$. Equation \eqref{eqn:interpolateTranslation} means that after the interpolation $E_{p(\mb z | \mb J_{pc}, \bs\Theta_c)}(\mb z)$ and $E_{p(\mb z | \mb J_{pc}, \bs\Theta_p)}(\mb z)$ should still contact each other when mapped to the physical space. $\mb E^p(\mb z|\mb J_{pc})$ is the abbreviation of $E_{p(\mb z | \mb J_{pc}, \bs\Theta_p)}(\mb z)$ in Algorithm \ref{alg:Interpolate}, similar for $\mb E^c(\mb z|\mb J_{pc})$. The final interpolation is generated by:
\begin{eqnarray}
(1-t)\mb S_i + t\mb S_j=\{\mb R_k^t\mb S_k^{ref}+\mb b_k^t + \bs\epsilon_k^t,k=1,...,m\},
\end{eqnarray}
where the residual $\bs\epsilon_k^t = \mb R_k^t\left((1-t){\mb R_k^i}^T\bs\epsilon_k^i + t{\mb R_k^j}^T\bs\epsilon_k^j\right)$ (the muscle movement) is blended in the latent space and is then mapped to the physical space.

\begin{figure}[ht]
	\centering
	\subfigure[\scriptsize$\mb S_3$]{
		\includegraphics[width=0.24\textwidth,trim=1cm 3.5cm 0.5cm 3.5cm, clip]{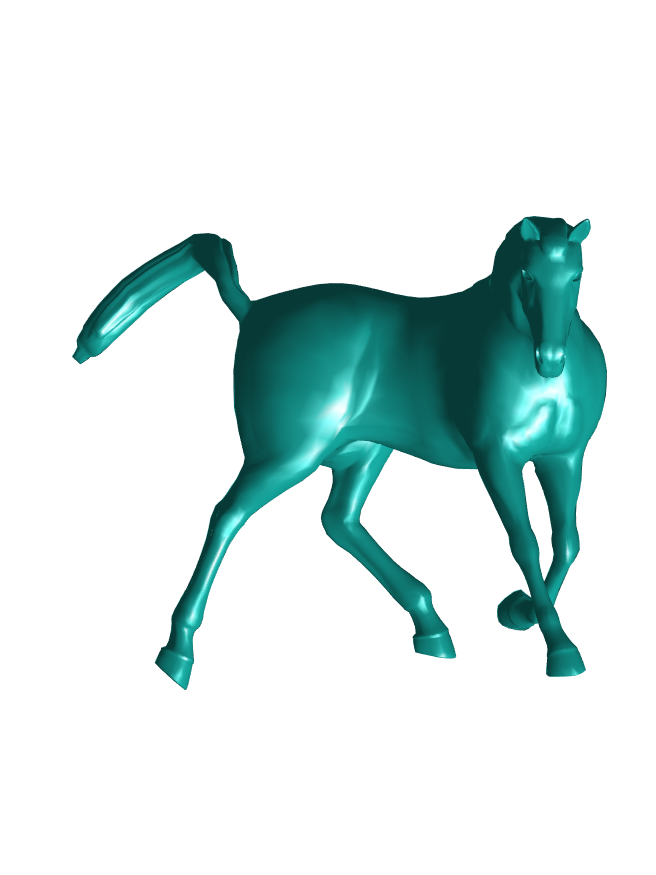}
		\label{fig:horsePoseInterpolate3A}
	}
	\subfigure[\scriptsize$0.75\mb S_3 + 0.25\mb S_{8}$]{
		\includegraphics[width=0.25\textwidth,trim=1cm 3.5cm 0.5cm 3.5cm, clip]{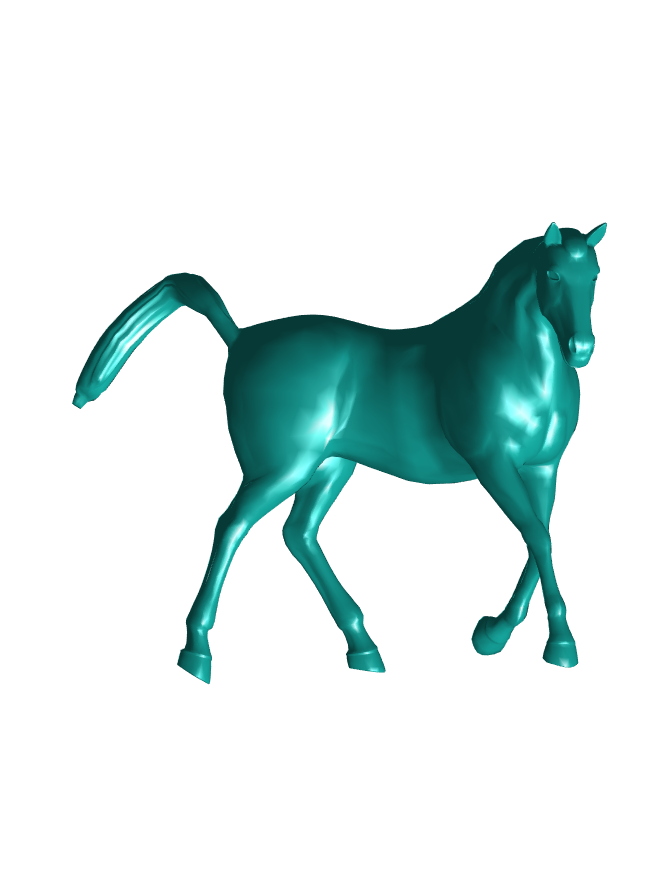}
		\label{fig:horsePoseInterpolate3B}
	}
	\subfigure[\scriptsize$0.5\mb S_3 + 0.5\mb S_{8}$]{
		\includegraphics[width=0.26\textwidth,trim=1cm 3.5cm 0.5cm 3.5cm, clip]{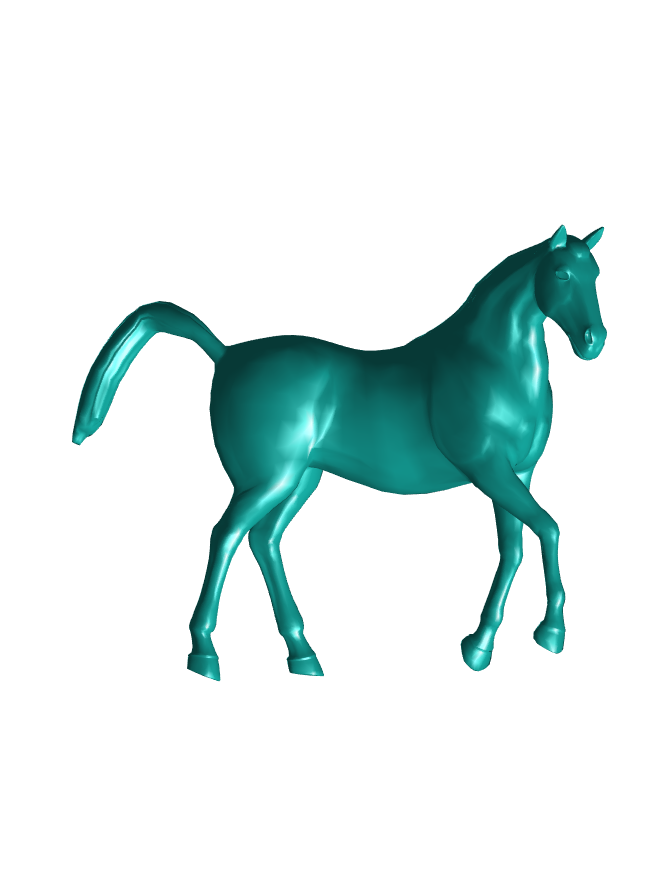}
		\label{fig:horsePoseInterpolate3C}
	}
	\subfigure[\scriptsize$0.25\mb S_3 + 0.75\mb S_{8}$]{
		\includegraphics[width=0.26\textwidth,trim=1cm 3.5cm 0.5cm 3.5cm, clip]{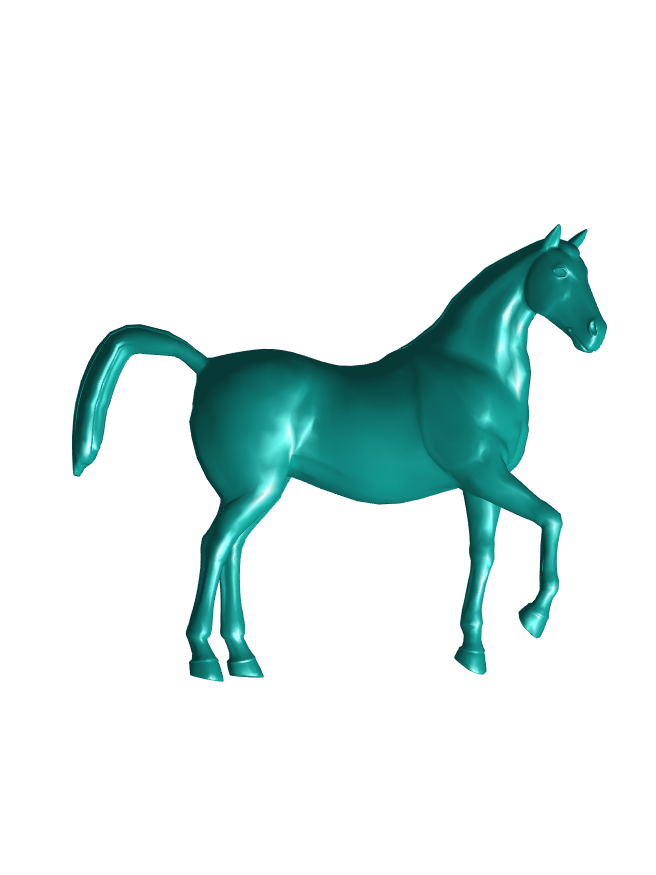}
		\label{fig:horsePoseInterpolate3D}
	}
	\subfigure[\scriptsize$\mb S_{8}$]{
		\includegraphics[width=0.26\textwidth,trim=1cm 3.5cm 0.5cm 3.5cm, clip]{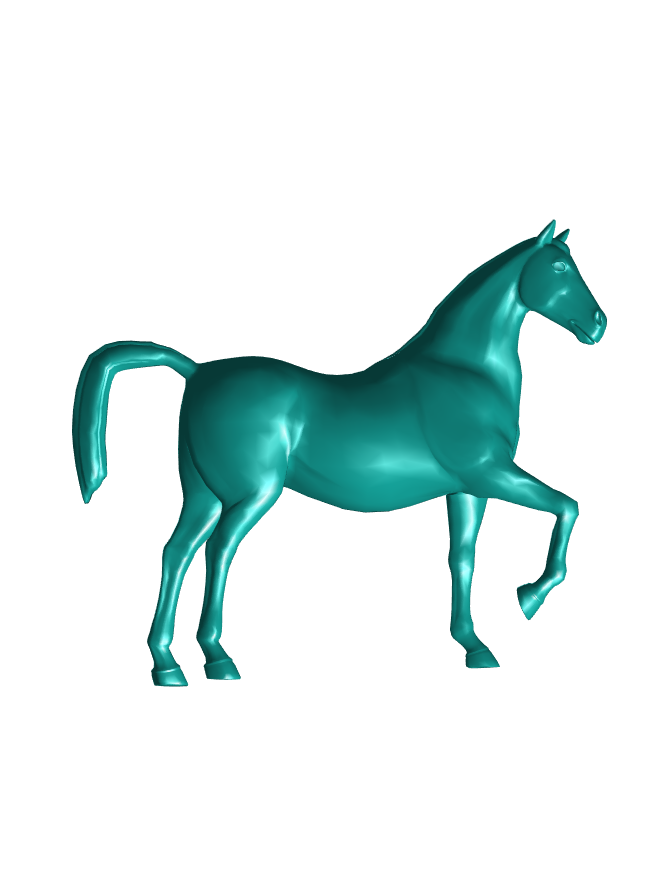}
		\label{fig:horsePoseInterpolate3E}
	}
	\caption{Pose interpolation between the horse shape $\mb S_3$ and $\mb S_{8}$.}
	\label{fig:horsePoseInterpolate3}
\end{figure}

\begin{figure}[ht]
	\centering
	\subfigure[\scriptsize$\mb S_1$]{
		\includegraphics[width=0.28\textwidth,trim=1cm 4cm 0.5cm 5cm, clip]{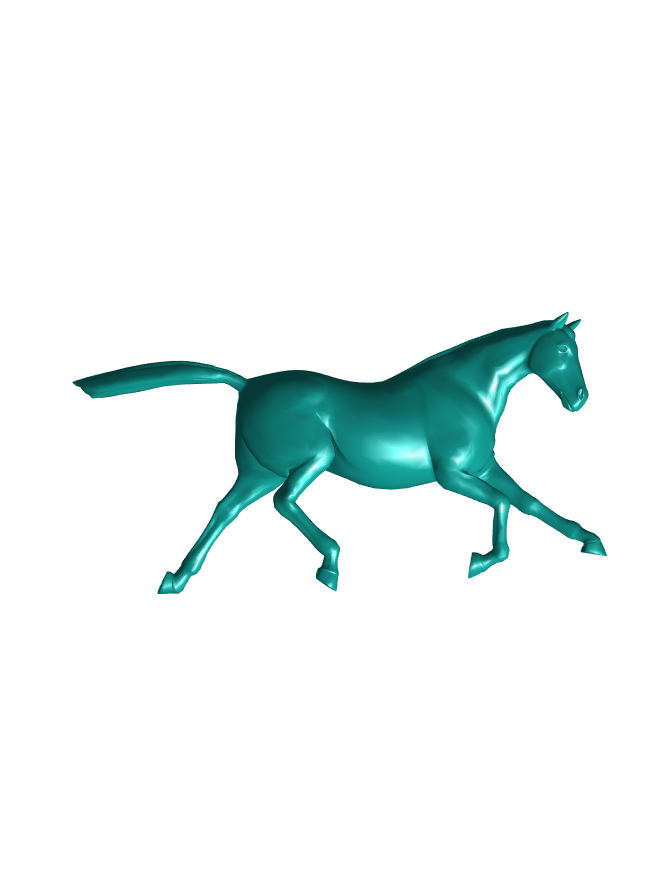}
		\label{fig:horsePoseInterpolate1A}
	}
	\subfigure[\scriptsize$0.5\mb S_1 + 0.5\mb S_{11}$]{
		\includegraphics[width=0.28\textwidth,trim=1cm 4cm 0.5cm 4.5cm, clip]{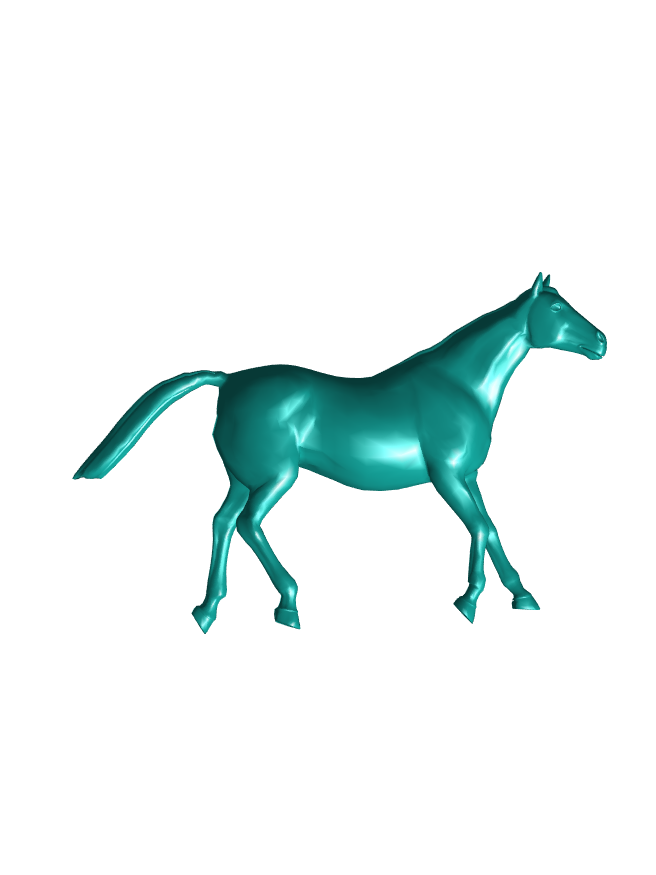}
		\label{fig:horsePoseInterpolate1C}
	}
	\subfigure[\scriptsize$\mb S_{11}$]{
		\includegraphics[width=0.24\textwidth,trim=1cm 3cm 0.5cm 3.5cm, clip]{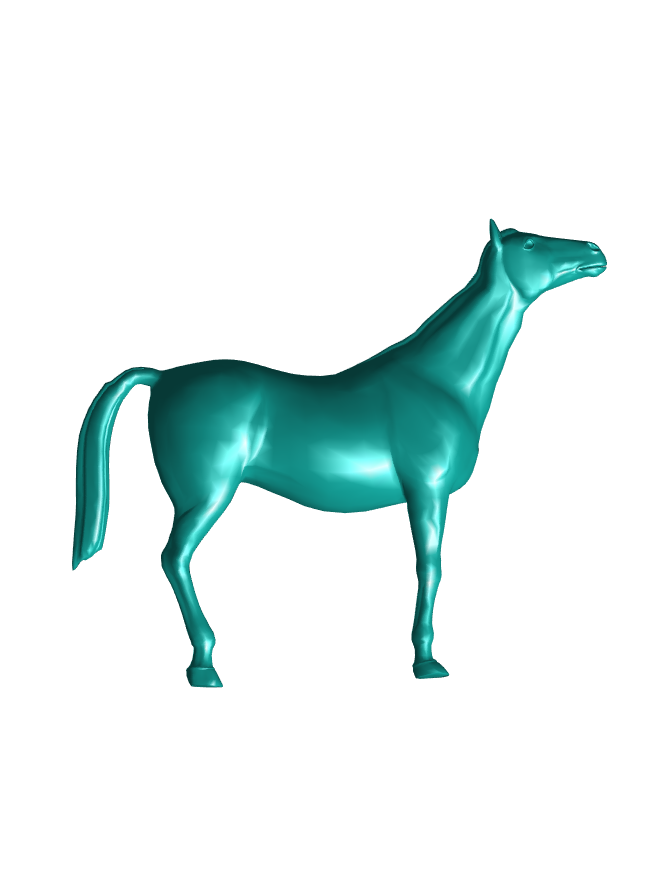}
		\label{fig:horsePoseInterpolate1E}
	}
	\caption{Pose interpolation between the horse shape $\mb S_1$ and $\mb S_{11}$.}
	\label{fig:horsePoseInterpolate1}
\end{figure}

As shown in Figure \ref{fig:horsePoseInterpolate3} are the results of pose interpolation between the horse shape $\mb S_3$ and horse shape $\mb S_{8}$. Figure \ref{fig:horsePoseInterpolate1} are the results of pose interpolation between the horse shape $\mb S_1$ and horse shape $\mb S_{11}$. From the results it can be seen that the interpolated poses are smooth and lifelike.

\section{Results}

In this section experimental results of the examples of horse, flamingo, camel, and cat are presented. As can be shown from the examples, the segmentation of horses, flamingos, and camels are very neat, all the joints are nicely segmented. The segmentation of cats are not as good:

1) The segmentation results in some regions (eg. on the neck and right shoulder of the cat example) are fragmented. This is because more muscle movements are involved in the pose variations of the specific regions of the cat and lion example when compared with the horses, flamingos and camels, so a simple rotation and translation are not enough to describe the movements in these regions (we need many independent rotations).

\begin{figure}[ht]
	\centering
	\subfigure{
		\includegraphics[width=0.22\textwidth,trim=0.5cm 3cm 0.5cm 3cm, clip]{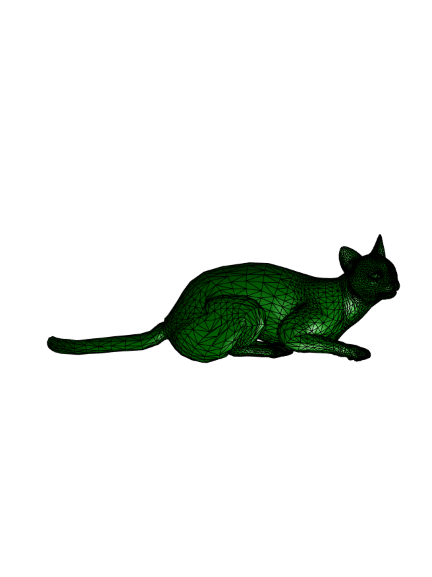}
		\label{fig:11catA}
	}
	\subfigure{
		\includegraphics[width=0.18\textwidth,trim=0.5cm 2cm 0.5cm 2cm, clip]{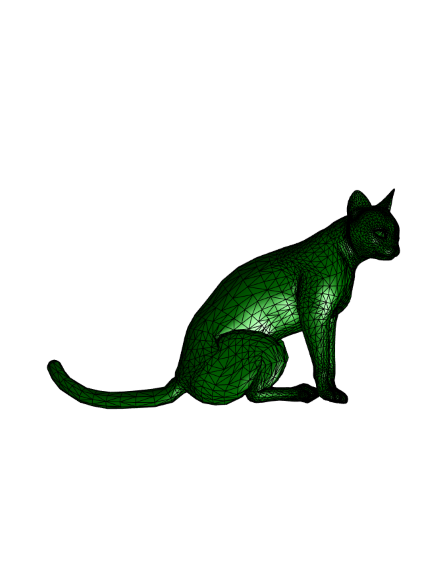}
	}
	\subfigure{
		\includegraphics[width=0.2\textwidth,trim=0.5cm 2cm 0.5cm 2cm, clip]{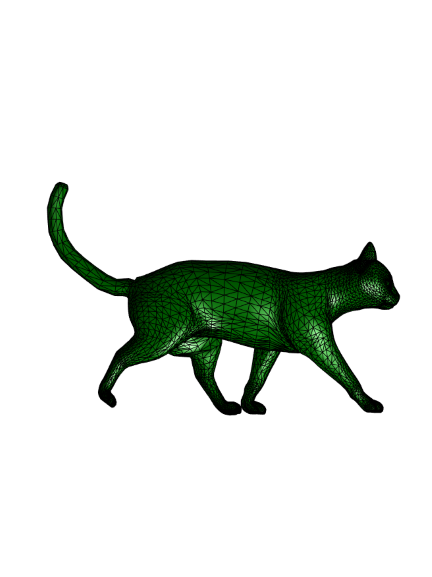}
	}
	\subfigure{
		\includegraphics[width=0.18\textwidth,trim=0.5cm 3cm 0.5cm 3cm, clip]{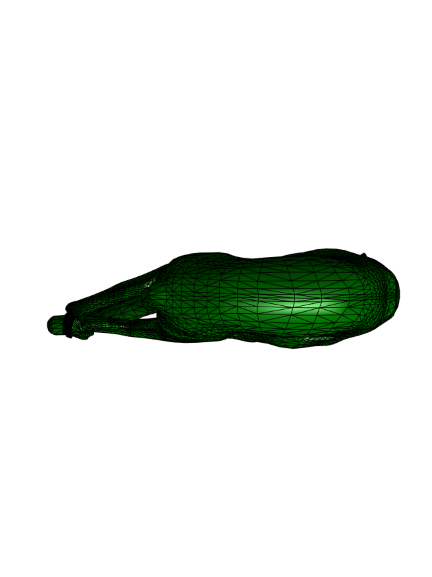}
	}
	\subfigure{
		\includegraphics[width=0.16\textwidth,trim=0.5cm 0.5cm 0.5cm 0.5cm, clip]{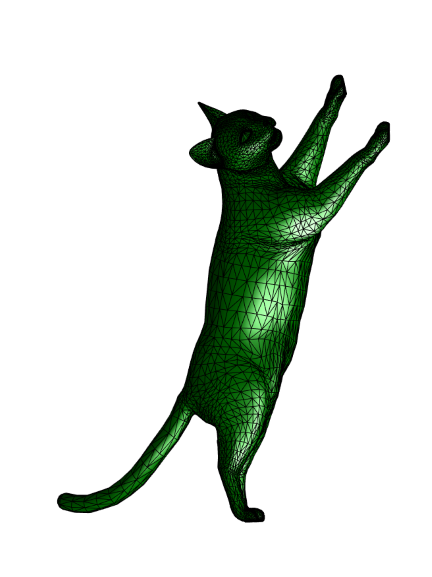}
	}
	\subfigure{
		\includegraphics[width=0.22\textwidth,trim=0.5cm 3cm 0.5cm 3cm, clip]{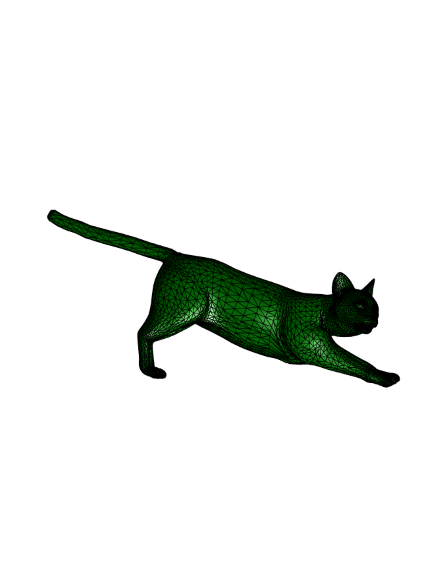}
	}
	\subfigure{
		\includegraphics[width=0.16\textwidth,trim=0.5cm 3cm 0.5cm 3cm, clip]{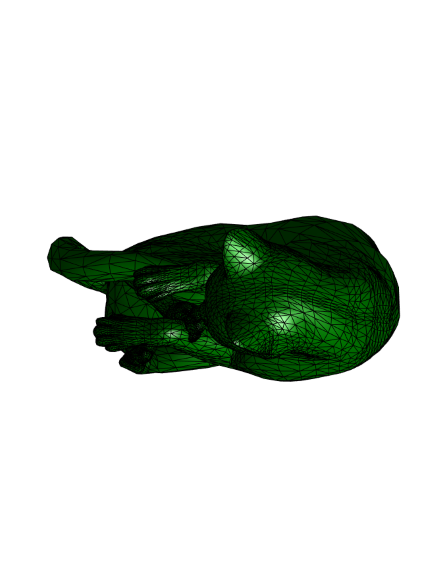}
	}
	\subfigure{
		\includegraphics[width=0.14\textwidth,trim=0.5cm 1.8cm 0.5cm 2cm, clip]{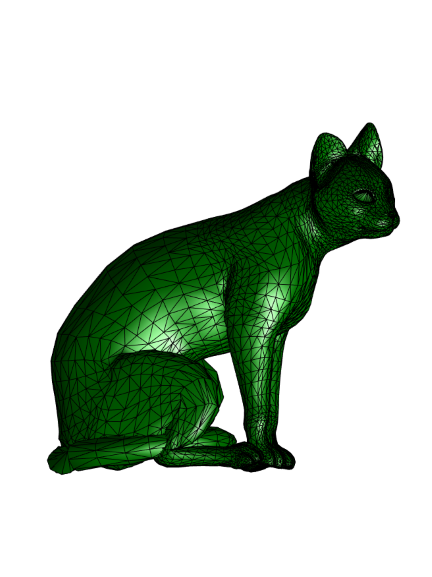}
	}
	\subfigure{
		\includegraphics[width=0.14\textwidth,trim=0.5cm 1.8cm 0.5cm 2cm, clip]{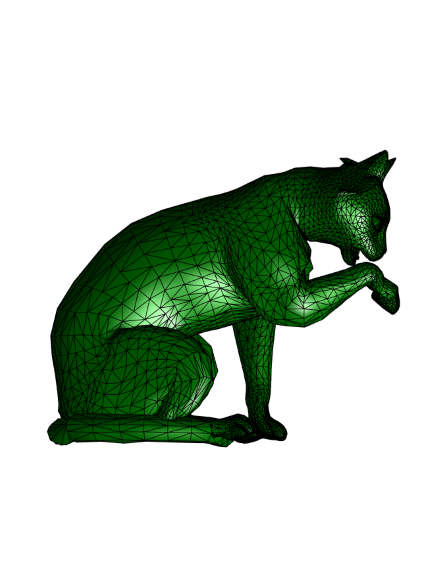}
	}
	\subfigure{
		\includegraphics[width=0.22\textwidth,trim=0.5cm 3cm 0.5cm 3cm, clip]{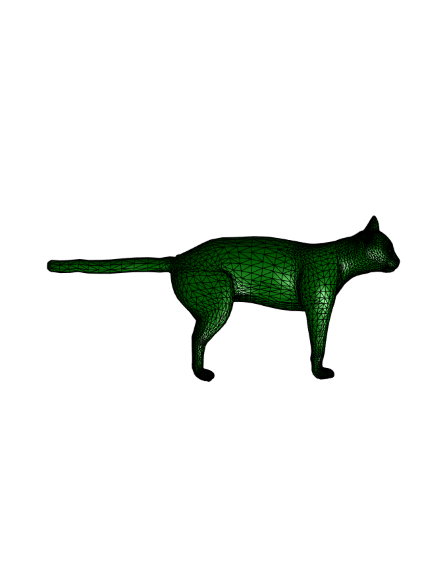}
	}
	\caption{The 10 cat training shapes in different poses.}
	\label{fig:11cat}
\end{figure}

2) The segmentation is very coarse in some regions (eg. the joints are not separated on left front leg of the cat model, and the joints are not separated on the right back leg of the elephant model). This is caused by the limited number of poses we have in the data set. For example, in Figure \ref{fig:11cat} the bending of the right front leg of the cat is observed (the 10th subfigure), but similar movements is missing for the left front leg, and that's why the segmentation of the left front leg is not as fine as the right front leg, since there is not enough movements to learn from. In the elephant example, the two front legs have much more movements than the two back legs, thus the segmentation of the two front legs is much finer than the segmentation of the two back legs.

It should be noted that the first defect (fragmented segmentation) will harm the results of interpolation, while the second defect will not harm the interpolation since such movements is not presented in the data.

\begin{figure}[ht]
	\centering
	\subfigure{
		\includegraphics[width=0.2\textwidth,trim=0.5cm 1cm 0.5cm 1.5cm, clip]{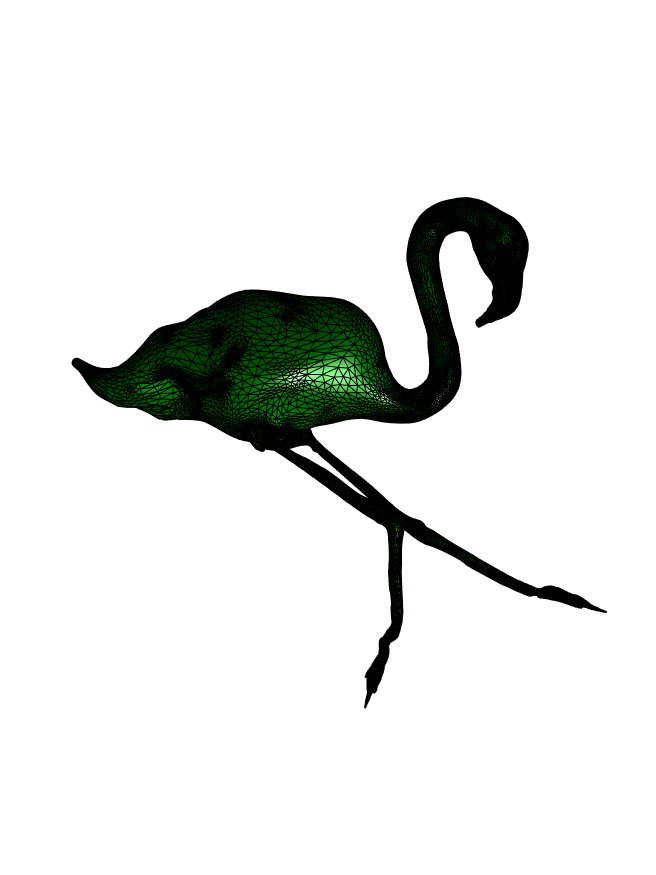}
		\label{fig:11flamA}
	}
	\subfigure{
		\includegraphics[width=0.2\textwidth,trim=0.5cm 1cm 0.5cm 1.5cm, clip]{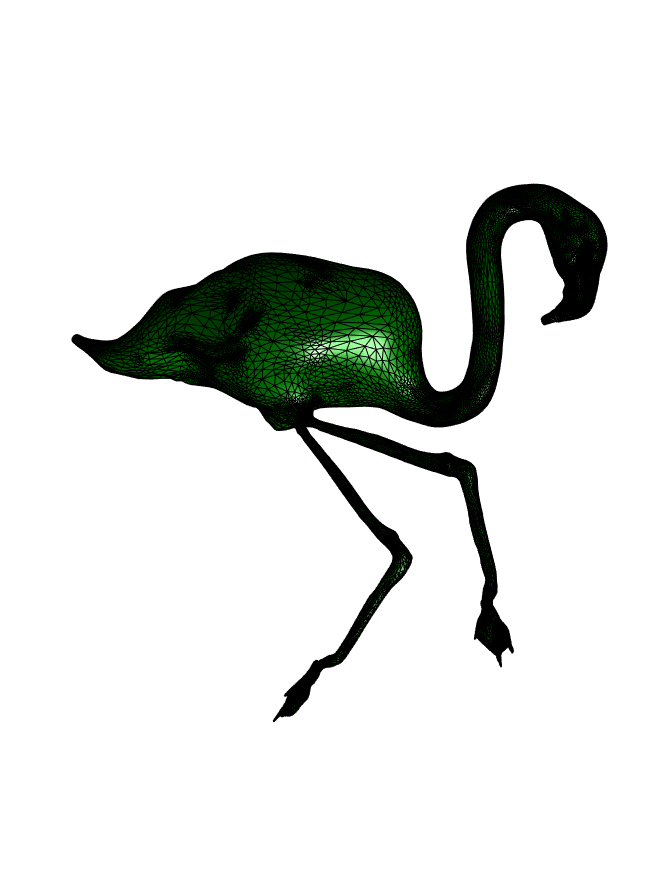}
	}
	\subfigure{
		\includegraphics[width=0.2\textwidth,trim=0.5cm 1cm 0.5cm 1.5cm, clip]{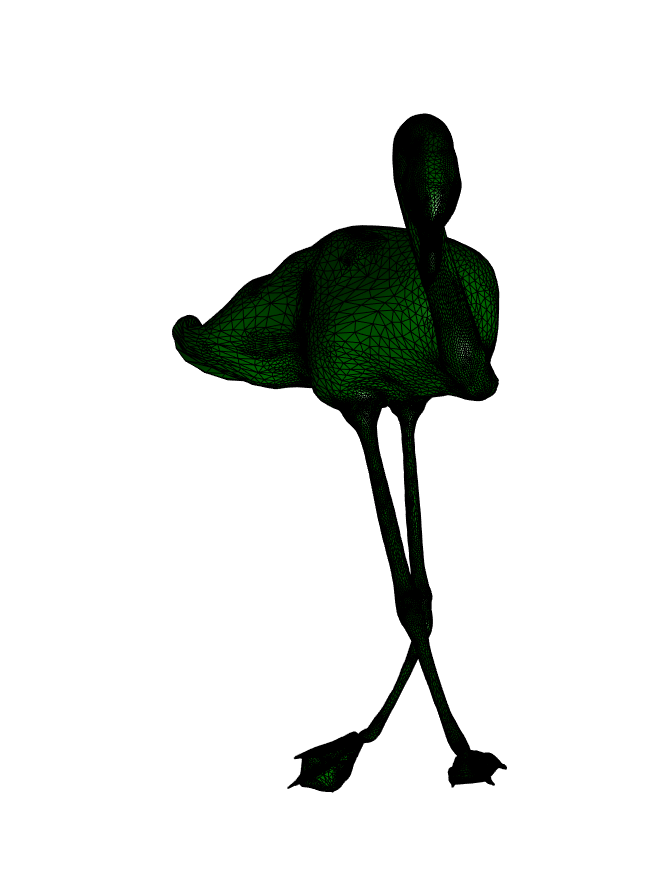}
	}
	\subfigure{
		\includegraphics[width=0.2\textwidth,trim=0.5cm 1cm 0.5cm 1.5cm, clip]{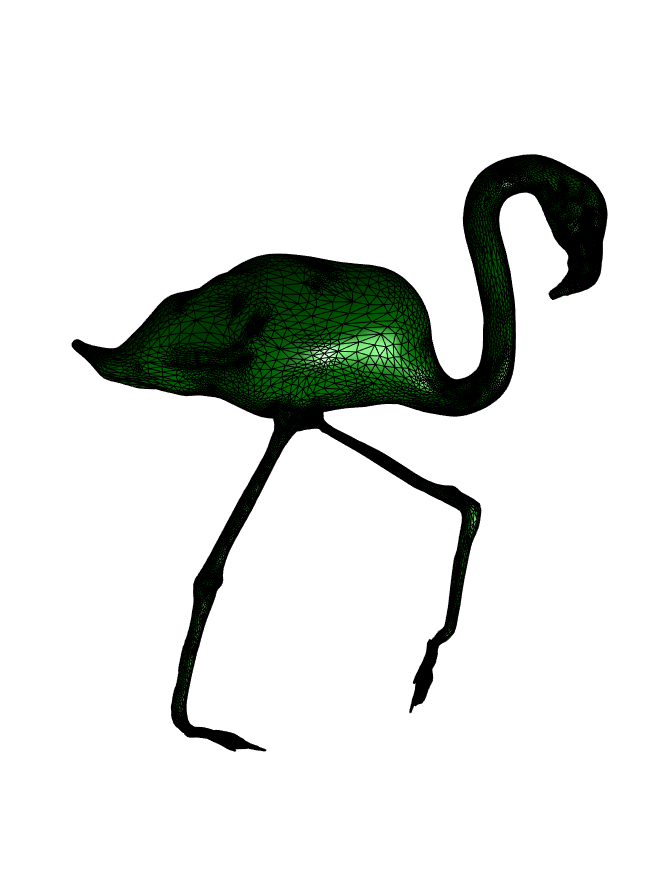}
	}
	\subfigure{
		\includegraphics[width=0.2\textwidth,trim=0.5cm 1cm 0.5cm 1.5cm, clip]{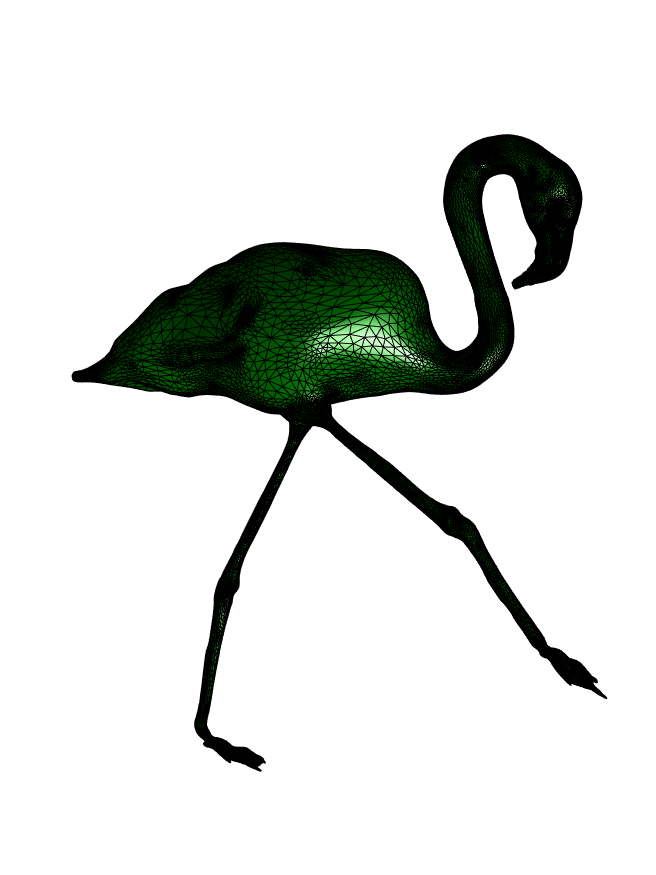}
	}
	\subfigure{
		\includegraphics[width=0.2\textwidth,trim=0.5cm 1cm 0.5cm 1.5cm, clip]{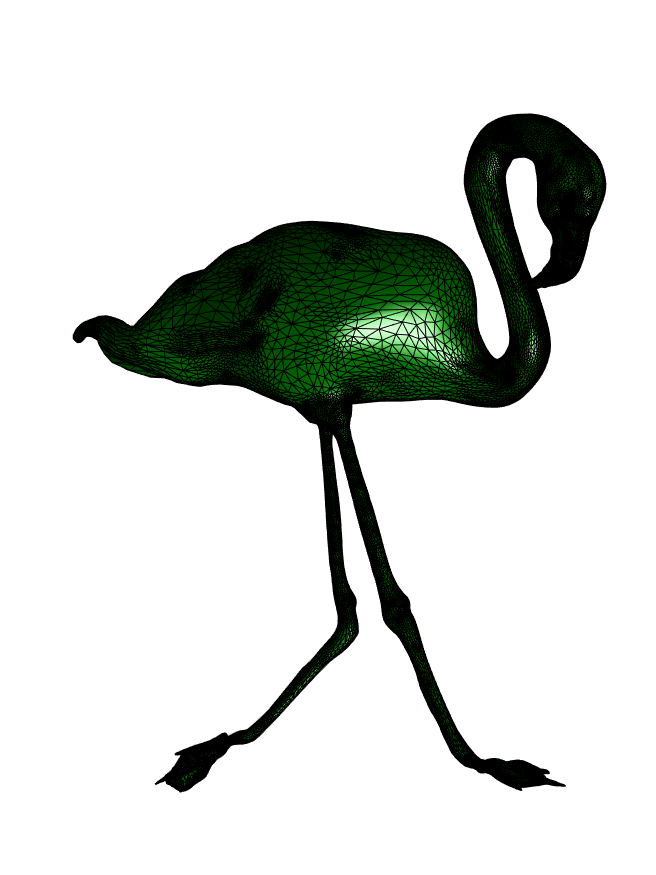}
	}
	\subfigure{
		\includegraphics[width=0.2\textwidth,trim=0.5cm 1cm 0.5cm 1.5cm, clip]{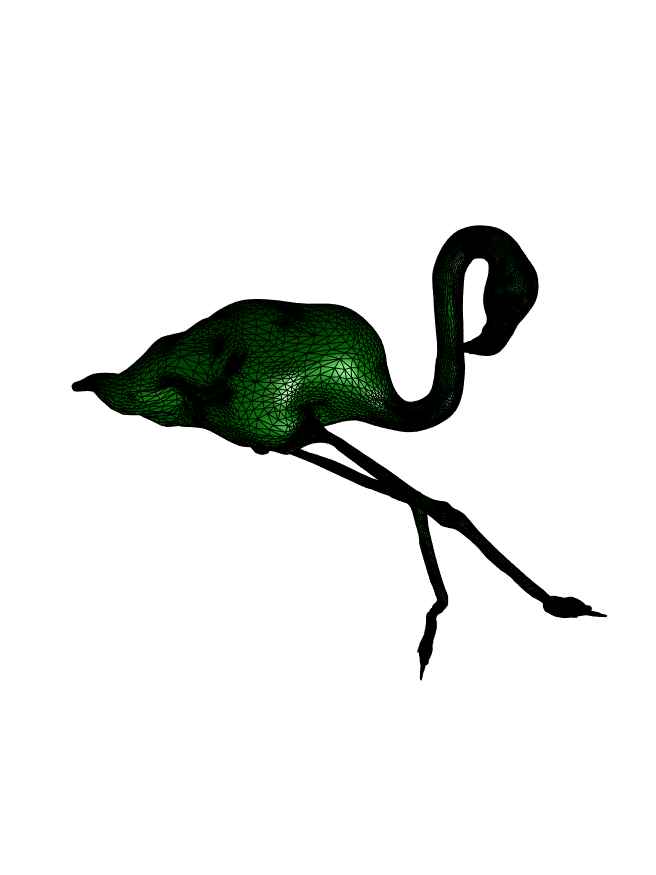}
	}
	\subfigure{
		\includegraphics[width=0.2\textwidth,trim=0.5cm 1cm 0.5cm 1.5cm, clip]{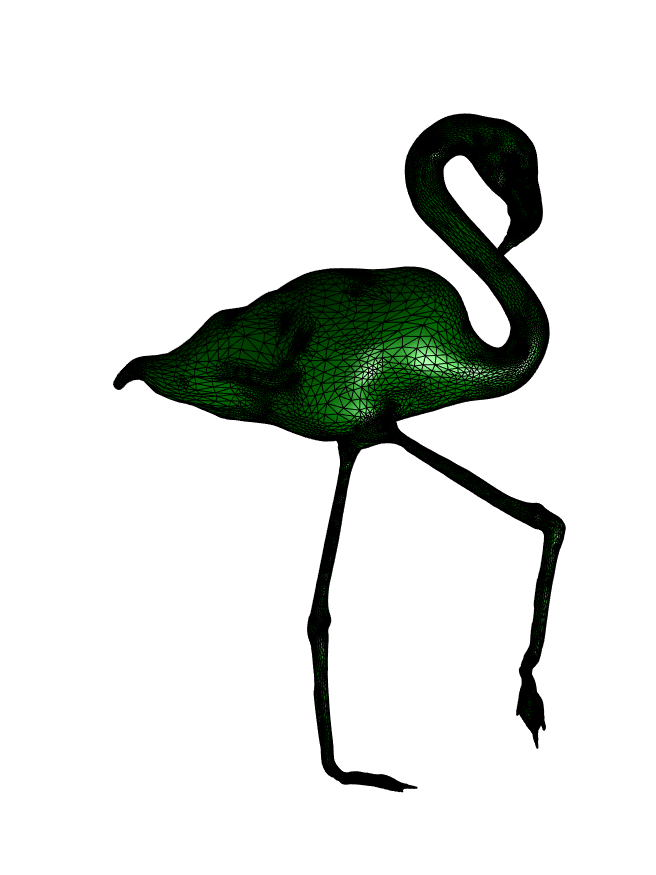}
	}
	\subfigure{
		\includegraphics[width=0.2\textwidth,trim=0.5cm 1cm 0.5cm 1.5cm, clip]{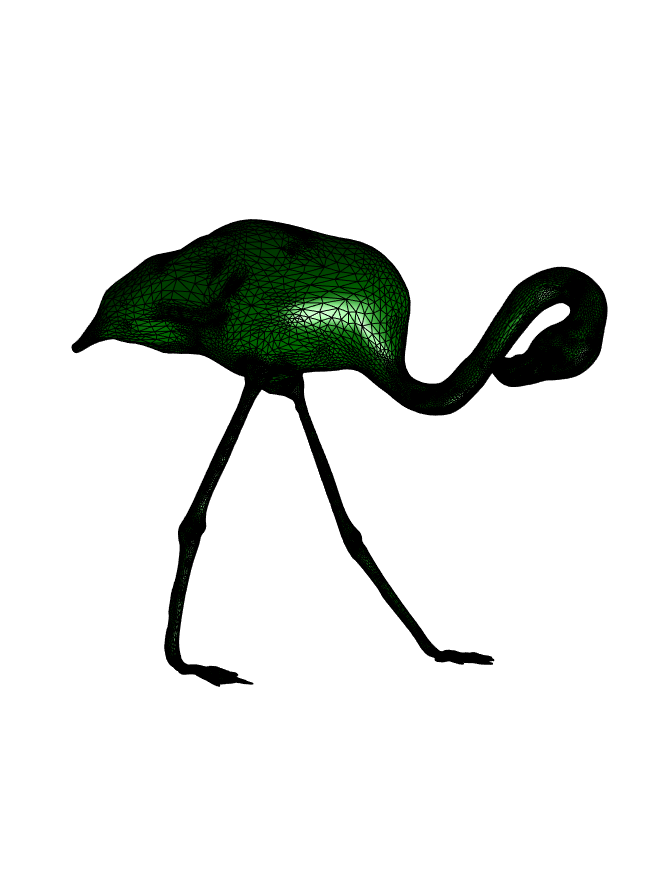}
	}
	\subfigure{
		\includegraphics[width=0.2\textwidth,trim=0.5cm 1cm 0.5cm 1.5cm, clip]{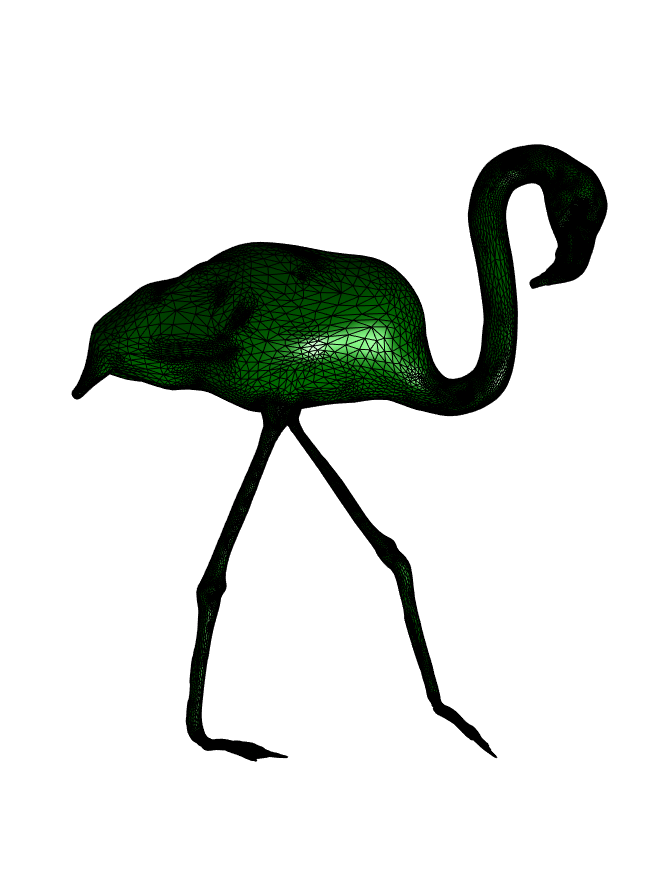}
	}
	\subfigure{
		\includegraphics[width=0.2\textwidth,trim=0.5cm 1.5cm 0.5cm 1.5cm, clip]{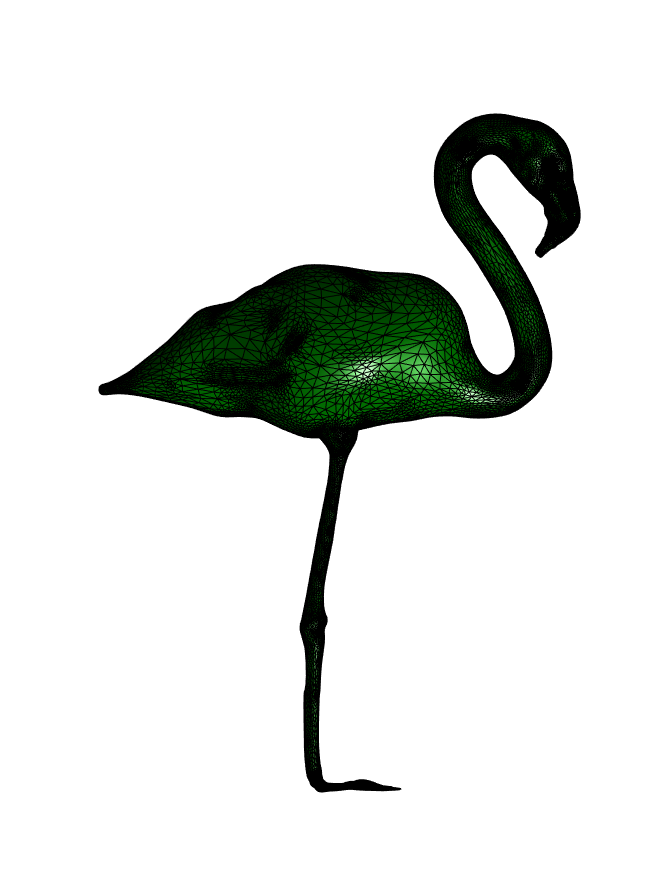}
	}
	\caption{The 11 flamingo training shapes in different poses.}
	\label{fig:11flam}
\end{figure}

\begin{figure}[ht]
	\centering
	\subfigure[]{
		\includegraphics[width=0.24\textwidth,trim=0.5cm 1.5cm 0.5cm 1.5cm, clip]{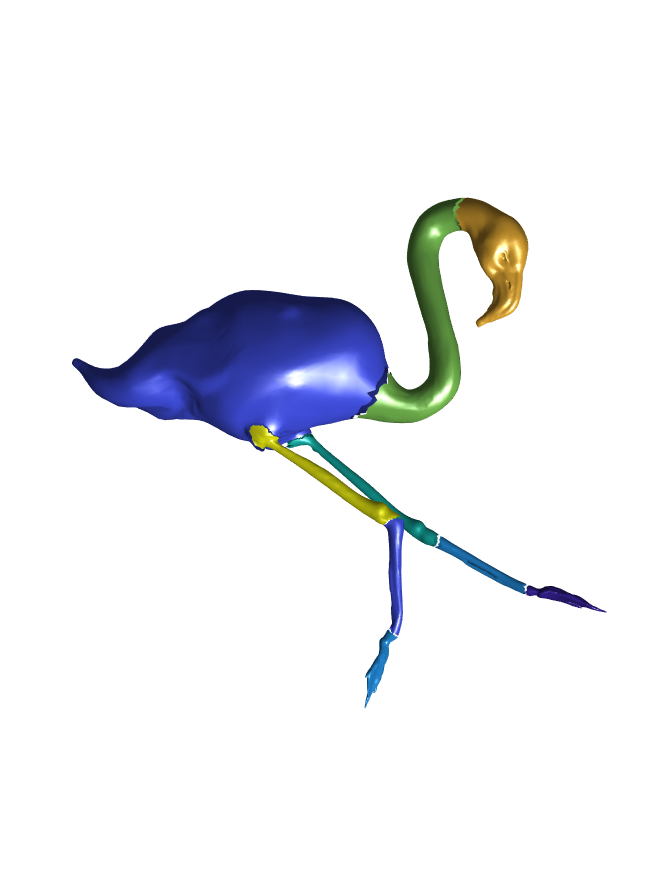}
		\label{fig:flamHierarchicalOptimizeA}
	}
	\subfigure[]{
		\includegraphics[width=0.24\textwidth,trim=0.5cm 1.5cm 0.5cm 1.5cm, clip]{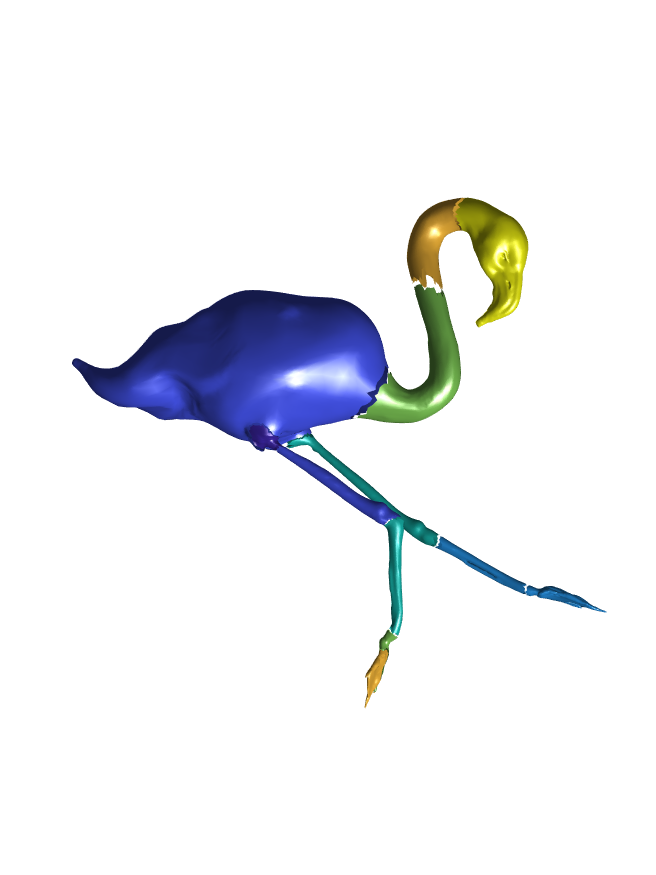}
		\label{fig:flamHierarchicalOptimizeB}
	}
	\subfigure[]{
		\includegraphics[width=0.24\textwidth,trim=0.5cm 1.5cm 0.5cm 1.5cm, clip]{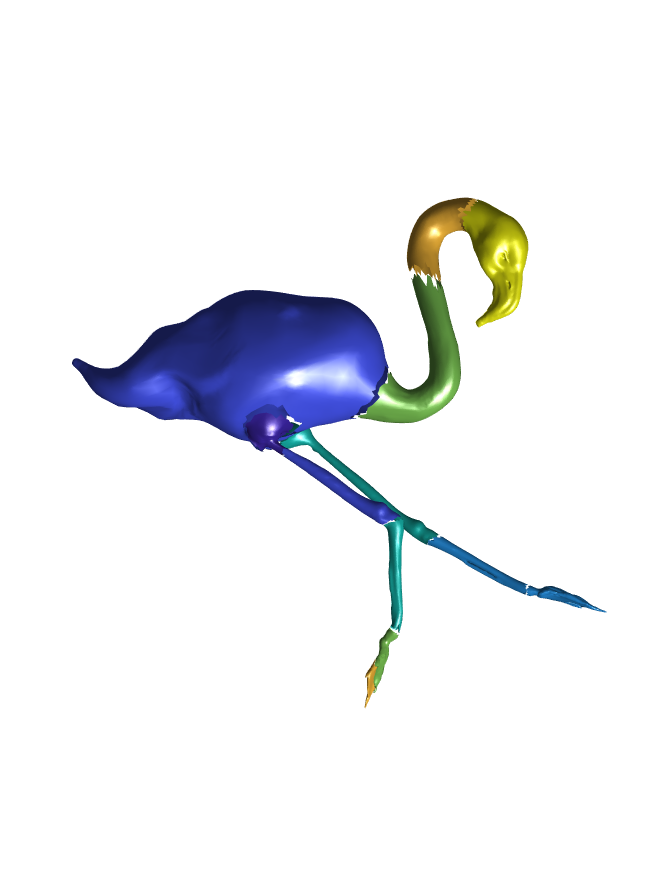}
		\label{fig:flamHierarchicalOptimizeC}
	}
	\caption{Hierarchical optimization of the flamingo example: a) initial segmentation with 9 clusters; b) refined segmentation with 12 clusters; c) final segmetation with 12 clusters.}
	\label{fig:flamHierarchicalOptimize}
\end{figure}

\begin{figure}[ht]
	\centering
	\subfigure[\scriptsize$\mb S_3$]{
		\includegraphics[width=0.26\textwidth,trim=0.5cm 1.5cm 0.5cm 1.5cm, clip]{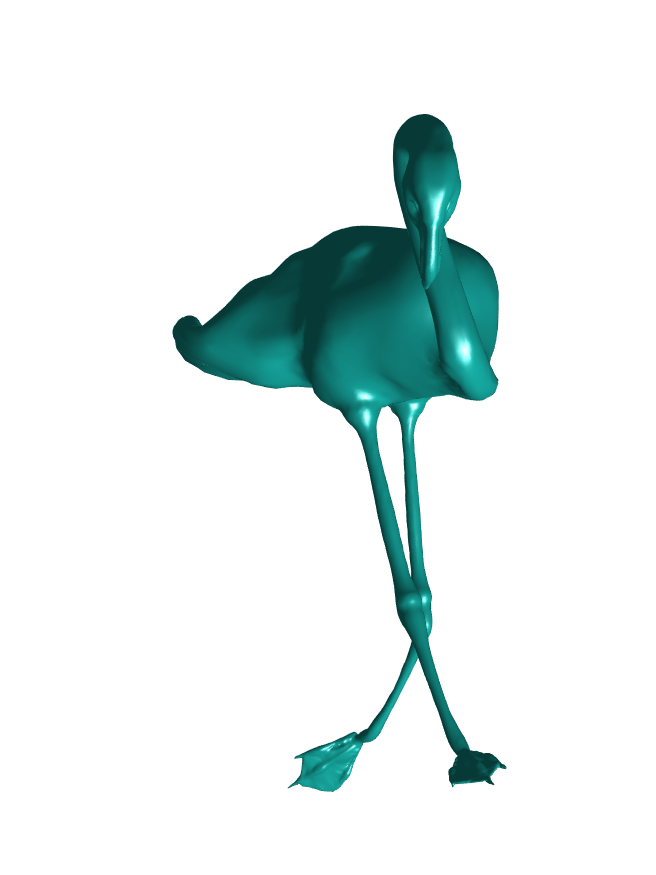}
		\label{fig:flamPoseInterpolate3A}
	}
	\subfigure[\scriptsize$0.75\mb S_3 + 0.25\mb S_{8}$]{
		\includegraphics[width=0.26\textwidth,trim=0.5cm 1.5cm 0.5cm 1.5cm, clip]{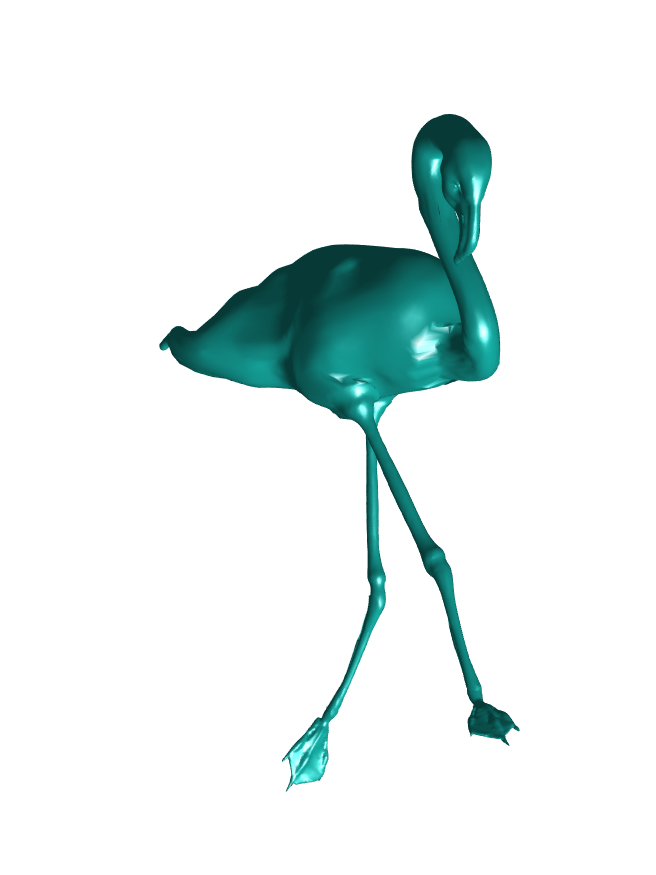}
		\label{fig:flamPoseInterpolate3B}
	}
	\subfigure[\scriptsize$0.5\mb S_3 + 0.5\mb S_{8}$]{
		\includegraphics[width=0.26\textwidth,trim=0.5cm 1.5cm 0.5cm 1.5cm, clip]{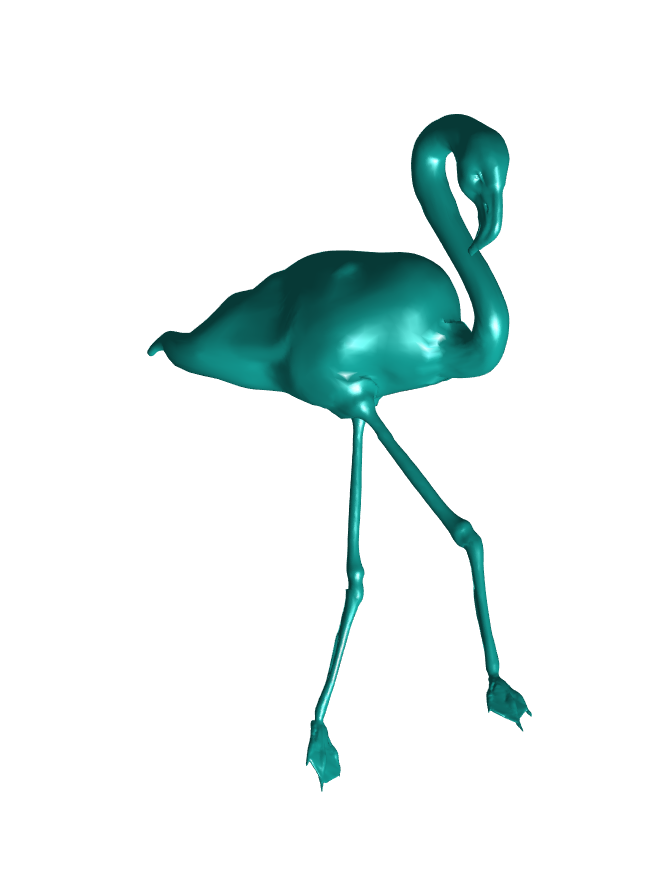}
		\label{fig:flamPoseInterpolate3C}
	}
	\subfigure[\scriptsize$0.25\mb S_3 + 0.75\mb S_{8}$]{
		\includegraphics[width=0.26\textwidth,trim=0.5cm 1.5cm 0.5cm 1.5cm, clip]{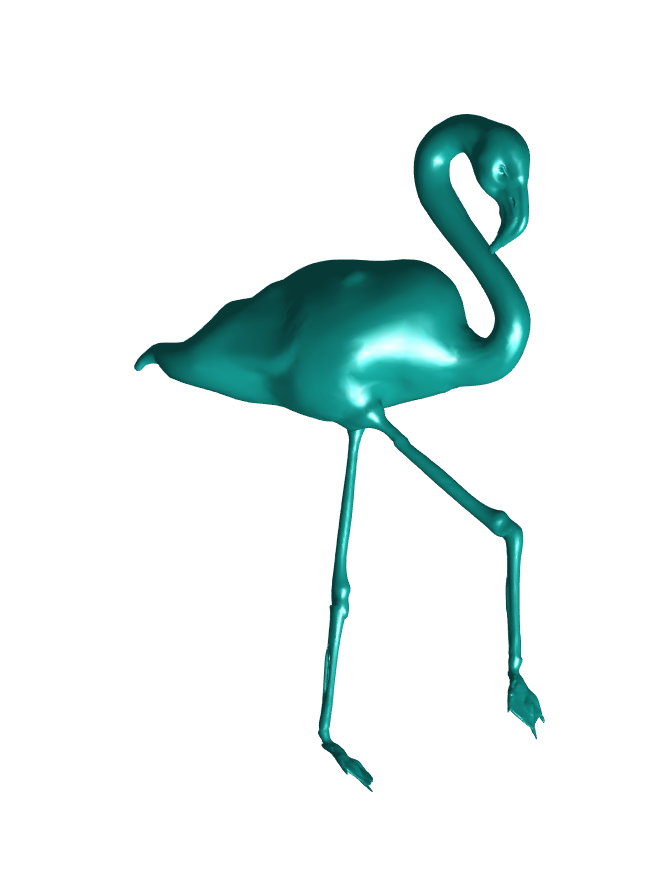}
		\label{fig:flamPoseInterpolate3D}
	}
	\subfigure[\scriptsize$\mb S_{8}$]{
		\includegraphics[width=0.26\textwidth,trim=0.5cm 1.5cm 0.5cm 1.5cm, clip]{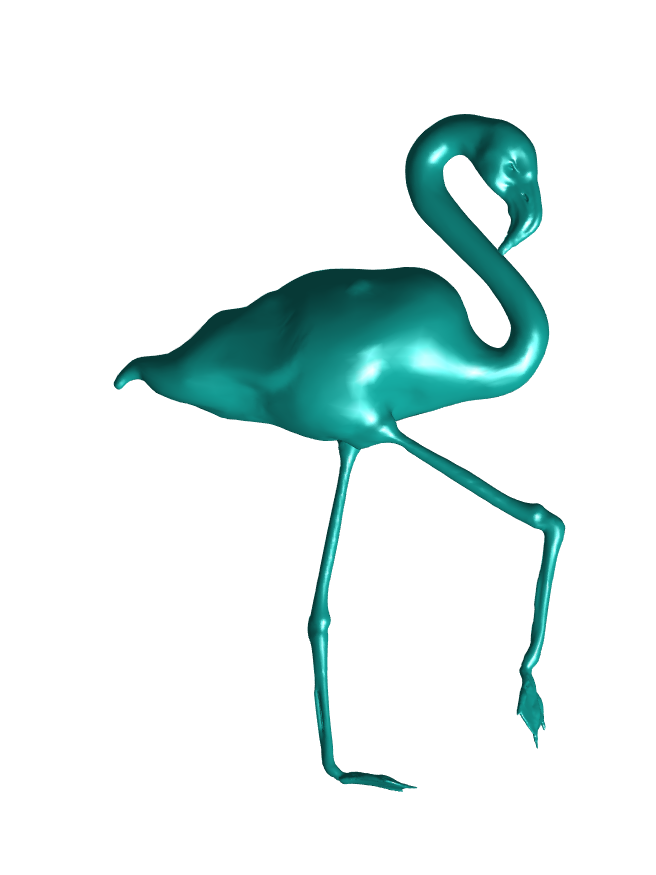}
		\label{fig:flamPoseInterpolate3E}
	}
	\caption{Pose interpolation between the two flam shapes $\mb S_3$ and $\mb S_8$.}
	\label{fig:flamPoseInterpolate3}
\end{figure}

\begin{figure}[ht]
	\centering
	\subfigure{
		\includegraphics[width=0.22\textwidth,trim=1cm 4cm 0.5cm 5cm, clip]{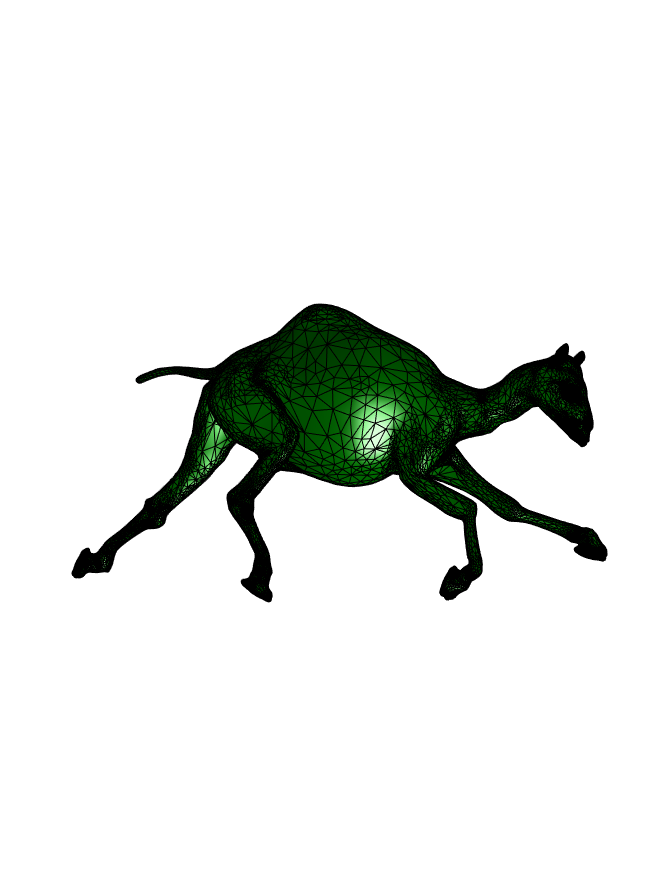}
		\label{fig:11camelA}
	}
	\subfigure{
		\includegraphics[width=0.2\textwidth,trim=1cm 4cm 0.5cm 4cm, clip]{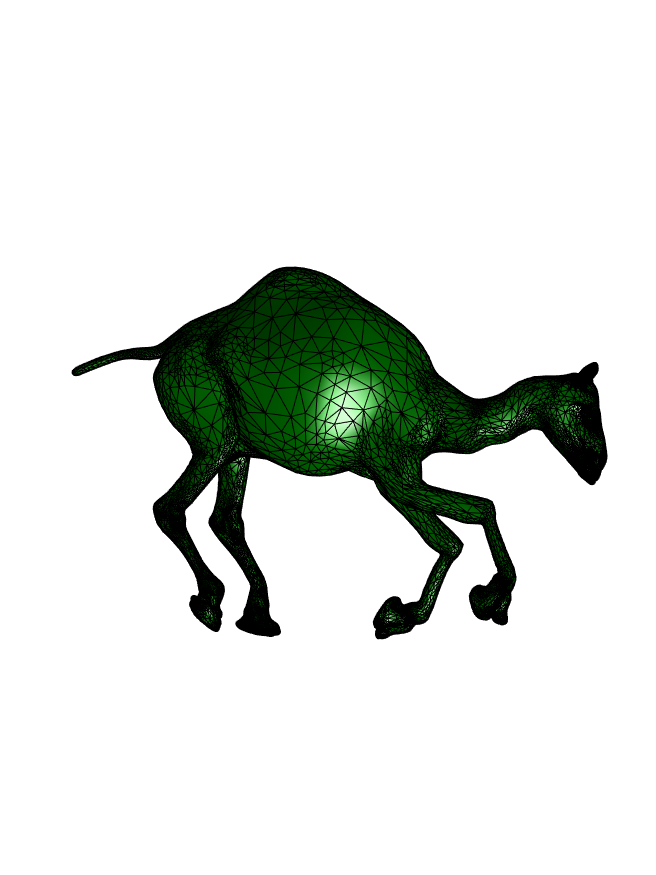}
	}
	\subfigure{
		\includegraphics[width=0.16\textwidth,trim=1cm 3cm 0.5cm 3cm, clip]{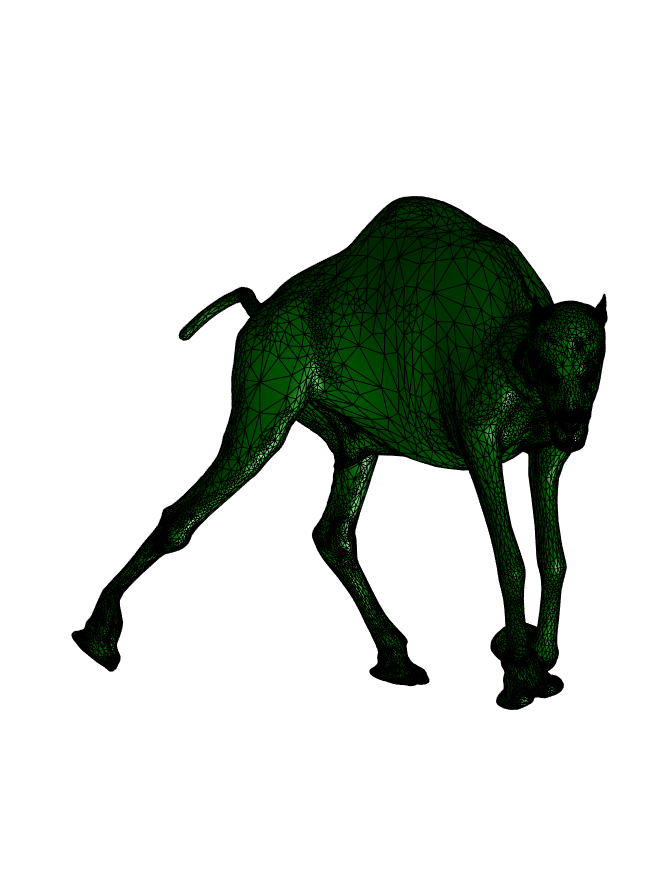}
	}
	\subfigure{
		\includegraphics[width=0.2\textwidth,trim=1cm 4cm 0.5cm 4cm, clip]{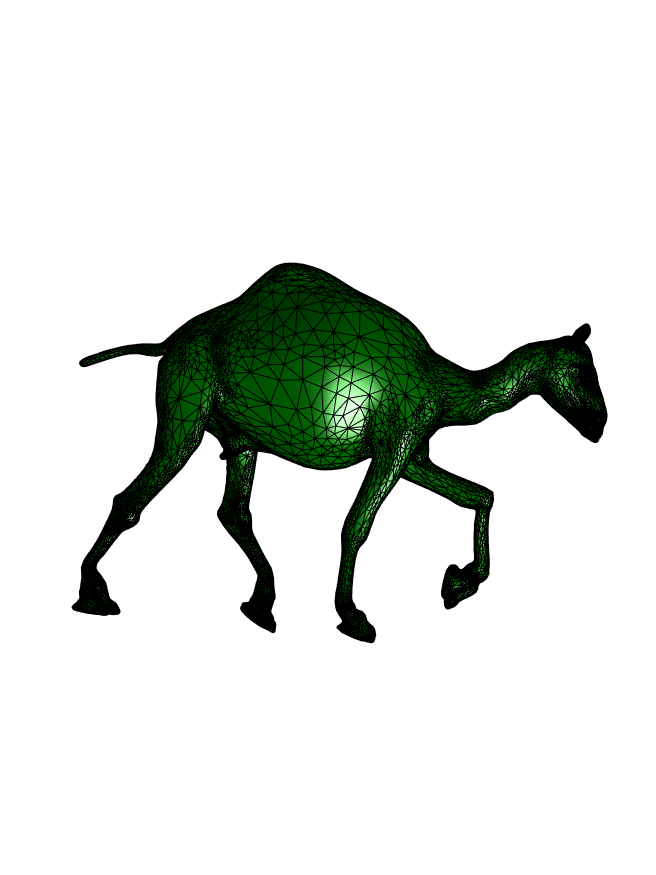}
	}
	\subfigure{
		\includegraphics[width=0.2\textwidth,trim=1cm 4cm 0.5cm 4.5cm, clip]{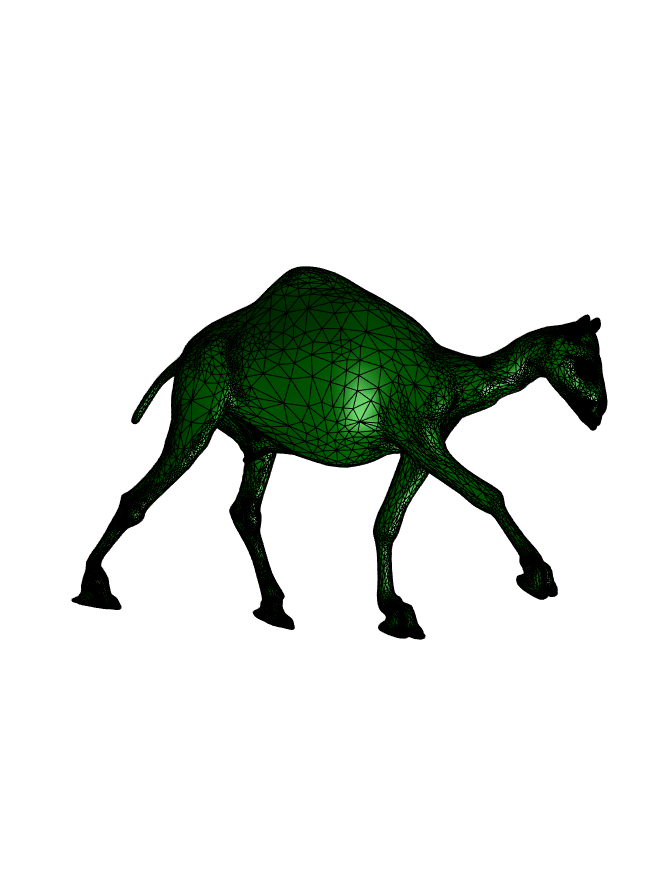}
	}
	\subfigure{
		\includegraphics[width=0.2\textwidth,trim=1cm 4cm 0.5cm 4.5cm, clip]{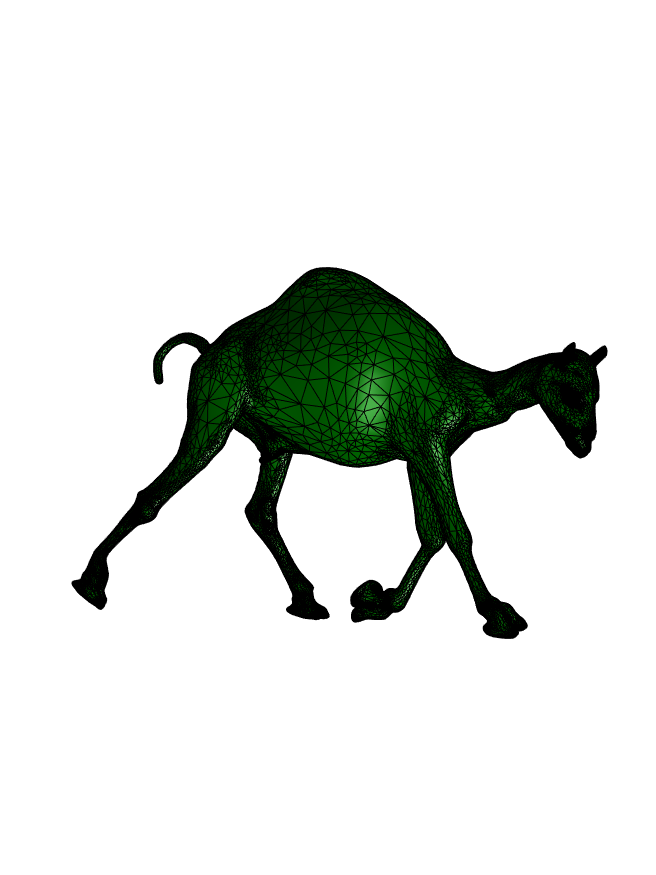}
	}
	\subfigure{
		\includegraphics[width=0.2\textwidth,trim=1cm 4.5cm 1cm 5cm, clip]{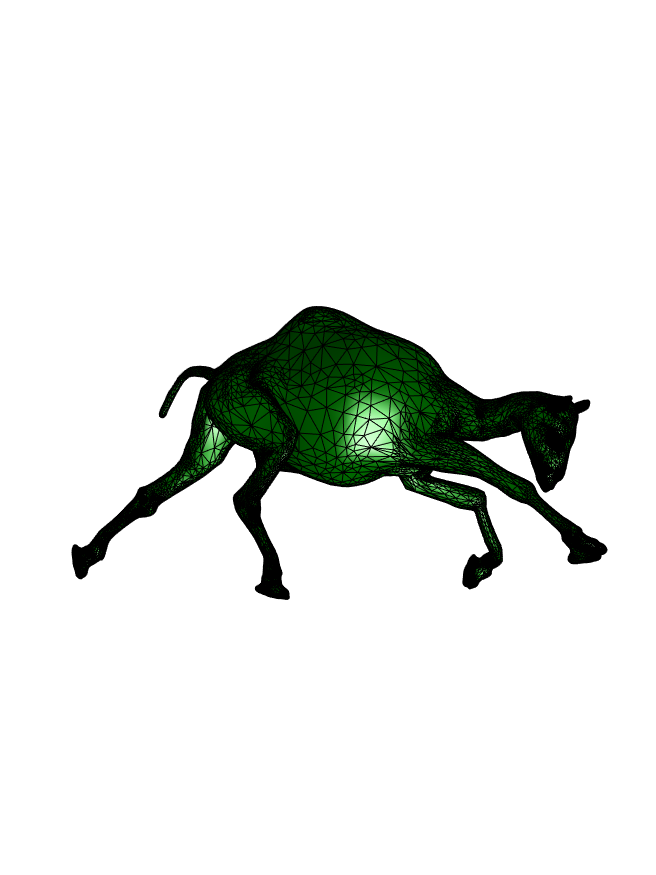}
	}
	\subfigure{
		\includegraphics[width=0.18\textwidth,trim=1cm 3cm 0.5cm 3.5cm, clip]{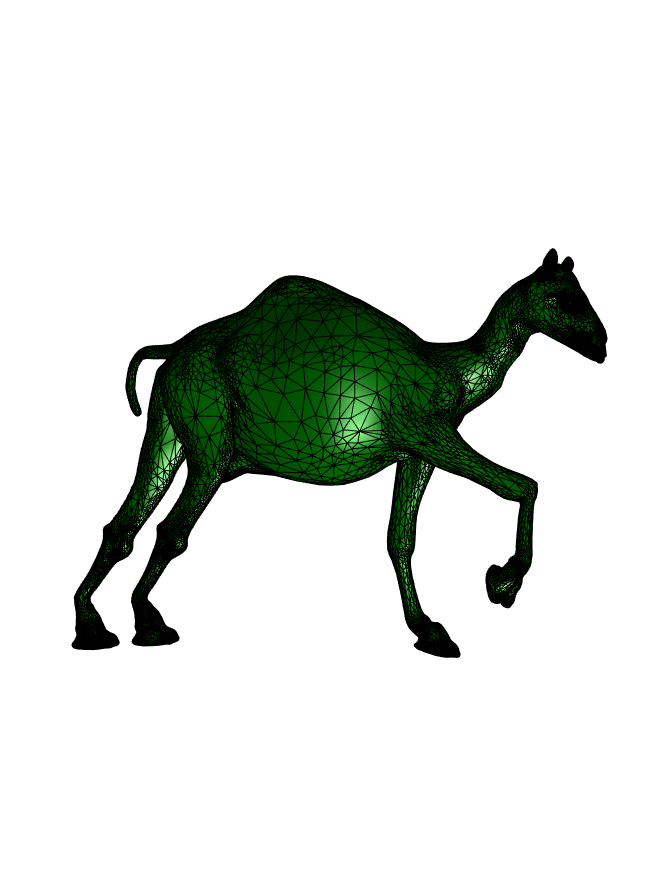}
	}
	\subfigure{
		\includegraphics[width=0.18\textwidth,trim=1cm 3cm 0.5cm 4cm, clip]{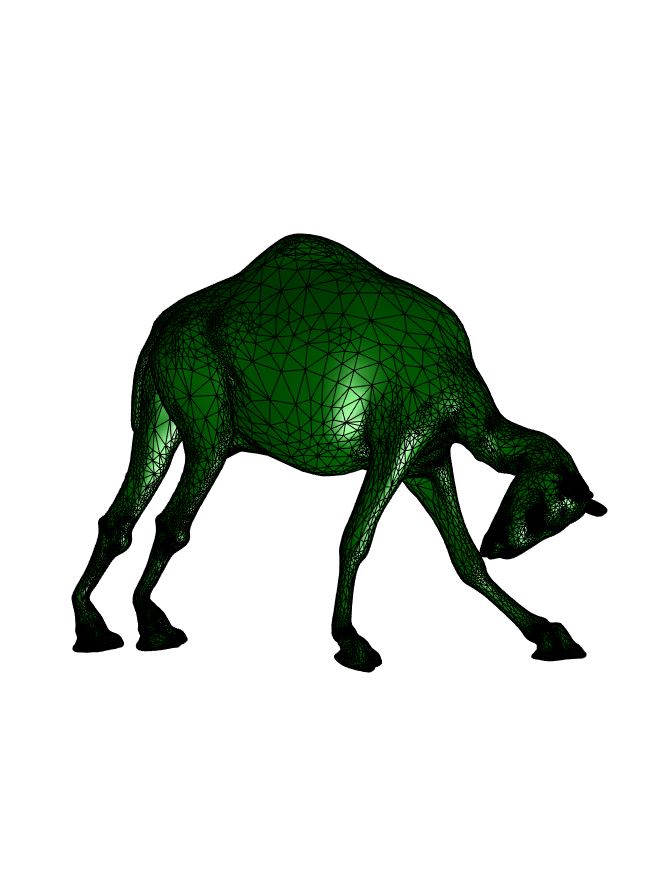}
	}
	\subfigure{
		\includegraphics[width=0.2\textwidth,trim=1cm 3.5cm 0.5cm 4cm, clip]{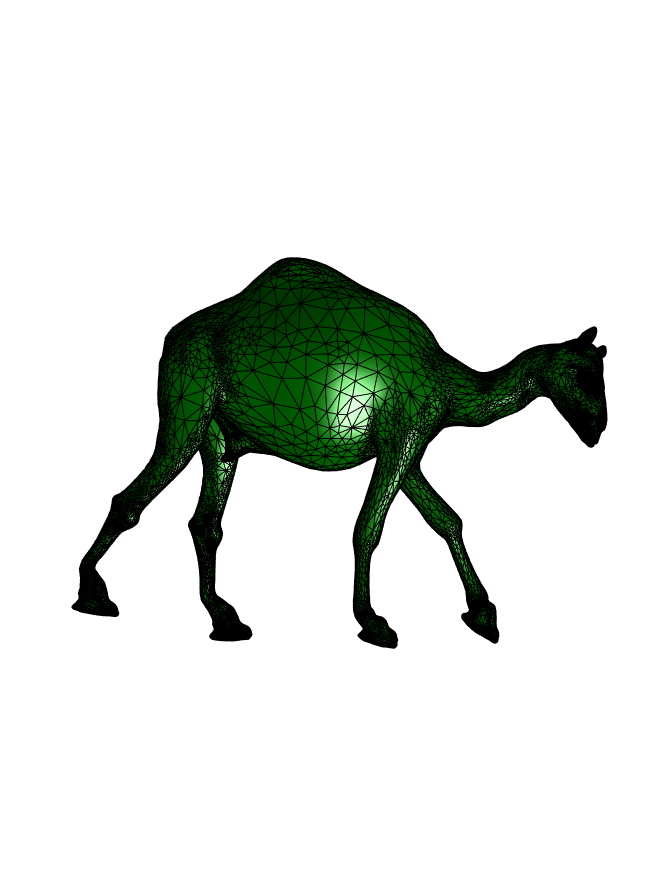}
	}
	\subfigure{
		\includegraphics[width=0.18\textwidth,trim=1cm 3cm 0.5cm 3cm, clip]{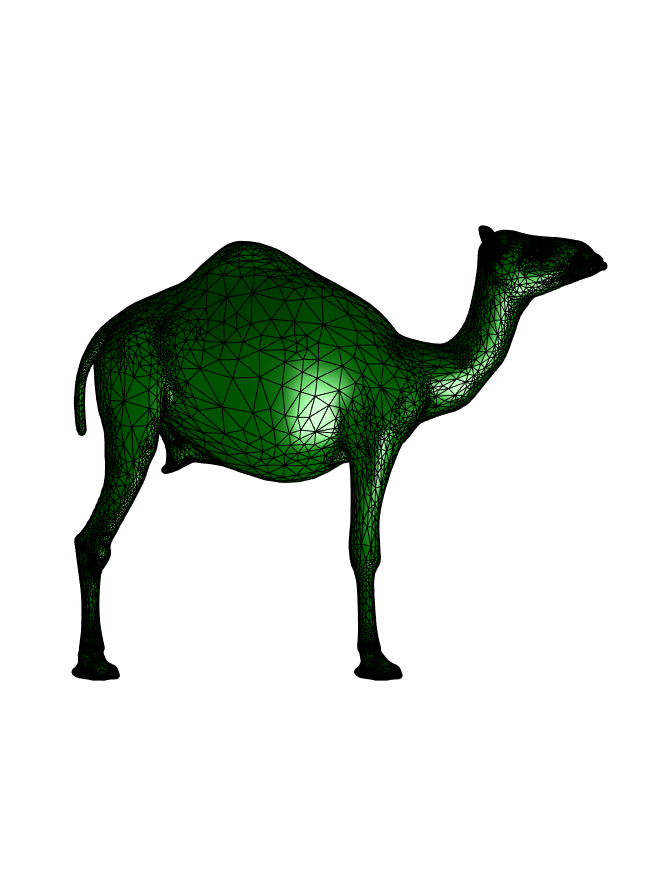}
	}
	\caption{The 11 camel training shapes in different poses.}
	\label{fig:11camel}
\end{figure}

\begin{figure}[ht]
	\centering
	\subfigure[]{
		\includegraphics[width=0.28\textwidth,trim=1cm 4cm 1cm 5cm, clip]{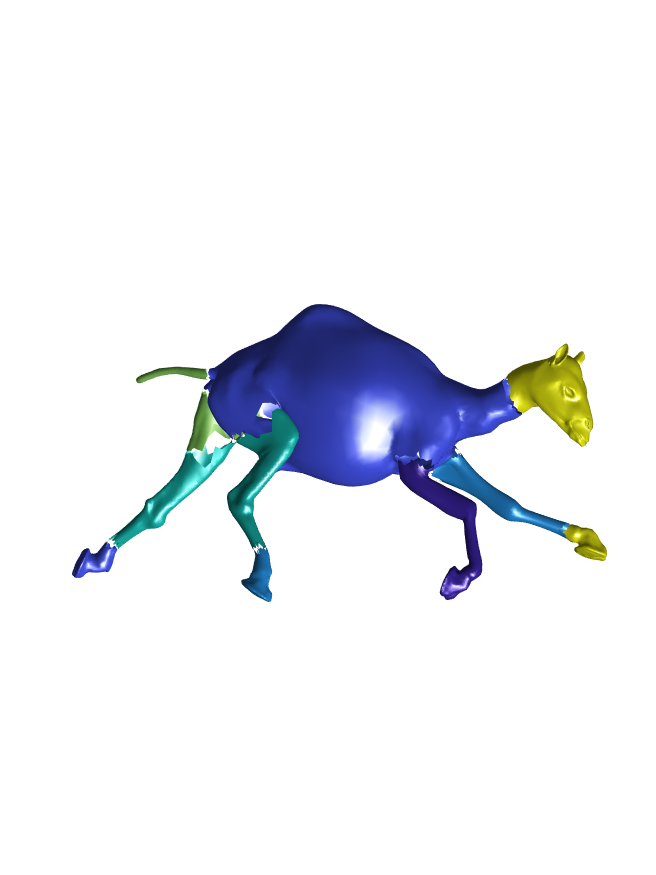}
		\label{fig:CamelHierarchicalOptimizeA}
	}
	\subfigure[]{
		\includegraphics[width=0.28\textwidth,trim=1cm 4cm 1cm 5cm, clip]{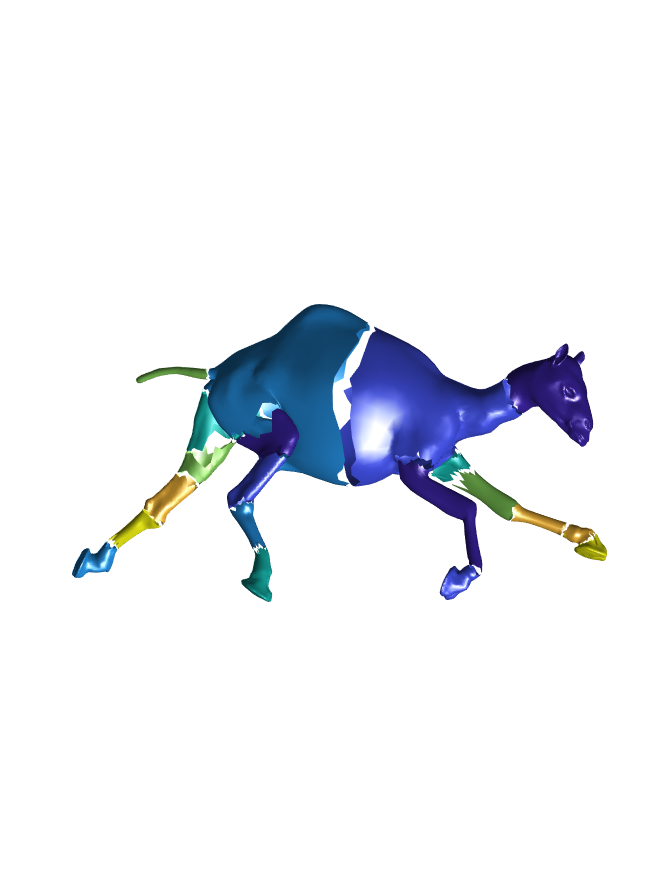}
		\label{fig:CamelHierarchicalOptimizeB}
	}
	\subfigure[]{
		\includegraphics[width=0.28\textwidth,trim=1cm 4cm 1cm 5cm, clip]{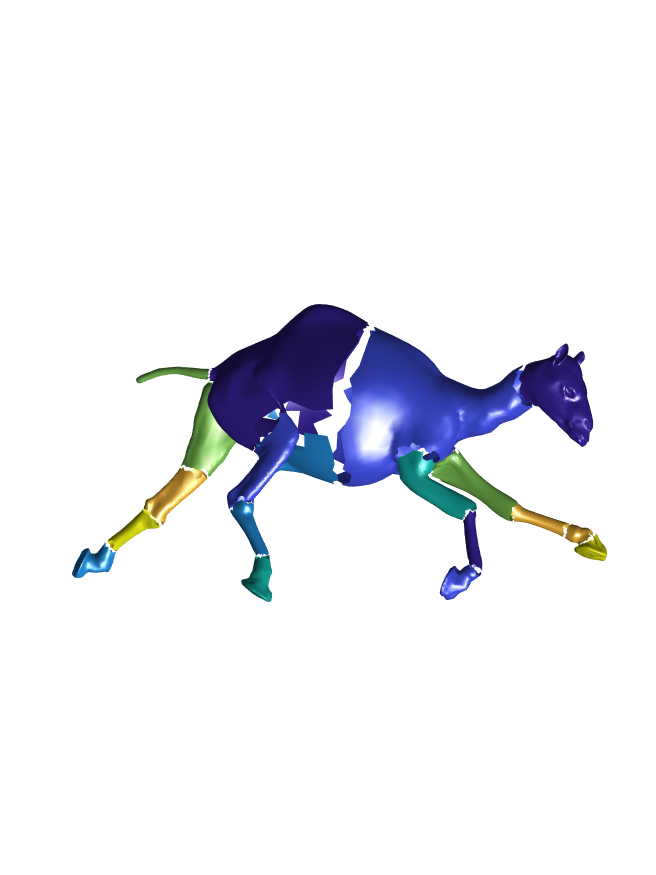}
		\label{fig:CamelHierarchicalOptimizeC}
	}
	\caption{Hierarchical optimization of the camel example: a) initial segmentation with 13 clusters; b) refined segmentation with 19 clusters; c) final segmetation with 19 clusters.}
	\label{fig:CamelHierarchicalOptimize}
\end{figure}

\begin{figure}[ht]
	\centering
	\subfigure[\scriptsize$\mb S_2$]{
		\includegraphics[width=0.27\textwidth,trim=0.5cm 3cm 0.5cm 3cm, clip]{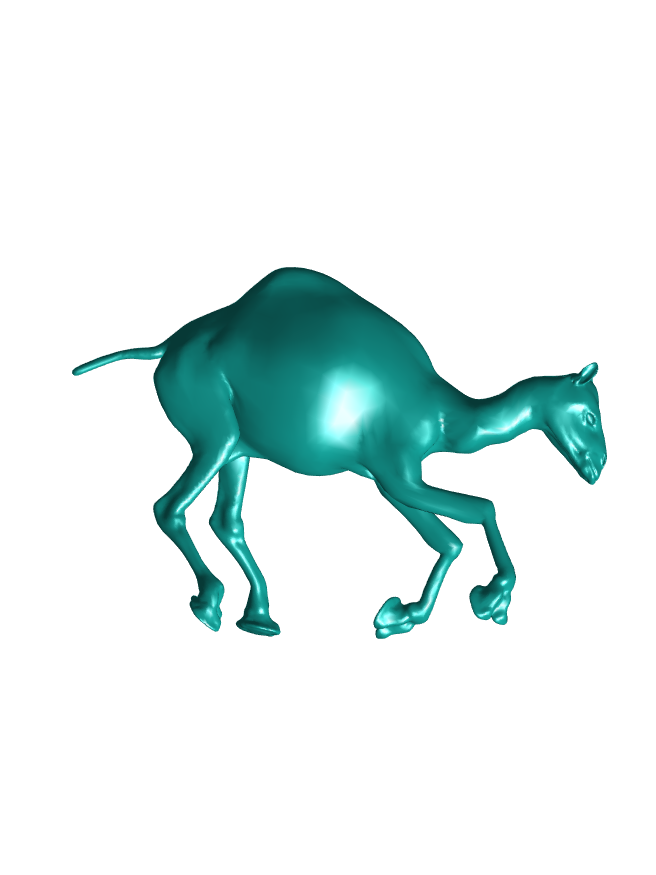}
		\label{fig:camelPoseInterpolate2A}
	}
	\subfigure[\scriptsize$0.75\mb S_2 + 0.25\mb S_{5}$]{
		\includegraphics[width=0.26\textwidth,trim=0.5cm 3cm 0.5cm 3cm, clip]{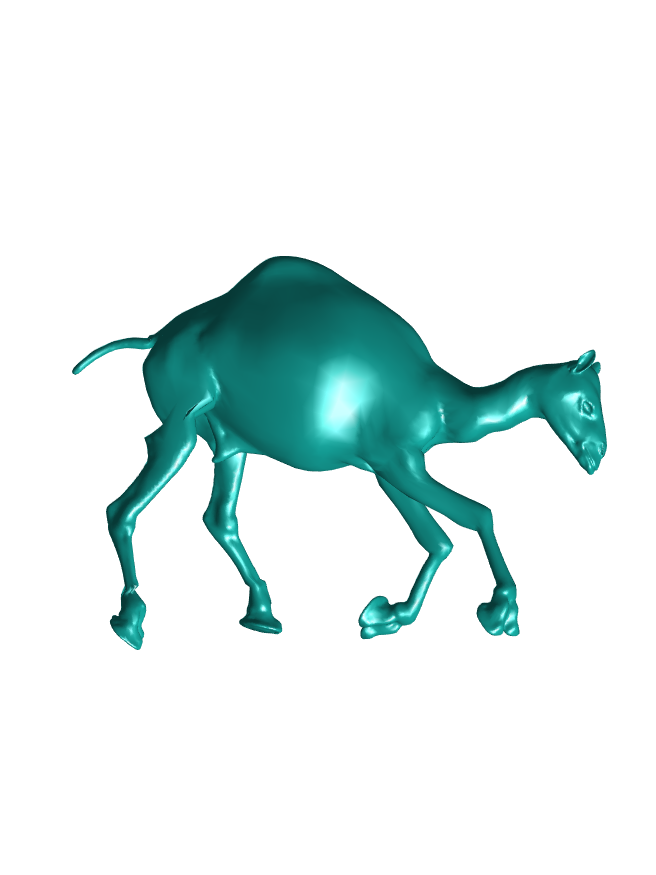}
		\label{fig:camelPoseInterpolate2B}
	}
	\subfigure[\scriptsize$0.5\mb S_2 + 0.5\mb S_{5}$]{
		\includegraphics[width=0.26\textwidth,trim=0.5cm 3cm 0.5cm 3cm, clip]{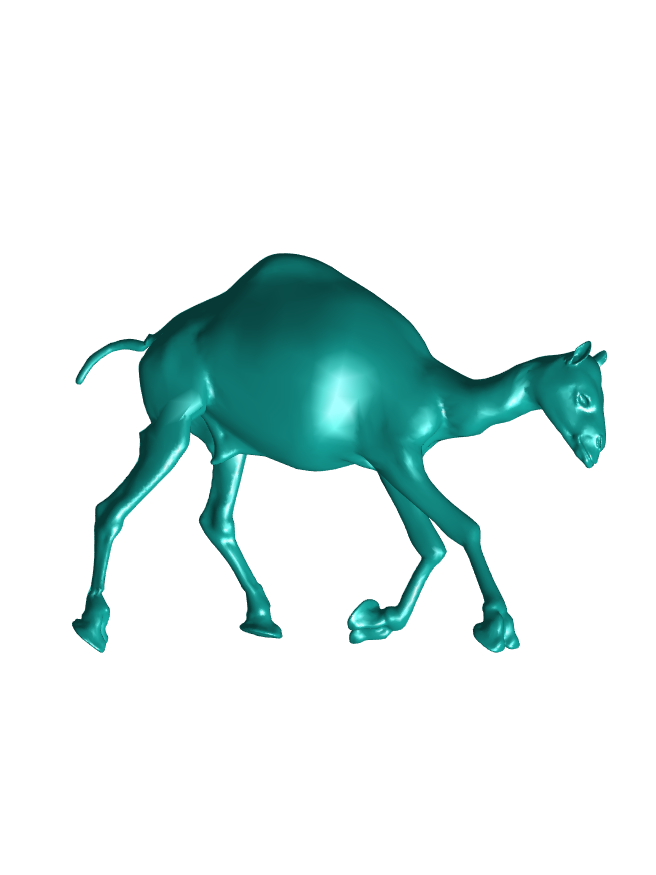}
		\label{fig:camelPoseInterpolate2C}
	}
	\subfigure[\scriptsize$0.25\mb S_2 + 0.75\mb S_{5}$]{
		\includegraphics[width=0.26\textwidth,trim=0.5cm 3cm 0.5cm 3cm, clip]{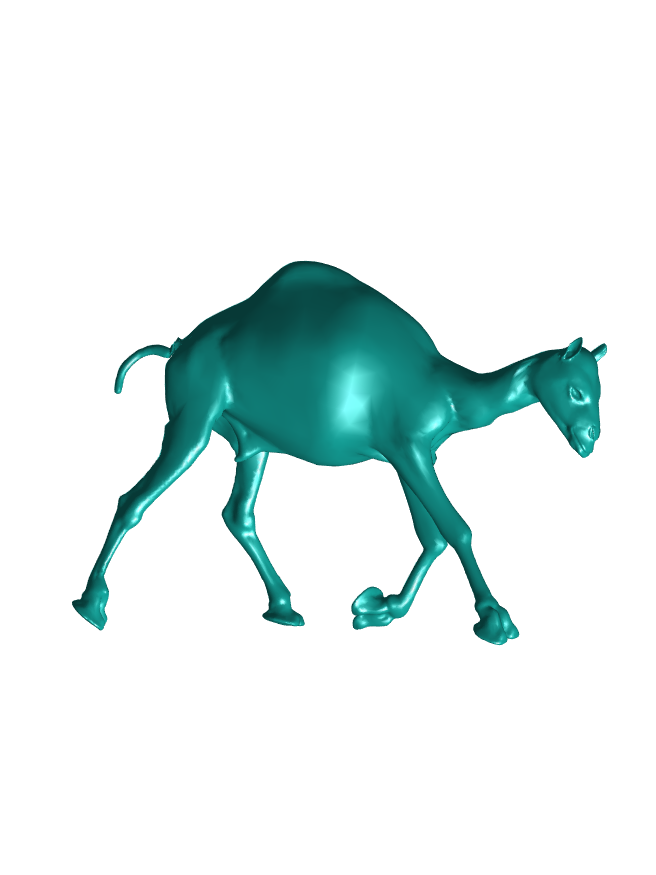}
		\label{fig:camelPoseInterpolate2D}
	}
	\subfigure[\scriptsize$\mb S_{5}$]{
		\includegraphics[width=0.26\textwidth,trim=0.5cm 3cm 0.5cm 3cm, clip]{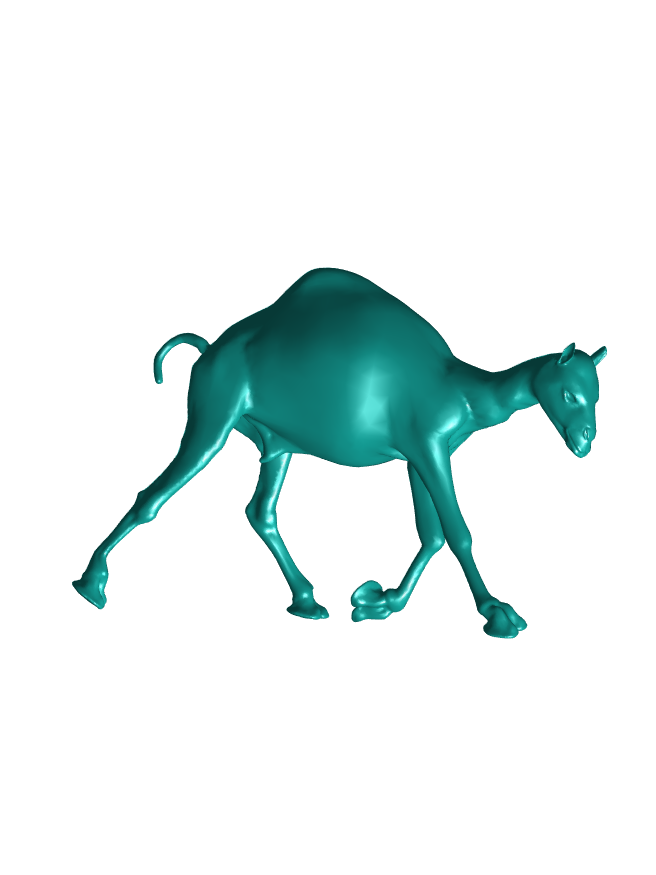}
		\label{fig:camelPoseInterpolate2E}
	}
	\caption{Pose interpolation between two camel shapes.}
	\label{fig:camelPoseInterpolate2}
\end{figure}

\begin{figure}[ht]
	\centering
		\includegraphics[width=0.33\textwidth,trim=1cm 4cm 0.5cm 4cm, clip]{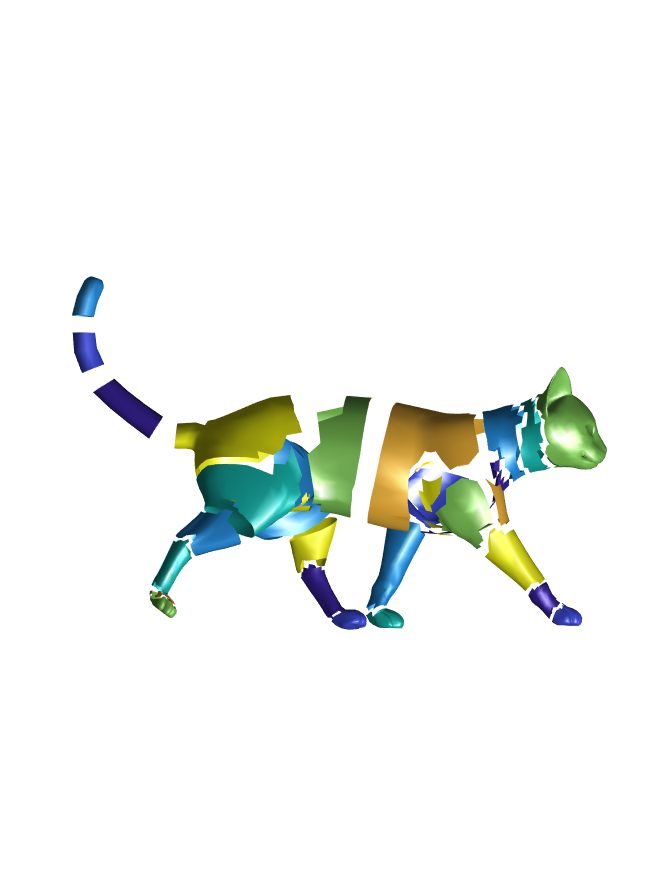}
	\caption{Articulated segmentation result of the cat example.}
	\label{fig:catHierarchicalOptimize}
\end{figure}

\begin{figure}[ht]
	\centering
	\subfigure[\scriptsize$\mb S_1$]{
		\includegraphics[width=0.27\textwidth,trim=0.5cm 4cm 0.5cm 3cm, clip]{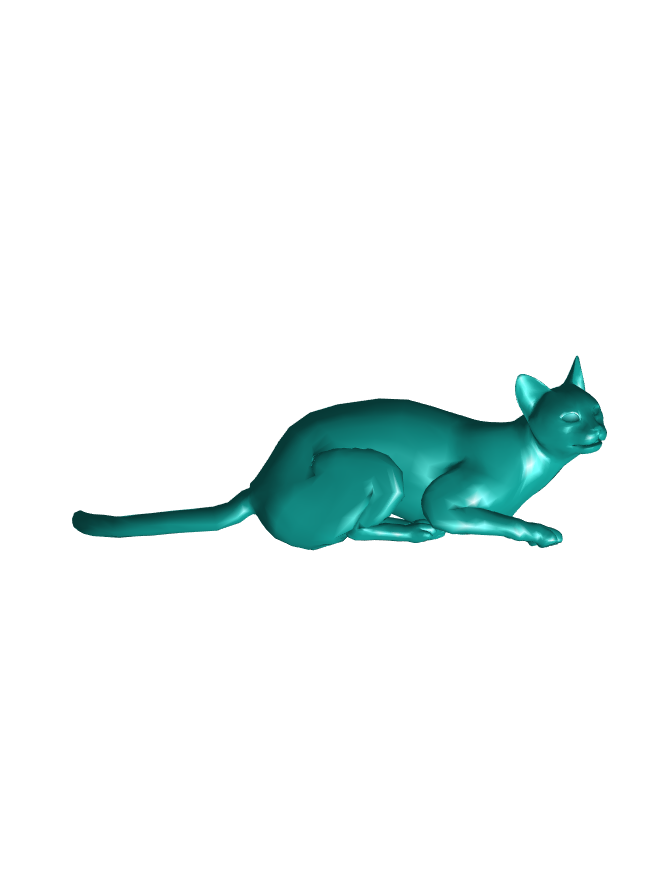}
		\label{fig:catPoseInterpolate1A}
	}
	\subfigure[\scriptsize$0.5\mb S_1 + 0.5\mb S_{10}$]{
		\includegraphics[width=0.27\textwidth,trim=0.5cm 4cm 0.5cm 3cm, clip]{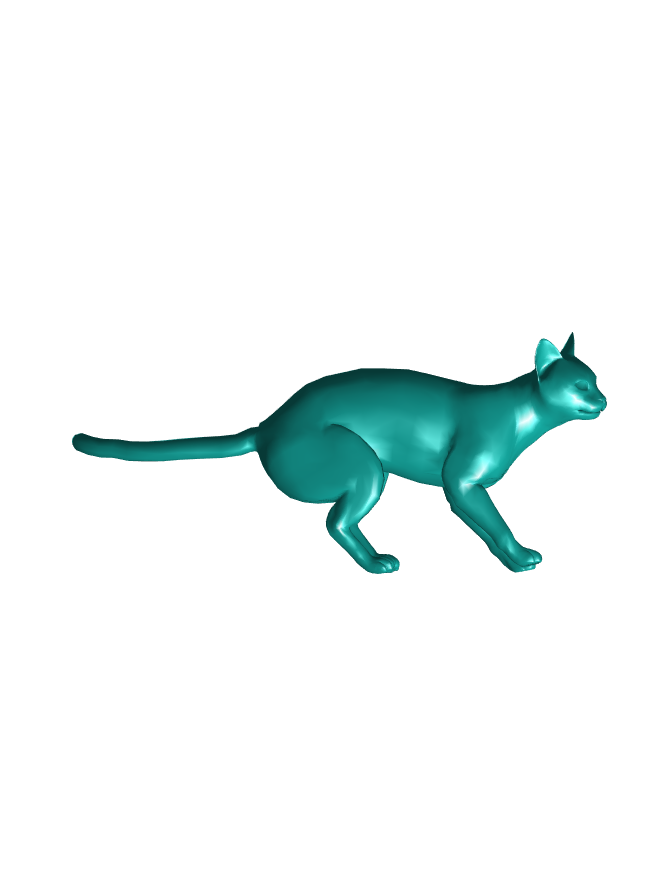}
		\label{fig:catPoseInterpolate1C}
	}
	\subfigure[\scriptsize$\mb S_{10}$]{
		\includegraphics[width=0.27\textwidth,trim=0.5cm 4cm 0.5cm 3cm, clip]{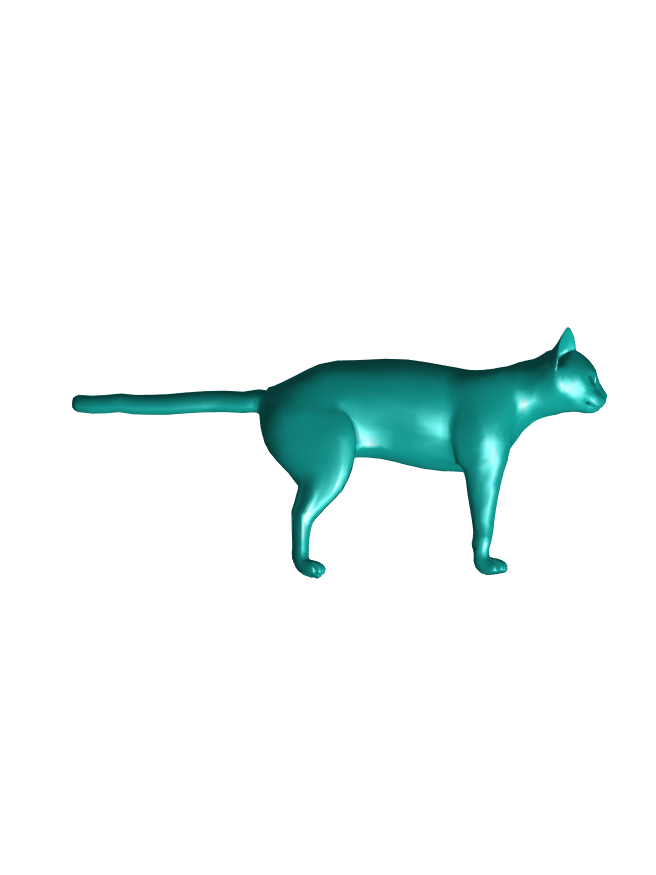}
		\label{fig:catPoseInterpolate1E}
	}
	\caption{Pose interpolation between two cat shapes.}
	\label{fig:catPoseInterpolate1}
\end{figure}

\section{Conclusion}

In this paper, an method based on mixtures of factor analyzers is proposed to learn the pose variations from the given shape population. The inputs are the training shapes and the initial number of mixtures. The outputs are the automatically refined mixtures from which we have the vertices labels and the rotations and translations associated with each rigid part of the training shapes. The spatially coherent articulated segmentation is guaranteed by the fact that the latent space and the physical space are homotopic. In all the examples the MFA algorithm converges in less than 20 iterations. The results have demonstrated the effectiveness of the proposed approach. The contributions of this work include: 1) formulating the problem of learning pose variation as the problem of mixtures of factor analyzers; 2) a hierarchical optimization algorithm to automatically refine the learned factor analyzers; 3) derivation of the closed form solution of the optimal factor loading matrices under the constraints that it is composed by rotation matrices; 4) Lemma 1 to gaurantee that the obtained rotation matrices are right handed; 5) a breadth first search algorithm for pose interpolation based on the obtained factor analyzers. Future work includes extending the method to larger data set of 3D shapes such as the CAESAR project \cite{robinette2002civilian}.

\section{Appendix}

In this section we derive the optimal values of the parameters $\{\pi_k\}$, $\{\mb b_k\}$, $\{\bs\Lambda_k\}$, $\{\mb R_k\}$, and $\{\bs\Phi_k\}$ in the $l+1$th iteration of the AECM algorithm. Though the optimal values of $\{\pi_k^{l+1}\}$ and $\{\mb b_k^{l+1}\}$ are known results in \cite{MclachlanModelling}, we still present their derivation for the self-containing of this paper.

The parameters $\{\pi_k^{l+1}\}$, $\{\mb b_k^{l+1}\}$, $\{\bs\Lambda_k^{l+1}\}$, $\{\mb R_k^{l+1}\}$, $\{\bs\Phi_k^{l+1}\}$ are derived following their order in the AECM algorithm, since the former, for example $\{\pi_k^{l+1}\}$ and $\{\mb b_k^{l+1}\}$, will be used in the derivation of their later parameters.

Observed that only the second term of Equation \eqref{eqn:EMFormula} contains $\{\pi_k\}$, the optimal values of $\{\pi_k\}$ in Equation \eqref{eqn:tildepi} is obtained by letting the derivatives of \eqref{eqn:piDerivation}  equal to zeros:
\begin{eqnarray}
\sum_{k=1}^{m}\left(\sum_{j = 1}^{n_v}\gamma_{kj}^{l+1}\right)\log\pi_k - \lambda(\sum_{k=1}^{m}\pi_k - 1),
\label{eqn:piDerivation}
\end{eqnarray}
where $\lambda(\sum_{k=1}^{m}\pi_k - 1)$ is the Lagrange that ensures the sum of unity of $\{\pi_k\}$.


The other parameters are in the first term of Equation \eqref{eqn:EMFormula}, factoring out the probability $P(\mb h_j| \mb z, \bs\Theta_k)$ in \eqref{eqn:EMFormula} and dropping the terms that are irrelevant to the parameters we have:
{\scriptsize
\begin{eqnarray}
f &=&-\frac{n_v}{2}\sum_{k=1}^{m}\pi_k^{l+1}\log |\boldsymbol{\Phi}_k|\nonumber\\
&&-\frac{1}{2}\sum_{k=1}^{m}\text{trace}\left(\boldsymbol{\Phi}_k^{-1}\sum_{j=1}^{n_v}\gamma_{kj}^{l+1}(\mb h_j-\mb b_k)(\mb h_j-\mb b_k)^T\right)\nonumber\\
&&+\sum_{k=1}^{m}\text{trace}\left(\mb A_k^T\boldsymbol{\Phi}_k^{-1}\sum_{j=1}^{n_v}\gamma_{kj}^{l+1}(\mb h_j-\mb b_k)E^k(\mb z|\mb h_j)^T\right)\nonumber\\
&&-\frac{1}{2}\sum_{k=1}^{m}\text{trace}\left(\mb A_k^T\boldsymbol{\Phi}_k^{-1}\mb A_k\sum_{j=1}^{n_v}\gamma_{kj}^{l+1}E^k(\mb z \mb z^T|\mb h_j)\right).
\label{eqn:detailedObjFunc}
\end{eqnarray}}

The optimal values of $\{\mb b_k\}$ in Equation \eqref{eqn:tildeb} are obtained by taking the partial derivatives of \eqref{eqn:detailedObjFunc} with respect to $\{\mb b_k\}$ and letting the derivatives equal to zero.

The optimal values of $\{\bs\Lambda_k\}$ is estimated as in \eqref{eqn:optLambda}, since by \eqref{eqn:optLambda} the image of the $k$th mixture $\mb v^{ref}=\bs\Lambda_k^{l+1}\mb z$ in the latent space is scaled to have the same variances as the $k$th mixture (rigid part) of the $n_s$ number of shapes, then it is possible to align the image and the rigid parts by rotations and translations.

Assume the variances $\bs\Phi_k$ is isotropic in the $x,y,z$ direction, we have $\bs\Phi_k = \text{diag}(s_k^1\mb I,s_k^2\mb I,...,s_k^{n_s}\mb I)$, where $\mb I$ is identity matrix if size $3 X 3$. Since $\mb A_k = \mb R_k\Lambda_k$, Equation \eqref{eqn:detailedObjFunc} can be further factored out:
{\scriptsize
\begin{eqnarray}
f &=& -\frac{n_v}{2}\sum_{k=1}^{m}\pi_k^{l+1}\log |\boldsymbol{\Phi}_k|\nonumber\\
&&-\frac{1}{2}\sum_{k=1}^{m}\text{trace}\left(\boldsymbol{\Phi}_k^{-1}\sum_{j=1}^{n_v}\gamma_{kj}^{l+1}(\mb h_j-\mb b_k^{l+1})(\mb h_j-\mb b_k^{l+1})^T\right)\nonumber\\
&&+\sum_{k=1}^{m}\sum_{i=1}^{n_s}\text{trace}\left(\bs\Lambda_k^{l+1}\sum_{j=1}^{n_v}\gamma_{kj}^{l+1}E^k(\mb z|\mb h_j)(\mb v_j^i-\mb b_k^{i,l+1})^T\bs\Phi_k^{-1}\mb R_k^i\right)\nonumber\\
&&-\frac{1}{2}\sum_{k=1}^{m}\sum_{i=1}^{n_s}\text{trace}\left({\bs\Lambda_k^{l+1}}^T\bs\Lambda_k^{l+1}/s_k^i\sum_{j=1}^{n_v}\gamma_{kj}^{l+1}E^k(\mb z \mb z^T|\mb h_j)\right).
\label{eqn:Rotdetail}
\end{eqnarray}}

It can be seen that only the third term of \eqref{eqn:Rotdetail} contains
$\{\mb R_k^i\}$. Assume we have the singular value decomposition $\bs\Lambda_k^{l+1}\sum_{j=1}^{n_v}\gamma_{kj}^{l+1}E^k(\mb z|h_j)(\mb h_j^i-\mb b_k^{i,l+1})^T = \mb U_{ki}^{l+1}\mb D_{ki}^{l+1}\mb {V_{ki}^{l+1}}^T$, substitute it into the third term of \eqref{eqn:Rotdetail}:
\begin{eqnarray}
g &=&\sum_{k=1}^{m}\sum_{i=1}^{n_s}\text{trace}(\mb U_{ki}^{l+1}\mb D_{ki}^{l+1}{\mb V_{ki}^{l+1}}^T\mb R_k^{i})\nonumber\\
&=&\sum_{k=1}^{m}\sum_{i=1}^{n_s}\text{trace}(\mb D_{ki}^{l+1}{\mb V_{ki}^{l+1}}^T\mb R_k^{i}\mb U_{ki}^{l+1}).
\label{eqn:RotFactorOut}
\end{eqnarray}
Assume $g_{ki}=\text{trace}(\mb D_{ki}^{l+1}{\mb V_{ki}^{l+1}}^T\mb R_k^{i}\mb U_{ki}^{l+1})$, When ${\mb V_{ki}^{l+1}}^T\mb R_k^{i}\mb U_{ki}^{l+1} = \mb I$, we have the maximum value of $g_{ik}$:
\begin{eqnarray}
\max g_{ki} = \text{trace}(\mb D_{ki}^{l+1}),
\label{eqn:maxGValueRight}
\end{eqnarray}
as demonstrated in the work of Procrustes Analysis \cite{gower1975generalized,ross2004procrustes}. Thus we have the optimal rotation:
\begin{eqnarray}
\mb R_k^{i,l+1} = \mb V_{ki}^{l+1}{\mb U_{ki}^{l+1}}^T.
\label{eqn:ProcrustesVURight}
\end{eqnarray}
However, the formulation in \eqref{eqn:ProcrustesVURight} doesn't guarantee the right handedness of the obtained rotation, and in this study it is observed that in a few cases $\mb R_k^{i,l+1}$ is left handed, which means the actual shape is mirrored. To ensure that we always obtain right handed rotation matrix, in the case that $\mb V_{ki}^{l+1}{\mb U_{ki}^{l+1}}^T$ is left handed, we have:
\begin{eqnarray}
\mb R_k^{i,l+1} = \mb V_{ki}^{l+1}\tilde{\mb I}{\mb U_{ki}^{l+1}}^T,
\label{eqn:ProcrustesVULeft}
\end{eqnarray}
where $\tilde{\mb I} = \text{diag}(1,1,-1)$. The proof is in Lemma 1.

\textbf{Lemma 1}: Assume $\mb U\mb D\mb V^T$ is the singular decomposition of the $3\times 3$ matrix $\mb Q$, the singular values are ordered from large to small in $\mb D$, and $\tilde{\mb I} = \text{diag}(1,1,-1)$. When $\mb V\mb U^T$ is left handed, $\mb R = \mb V\tilde{\mb I}\mb U^T$ is the rotation that maximizes $\text{trace}(\mb Q\mb R)$ among the all right handed rotations and the maximum value is $\text{trace}(\mb D\tilde{\mb I})$.

\textbf{Proof}: "$\mb V\mb U^T$ is left handed" means that either $\mb V$ or $\mb U$ is left handed, let's assume $\mb V$ is left handed first, we have:
\begin{eqnarray}
\text{trace}(\mb Q\mb R) &=& \text{trace}(\mb U\mb D\mb V^T\mb R)\nonumber\\
&=& \text{trace}(\mb U\mb D\tilde{\mb I}\tilde{\mb I}\mb V^T\mb R)\nonumber\\
&=& \text{trace}(\mb D\tilde{\mb I}(\mb V\tilde{\mb I})^T\mb R\mb U)
\end{eqnarray}
It is obvious that $(\mb V\tilde{\mb I})^T\mb R\mb U$ is right handed rotation matrix. Assume $\theta$ the corresponding rotation angle and $[A_1, A_2, A_3]$ the corresponding rotation axis, note $c = \cos(\theta), s = \sin(\theta)$, we have
\begin{eqnarray}
\text{tr}(\mb Q\mb R) &=& \sum_{i=1}^{2}D_i(c+(1-c)A_i^2)-D_3(c+(1-c)A_3^2)\nonumber\\
&=& D_1 + D_2 - D_3 - (D_1-D_3)(1-c)A_2^2 \nonumber\\
&&-(D_2-D_3)(1-c)A_1^2 - (D_1+D_2)(1-c)A_3^2\nonumber\\
&\leq& D_1 + D_2 - D_3.
\end{eqnarray}
The second line is obtained by the fact that $A_1^2 + A_2^2 + A_3^2 = 1$. The less and equal relationship in the last line is obtained by the facts that $D_1\geq D_2\geq D_3$ and $c \leq 1$. The equal sign is achieved when $c = 1$, or equivalently when $(\mb V\tilde{\mb I})^T\mb R\mb U = \mb I$, from which we have $\mb R = \mb V\tilde{\mb I}\mb U^T$. The proof is similar when only $\mb U$ is left handed. To the best of my knowledge, Equation \eqref{eqn:ProcrustesVULeft} is not seen in the Procrustes Analysis literature. Its application is not limited to this paper and can also be applied in the Procrustes Analysis to ensure the right handedness of the obtained orthogonal matrix.

The optimal values of $\{\bs\Phi_k\}$ in Equation \eqref{eqn:tildePhi} are obtained by taking the partial derivatives of \eqref{eqn:detailedObjFunc} with respect to $\mb\Phi_k^{-1}$ and letting the derivatives equal to zero.

\bibliography{LearningPoseVariationsByMFA2}

\begin{thebibliography}{}

\bibitem[\protect\citename{Amberg {\em et~al.}, }2007]{amberg2007optimal}
Amberg, Brian, Romdhani, Sami, \& Vetter, Thomas. 2007.
\newblock Optimal step {Non-rigid} {ICP} algorithms for surface registration.
\newblock {\em Pages  1--8 of:} {\em Computer Vision and Pattern Recognition,
  2007. CVPR'07. IEEE Conference on}.
\newblock IEEE.

\bibitem[\protect\citename{Anguelov {\em et~al.},
  }2004]{Anguelov2012Recovering}
Anguelov, Dragomir, Koller, Daphne, Pang, Hoi~Cheung, Srinivasan, Praveen, \&
  Thrun, Sebastian. 2004.
\newblock Recovering Articulated Object Models from 3D Range Data.
\newblock {\em Proceedings of the 20th conference on Uncertainty in artificial
  intelligence},  18--26.

\bibitem[\protect\citename{Anguelov {\em et~al.}, }2005]{Anguelov2005SCAPE}
Anguelov, Dragomir, Srinivasan, Praveen, Koller, Daphne, Thrun, Sebastian, \&
  Davis, James. 2005.
\newblock SCAPE: Shape completion and animation of people.
\newblock {\em Acm Transactions on Graphics}, {\bf 24}(3), 408--416.

\bibitem[\protect\citename{Baek {\em et~al.}, }2010]{Baek2010Mixtures}
Baek, Jangsun, Mclachlan, Geoffrey~J, \& Flack, Lloyd~K. 2010.
\newblock Mixtures of Factor Analyzers with Common Factor Loadings:
  Applications to the Clustering and Visualization of High-Dimensional Data.
\newblock {\em IEEE Transactions on Pattern Analysis and Machine Intelligence},
  {\bf 32}(7), 1298--1309.

\bibitem[\protect\citename{Baek \& Lee, }2012]{baek2012parametric}
Baek, Seung-Yeob, \& Lee, Kunwoo. 2012.
\newblock Parametric human body shape modeling framework for human-centered
  product design.
\newblock {\em Computer-Aided Design}, {\bf 44}(1), 56--67.

\bibitem[\protect\citename{Boscaini {\em et~al.}, }2016]{boscaini2016learning}
Boscaini, Davide, Masci, Jonathan, Rodol{\`a}, Emanuele, \& Bronstein, Michael.
  2016.
\newblock Learning shape correspondence with anisotropic convolutional neural
  networks.
\newblock {\em Pages  3189--3197 of:} {\em Advances in neural information
  processing systems}.

\bibitem[\protect\citename{Bronstein {\em et~al.},
  }2017]{bronstein2017geometric}
Bronstein, Michael~M, Bruna, Joan, LeCun, Yann, Szlam, Arthur, \&
  Vandergheynst, Pierre. 2017.
\newblock Geometric deep learning: going beyond euclidean data.
\newblock {\em IEEE Signal Processing Magazine}, {\bf 34}(4), 18--42.

\bibitem[\protect\citename{Chu {\em et~al.}, }2010]{chu2010exemplar}
Chu, Chih-Hsing, Tsai, Ya-Tien, Wang, Charlie~CL, \& Kwok, Tsz-Ho. 2010.
\newblock Exemplar-based statistical model for semantic parametric design of
  human body.
\newblock {\em Computers in Industry}, {\bf 61}(6), 541--549.

\bibitem[\protect\citename{De~Aguiar {\em et~al.}, }2008]{de2008automatic}
De~Aguiar, Edilson, Theobalt, Christian, Thrun, Sebastian, \& Seidel,
  Hans-Peter. 2008.
\newblock Automatic conversion of mesh animations into skeleton-based
  animations.
\newblock {\em Pages  389--397 of:} {\em Computer Graphics Forum},  vol. 27.
\newblock Wiley Online Library.

\bibitem[\protect\citename{Dryden \& Mardia, }1998]{dryden1998statistical}
Dryden, Ian~L, \& Mardia, Kanti~V. 1998.
\newblock {\em Statistical Shape Analysis}.
\newblock  Vol. 4.
\newblock J. Wiley Chichester.

\bibitem[\protect\citename{Ghahramani {\em et~al.},
  }1996]{ghahramani1996algorithm}
Ghahramani, Zoubin, Hinton, Geoffrey~E, {\em et~al.} 1996.
\newblock {\em The EM algorithm for mixtures of factor analyzers}.
\newblock Tech. rept. Technical Report CRG-TR-96-1, University of Toronto.

\bibitem[\protect\citename{Gower, }1975]{gower1975generalized}
Gower, John~C. 1975.
\newblock Generalized procrustes analysis.
\newblock {\em Psychometrika}, {\bf 40}(1), 33--51.

\bibitem[\protect\citename{Hasler {\em et~al.}, }2009]{Hasler2009A}
Hasler, Nils, Stoll, Carsten, Sunkel, Martin, Rosenhahn, Bodo, \& Seidel,
  Hans~Peter. 2009.
\newblock A Statistical Model of Human Pose and Body Shape.
\newblock {\em Computer Graphics Forum}, {\bf 28}(2), 337--346.

\bibitem[\protect\citename{Heimann \& Meinzer, }2009]{heimann2009statistical}
Heimann, Tobias, \& Meinzer, Hans-Peter. 2009.
\newblock Statistical shape models for {3D} medical image segmentation: a
  review.
\newblock {\em Medical Image Analysis}, {\bf 13}(4), 543--563.

\bibitem[\protect\citename{Hinton \& Salakhutdinov, }2006]{hinton2006reducing}
Hinton, Geoffrey~E, \& Salakhutdinov, Ruslan~R. 2006.
\newblock Reducing the dimensionality of data with neural networks.
\newblock {\em science}, {\bf 313}(5786), 504--507.

\bibitem[\protect\citename{James \& Twigg, }2005]{James2005Skinning}
James, Doug~L., \& Twigg, Christopher~D. 2005.
\newblock Skinning mesh animations.
\newblock {\em Acm Transactions on Graphics}, {\bf 24}(3), 399--407.

\bibitem[\protect\citename{Kabsch, }1978]{kabsch1978discussion}
Kabsch, Wolfgang. 1978.
\newblock A discussion of the solution for the best rotation to relate two sets
  of vectors.
\newblock {\em Acta Crystallographica Section A: Crystal Physics, Diffraction,
  Theoretical and General Crystallography}, {\bf 34}(5), 827--828.

\bibitem[\protect\citename{Le \& Deng, }2012]{le2012smooth}
Le, Binh~Huy, \& Deng, Zhigang. 2012.
\newblock Smooth skinning decomposition with rigid bones.
\newblock {\em ACM Transactions on Graphics (TOG)}, {\bf 31}(6), 1--10.

\bibitem[\protect\citename{Lengyel, }2001]{lengyel2001mathematics}
Lengyel, Eric. 2001.
\newblock {\em Mathematics for 3D game programming and computer graphics}.
\newblock Charles River Media, Inc.

\bibitem[\protect\citename{Li {\em et~al.}, }2017]{li2017grass}
Li, Jun, Xu, Kai, Chaudhuri, Siddhartha, Yumer, Ersin, Zhang, Hao, \& Guibas,
  Leonidas. 2017.
\newblock Grass: Generative recursive autoencoders for shape structures.
\newblock {\em ACM Transactions on Graphics (TOG)}, {\bf 36}(4), 1--14.

\bibitem[\protect\citename{Litany {\em et~al.}, }2017]{litany2017deep}
Litany, Or, Remez, Tal, Rodol{\`a}, Emanuele, Bronstein, Alex, \& Bronstein,
  Michael. 2017.
\newblock Deep functional maps: Structured prediction for dense shape
  correspondence.
\newblock {\em Pages  5659--5667 of:} {\em Proceedings of the IEEE
  International Conference on Computer Vision}.

\bibitem[\protect\citename{Mclachlan {\em et~al.}, }2003]{MclachlanModelling}
Mclachlan, G.~J., Peel, D., \& Bean, R.W. 2003.
\newblock Modelling high-dimensional data by mixtures of factor analyzers.
\newblock {\em Computational Statistics and Data Analysis}, {\bf 41}(3-4),
  379--388.

\bibitem[\protect\citename{Meng, }1993]{Meng1993Maximum}
Meng, X.~L. 1993.
\newblock Maximum likelihood estimation via the ECM algorithm: A general
  frame-work.
\newblock {\em Biometrika}, {\bf 80}(2), 267--278.

\bibitem[\protect\citename{Meng \& Van~Dyk, }1997]{meng1997algorithm}
Meng, Xiao-Li, \& Van~Dyk, David. 1997.
\newblock The EM algorithm—an old folk-song sung to a fast new tune.
\newblock {\em Journal of the Royal Statistical Society: Series B (Statistical
  Methodology)}, {\bf 59}(3), 511--567.

\bibitem[\protect\citename{Nash \& Williams, }2017]{nash2017shape}
Nash, Charlie, \& Williams, Christopher~KI. 2017.
\newblock The shape variational autoencoder: A deep generative model of
  part-segmented 3D objects.
\newblock {\em Pages  1--12 of:} {\em Computer Graphics Forum},  vol. 36.
\newblock Wiley Online Library.

\bibitem[\protect\citename{Robinette {\em et~al.},
  }2002]{robinette2002civilian}
Robinette, Kathleen~M, Blackwell, Sherri, Daanen, Hein, Boehmer, Mark, \&
  Fleming, Scott. 2002.
\newblock {\em Civilian American and European Surface Anthropometry Resource
  (CAESAR), Final Report. Volume 1. Summary}.
\newblock Tech. rept. DTIC Document.

\bibitem[\protect\citename{Ross, }2004]{ross2004procrustes}
Ross, Amy. 2004.
\newblock Procrustes analysis.
\newblock {\em Course report, Department of Computer Science and Engineering,
  University of South Carolina}, {\bf 26}.

\bibitem[\protect\citename{Schaefer \& Yuksel, }2007]{Schaefer2007Example}
Schaefer, Scott, \& Yuksel, Can. 2007.
\newblock Example-based skeleton extraction.
\newblock {\em In:} {\em Proceedings of the Fifth Eurographics Symposium on
  Geometry Processing, Barcelona, Spain, July 4-6, 2007}.

\bibitem[\protect\citename{Shu {\em et~al.}, }2018]{shu2018deforming}
Shu, Zhixin, Sahasrabudhe, Mihir, Alp~Guler, Riza, Samaras, Dimitris, Paragios,
  Nikos, \& Kokkinos, Iasonas. 2018.
\newblock Deforming autoencoders: Unsupervised disentangling of shape and
  appearance.
\newblock {\em Pages  650--665 of:} {\em Proceedings of the European Conference
  on Computer Vision (ECCV)}.

\bibitem[\protect\citename{Sumner \& Popovi{\'c}, }2004]{sumner2004deformation}
Sumner, Robert~W, \& Popovi{\'c}, Jovan. 2004.
\newblock Deformation transfer for triangle meshes.
\newblock {\em ACM Transactions on graphics (TOG)}, {\bf 23}(3), 399--405.

\bibitem[\protect\citename{Tang {\em et~al.}, }2012]{tang2012deep}
Tang, Yichuan, Salakhutdinov, Ruslan, \& Hinton, Geoffrey. 2012.
\newblock Deep mixtures of factor analysers.
\newblock {\em In:} {\em 29th International Conference on Machine Learning,
  ICML 2012}.

\bibitem[\protect\citename{Tierny {\em et~al.}, }2008]{tierny2008fast}
Tierny, Julien, Vandeborre, Jean-Philippe, \& Daoudi, Mohamed. 2008.
\newblock Fast and precise kinematic skeleton extraction of 3d dynamic meshes.
\newblock {\em Pages  1--4 of:} {\em 2008 19th International Conference on
  Pattern Recognition}.
\newblock IEEE.

\bibitem[\protect\citename{Wang \& Qian, }2016]{wang2016statistical}
Wang, Xilu, \& Qian, Xiaoping. 2016.
\newblock A statistical atlas based approach to automated subject-specific {FE}
  modeling.
\newblock {\em Computer-Aided Design}, {\bf 70}, 67--77.

\bibitem[\protect\citename{Yi {\em et~al.}, }2017]{yi2017syncspeccnn}
Yi, Li, Su, Hao, Guo, Xingwen, \& Guibas, Leonidas~J. 2017.
\newblock Syncspeccnn: Synchronized spectral cnn for 3d shape segmentation.
\newblock {\em Pages  2282--2290 of:} {\em Proceedings of the IEEE Conference
  on Computer Vision and Pattern Recognition}.

\end{thebibliography}

\end{document}